%% file: main.tex
\documentclass[a4paper]{article}
\usepackage[margin=1in]{geometry}
\usepackage{times}
\usepackage{hyperref}
\title{Generalization of LiNGAM that allows confounding}

\author{Joe Suzuki and Tian-Le Yang}
\usepackage{subfigure} 

\usepackage{multirow} 
\usepackage{tikz}
\usepackage{amssymb}
\usepackage{graphicx}
\usepackage{ascmac}
\usepackage{wrapfig}
\usepackage{color}
\newtheorem{teiri}{Theorem}
\newtheorem{prop}{Proposition}

\newtheorem{algo}{Algorithm}

\def\ci{\perp\!\!\!\perp}
\usepackage{natbib}
\begin{document}

\maketitle

\begin{abstract}
LiNGAM determines the variable order from cause to effect using additive noise models, but it faces challenges with confounding. Previous methods maintained LiNGAM's fundamental structure while trying to identify and address variables affected by confounding. As a result, these methods required significant computational resources regardless of the presence of confounding, and they did not ensure the detection of all confounding types. In contrast, this paper enhances LiNGAM by introducing LiNGAM-MMI, a method that quantifies the magnitude of confounding using KL divergence and arranges the variables to minimize its impact. This method efficiently achieves a globally optimal variable order through the shortest path problem formulation. LiNGAM-MMI processes data as efficiently as traditional LiNGAM in scenarios without confounding while effectively addressing confounding situations. Our experimental results suggest that LiNGAM-MMI more accurately determines the correct variable order, both in the presence and absence of confounding. The code is in the supplementary file in this link: \url{https://github.com/SkyJoyTianle/ISIT2024}. 
\end{abstract}

\section{Introduction}
When multiple events occur, determining which is the cause and which is the effect is a problem we commonly encounter in daily life. This paper examines the challenge of finding the order from causes to effects when dealing with multiple random variables.

Similar to, yet distinct from, this type of causal inference is the problem of identifying the structure of a Bayesian Network (BN) \citep{pearl,spirtes,bollen,apo13}. A BN represents the conditional independence between random variables using a Directed Acyclic Graph (DAG). In a BN, each vertex represents a variable, and directed edges are drawn from vertices corresponding to
$X_1,\ldots,X_m$ to the vertex corresponding to $X$
when the joint distribution is expressed as the product of conditional probabilities $P(X|X_1,\ldots,X_m)$. However, the structure of a BN can be represented differently, depending on how the joint distribution is factorized, and the direction of the arrows does not necessarily indicate the direction of causality. For instance, $P(X,Y)=P(X)P(Y|X)=P(X|Y)P(Y)$ can be written in such a way that it is unclear whether the correct representation is $X\rightarrow Y$ or $Y\rightarrow X$.
We refer to different graph expression that shares the same distribution, such as $X\rightarrow Y$ and $Y\rightarrow X$ as {\it Markov equivalent} models.

\cite{kano} proposed a causal inference framework called the Additive Noise Model. Consider two variables $X,Y$, with means of zero that can be expressed as $Y=aX+e$ using some constant $a$ and noise $e$. The Additive Noise Model posits that $X$ and $e$ are independent, with $X$ being the cause and $Y$ the effect. Some constants $a'$ and noise $e'$ may exist, such as $X=a'Y+e'$. However, both\footnote{We write $X\ci Y$ to denote that $X$ and $Y$ are independent} $X\ci e$ and $Y\ci e'$ (i.e., the order is not identifiable) 
is equivalent to  $X$ and $Y$ being Gaussian. For more details, refer to section 2.2 of this paper. Assuming a non-Gaussian distribution, a procedure was developed to determine which of $X$ or $Y$ is the cause and which is the effect. This approach is known as LiNGAM (Linear non-Gaussian Acyclic Model \citep{shimizu06,shimizu11,apo13}).

However, the Additive Noise Model is considered to have a limited scope of application. It does not encompass cases where neither $X\ci e$ nor $Y\ci e'$ is true. This paper, along with existing studies in causal inference, refers to such cases as 'the presence of confounding.' As the number of variables $p$ increases, the likelihood of encountering situations without confounding diminishes.
Even when focusing specifically on LiNGAM, a significant amount of research in causal inference accommodates the presence of confounding. While existing studies will be discussed later, this paper offers a fundamentally different perspective.

Most of these studies (\cite{entner2010causal},\cite{tashiro14},\cite{RCD},\cite{saber2020},\cite{Wang2023} ) aim to identify variables affected by confounding, eliminate their influence, and then apply the traditional LiNGAM model, which does not account for confounding. As a result, they presuppose a specific type of confounding and require computation times that grow exponentially with the number of variables $p$. 
However, there has been scant discussion regarding the challenges of causal inference in the presence of confounding, specifically its computational intensity, which often renders it impractical.

This paper explores an approach to finding the order of variables without identifying those affected by confounding, by extending LiNGAM. It specifically aims to quantify the magnitude of confounding among $p$ variables and to determine the sequence of these variables that minimizes this confounding. For each of the $p!$ possible orders $X_1\rightarrow \ldots \rightarrow X_p$, a corresponding series of noise terms $e_1=X_1$, $e_i=X_i-\sum_{j=1}^{i-1}\beta_{i,j}X_j$ ($i=2,\ldots,p$) is identified, where $\beta_{i,j}$ are constants. Traditional LiNGAM presupposes that in one specific order of variables, $e_1,\ldots,e_p$ are independent, and this configuration represents the true model. This paper proposes to quantify the degree of confounding using the Kullback-Leibler (KL) divergence between the joint distribution of these noise terms $P(e_1,\ldots,e_p)$ and the product of their marginal probabilities $P(e_1)\cdots P(e_p)$. The assumption is that the order of variables $X_1\rightarrow \ldots \rightarrow X_p$ that minimizes the KL divergence represents the true. This approach encompasses the existing Additive Noise Model and LiNGAM as particular instances.

Regarding computation time, finding the order that minimizes confounding is addressed through the shortest path problem, as the KL divergence representing confounding can be expressed as the sum of $p-1$ mutual information quantities. Despite this, the worst-case computational complexity remains $O(p!)$. Nevertheless, when confounding is minimal, the computational demand decreases substantially, and in cases with no confounding, the procedure completes in a timeframe comparable to that of the standard LiNGAM.

When applying LiNGAM, the order is sequentially determined from the causal variables to the resultant ones. This method works well without confounding, but it is not optimal when noise is not independent, i.e., when confounding is present. For instance, consider the variables $X, Y, Z$; even if $X$ is independent of the residuals $Y_X, Z_X$ (the residuals of $Y, Z$ after removing the influence of $X$), it is possible that both $Y_X, Z_{XY}$ and $Z_X, Y_{XZ}$ are far from independent, where  $Y_{XZ}$ and $Z_{XY}$ are the residuals of $Y$ and $Z$ after removing the influence of $X, Z$, and $X, Y$, respectively. In such cases, a greedy search approach can fail. When conducting a causal search with a finite sample, it becomes impossible to reliably distinguish whether confounding is present compared to the true distribution. Therefore, issues related to greedy search can always potentially arise. However, a search based on the shortest path problem enables a more global approach, thereby avoiding these issues.

\begin{figure}
\input{fig031}
\caption{\label{fig09} 
Confounders in red of the right figure affect more than one variable. The previous methods can deal with only non-consecutive variables such as $X_1$ and $X_3$.
}
\end{figure}
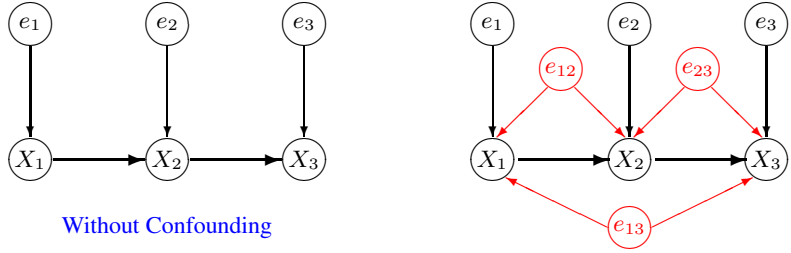

This method's versatility is further enhanced because it does not require prior knowledge of the confounding characteristics or assumptions. On the other hand, for example, the existing procedures preclude the case that confounders affect consecutive variables, which is fatal in real applications (Figure \ref{fig09}).

While existing methods demand exponential computation time regardless of the data, the approach proposed in this paper can be executed in a significantly shorter duration for most cases. Additionally, unlike current methods, it obviates the need for the BN structure learning procedure, simplifying implementation. 

In this paper, we refer our proposed procedure to the {\it LiNGAM-MMI} because it minimizes the total sum of MI along the path.
This paper provides the following contributions:
\begin{enumerate}
\item It quantifies confounding by using the KL divergence between the distributions of the noise and defines the true order as the one that minimizes it.
\item The optimal order is determined by formulating the problem as a shortest path problem, with the distance measured by the KL divergence. This global search approach requires relatively small computation, especially when the confounding is small.
\item Experiments show that the proposed procedure with the copula mutual information outperforms DirectLiNGAM \citep{shimizu06} with the Hilbert-Schmidt Independence Criterion (HSIC) \citep{hsic} even in the absence of confounding due to the global nature of the search and becomes more pronounced when $p$ (the number of variables) is large.
\end{enumerate}

The organization of this paper is as follows: Section 2 provides the necessary background information for understanding the results and places them in the context of existing work. Section 3 presents the main results of this study, with a particular focus on the underlying principles and methodologies. Section 4 illustrates these concepts with an example and experiments, assessing the effectiveness of LiNGAM-MMI. Section 5 concludes the paper with a summary of the findings and a discussion of potential directions for future research.

\subsection{Related Work}
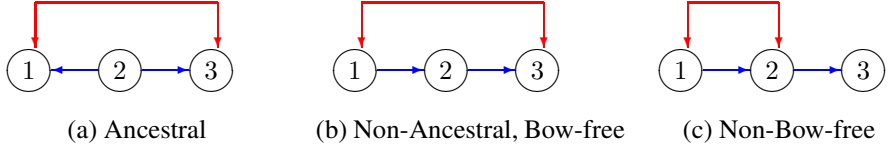
\begin{figure}
\input{ancestral}
\caption{\label{ancestral}
Ancestral and bow-free graphs. (a) The vertices 1 and 3 are siblings, but none of them are ancestors of the other. 
(b) The vertices 1 and 3 are siblings, but the former is an ancestor of the latter. (c) The vertices 1 and 2 consist of a bow.
}
\end{figure}

In a Directed Acyclic Graph (DAG), connecting the directed edges $u \rightarrow v$ and $v \rightarrow w$ creates a directed path from $u$ to $w$. Within this framework, $u$ is considered an ancestor of $w$. This interpretation holds irrespective of the number of directed edges involved; the origin of a directed path is consistently seen as the ancestor of its endpoint.
Moving on to mixed graphs, these incorporate edges with arrows in both directions, referred to as bidirectional edges such as $u\leftrightarrow v$, into a DAG. In such graphs, the simultaneous presence of directed edges and bidirectional edges is permissible (these are known as bows). 
A graph is defined as an ancestral graph if it does not include any pair of vertices that are simultaneously ancestors and siblings of each other (Figure \ref{ancestral}). In existing research, ancestral graphs, or their variants such as bow-free graphs, often represent confounding between two variables by using bidirected edges, like $u \leftrightarrow v$, between vertices.

Adding bidirected edges allows every ancestral graph $G$ to be transformed into a maximal ancestral graph (MAG) while preserving the conditional independence relations in $G$. However, such an MAG is not unique, and another Markov equivalent graph exists. They can be compactly represented by a partial ancestral graph (PAG) \citep{ali2009markov}. \cite{fci} proposed the Fast Causal Inference algorithm (FCI) to estimate the PAG corresponding to the underlying causal graph. \cite{Zhang2008-ZHAOTC-3} added additional orientation rules such that the output of FCI is complete. \cite{Colombo}, \cite{Claassen} and \cite{chen2021causal} followed in this direction. However, only an equivalence class of graphs a PAG represents can be discovered. 

In contrast, \cite{shimizu06} show that when the true model is a recursive linear SEM with non-Gaussian errors, the exact graph - not just an equivalence class - can be identified from observational data using independent component analysis (ICA). Instead of ICA, the subsequent DirectLiNGAM \citep{shimizu11} and Pairwise LiNGAM \citep{apo13} methods use an iterative procedure to estimate a causal ordering. 

\cite{hoyer08} consider the setting where a LiNGAM model generates the data, but some variables are unobserved. Using overcomplete ICA, they show that the canonical DAG can be identified when all parent-child pairs in the observed set are unconfounded. 
\cite{shimizu14} point out that "current versions of the overcomplete ICA algorithms are unreliable since they often suffer from local optima." 
To avoid using overcomplete ICA and improve practical performance, \cite{entner2010causal} and \cite{tashiro14} both propose procedures that test subsets of the observed variables and seek to identify as many pairwise ancestral relationships as possible. For more details, refer to section~\ref{sec2.5} of this paper. 

\cite{RCD} proposes the Repetitive Causal Discovery (RCD) method for discovering mixed graphs. In contrast to the approach by \cite{tashiro14}, RCD iteratively utilizes previously discovered structures to inform subsequent steps. Similar to \cite{hoyer08}, \cite{saber2020} employ overcomplete ICA, which crucially requires all confounding to be linear.
\cite{Wang2023} introduce a method for identifying bow-free confounding, diverging from existing works that have assumed confounding to be ancestral. However, all these studies, including \cite{Wang2023}, deal only with non-consecutive confounding, as shown in Figure \ref{fig09}.

The top author of this paper has published the LiNGAM-MMI for the binary case \citep{binary} while this paper deals with the structure equation models \citep{kano}.

\section{Preliminaries}
This section provides a background on covariance, independence, additive noise models, LiNGAM (Linear Gaussian Models), HSIC (Hilbert-Schmidt Information Criterion), and confounding factors.

\subsection{Covariance and Independence}
Let $X,Y, e$ be zero mean random variables\footnote{
Hereafter, we refer to the random variables simply as 'variables' when no confusion arises.
} related by 
\begin{eqnarray}\label{eq0}
Y=aX+e
\end{eqnarray}
 with $a\in {\mathbb R}$.
We determine the constant $a$ so that the covariance of $X,e$ is zero:
\begin{eqnarray}\label{eq1}
{\rm Cov}[X,e]={\rm Cov}[X,Y-aX]=0\ ,
\end{eqnarray}
which means 
\begin{eqnarray}\label{eq6}
a=\frac{{\rm Cov}[X,Y]}{V[X]}\ ,
\end{eqnarray}
where ${\rm Cov}[\cdot,\cdot]$ and $V[\cdot]$ are the covariance and variance operations.

Let $N(\mu,\sigma^2)$ be the Gaussian distribution with mean $\mu$ and variance $\sigma^2$.
We know that, in general, the converse of the implication
\begin{eqnarray}\label{eq2}
Z\ci W \Longrightarrow {\rm Cov}[Z,W]=0
\end{eqnarray}
is not true. 
For example, suppose $Z \sim N(0,1)$ and $W=Z^2$. They are not independent but
$${\rm Cov}[Z,W]={\rm Cov}[Z,Z^2]=E[(Z-0)(Z^2-1)]=E[Z^3]-E[Z]=0$$
because $E[Z]=E[Z^3]=0$ and $E[Z^2]=1$, where $E[\cdot]$ is the expectation operation.
In this sense, (\ref{eq1}) does not mean $X\ci e$.

On the other hand, suppose $U,V\sim N(0,1)$ with $U\ci V$.
Variables $Z,W$ are said to be jointly Gaussian if there exists a matrix $A\in {\mathbb R}^{2\times 2}$  such that 
\begin{eqnarray*}
\left[
\begin{array}{c}
Z\\
W
\end{array}
\right]
=A
\left[
\begin{array}{c}
U\\
V
\end{array}
\right]+
\left[
\begin{array}{c}
E[Z]\\
E[W]
\end{array}
\right]
\ .
\end{eqnarray*}
If $Z,W$ are jointly Gaussian, then they are Gaussian. Moreover, if $Z,W$ are jointly Gaussian, the converse ($\Longleftarrow$) of (\ref{eq2}) holds.
Just because $Z,W$ are Gaussian does not imply the converse of (\ref{eq2}).
In fact, let $Z\in N(0,1)$ and $R\in \{\pm 1\}$ equiprobable, and assume $Z\ci R$.
Then, $Z$ and $W=ZR$ are not independent although $E[R]=E[Z]=E[ZR]=0$ and 
$${\rm Cov}(Z,W) = E[(Z-0)(ZR-0)]=E[Z^2R]=E[Z^2]E[R]=0\ .$$
In this paper, we refer to $Z$ and $W$ as 'Gaussian' only when they are jointly Gaussian, provided this does not lead to confusion. 
Thus, if $X,e$ are Gaussian in (\ref{eq0}), we have
\begin{eqnarray}\label{eq3}
{\rm Cov}[X,e]=0 \Longleftrightarrow X\ci e\ .
\end{eqnarray}

\subsection{Additive Noise Model and LiNGAM}
\begin{figure}
\input{fig029}
\caption{
Additive Noise Model: Either $e_1\ci e_2$ ($X\rightarrow Y$) or $e_1'\ci e_2'$ ($Y\rightarrow X$) must hold, but not both.
}
\end{figure}
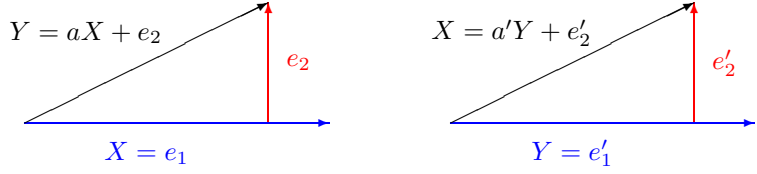
\citet{kano} considered a causal model in 
which cause $X$ and effect $Y$ are related by
\begin{eqnarray}\label{eq10a}
\left\{
\begin{array}{l}
X=e_1\\
Y=aX+e_2
\end{array}
\right.\ ,
\end{eqnarray}
where $a$ is given by (\ref{eq6}). 
In particular, they assumed that $e_1, e_2$ should be independent.
Then, we write $X\rightarrow Y$.
Similarly, we can construct the opposite model in which they are related by
\begin{eqnarray}\label{eq10b}
\left\{
\begin{array}{l}
Y=e_1'\\
X=a'Y+e_2'
\end{array}
\right.
\end{eqnarray}
with $e_1'\ci e_2'$, and write $Y\rightarrow X$.
They assumed that exactly one of (\ref{eq10a}) with  
$e_1\ci e_2$
and (\ref{eq10b}) with $e_1'\ci e_2'$ happens, and identifies which of $X,Y$ are the cause and effect, 
depending on which of (\ref{eq10a}) and (\ref{eq10b}) is correct. 
Note that $a$ and
$$\displaystyle a'=\frac{{\rm Cov}[X,Y]}{V[Y]}$$ are determined so that 
\begin{equation}\label{eq7}
\left\{
\begin{array}{l}
{\rm Cov}[e_1,e_2]=0\\
{\rm Cov}[e_1',e_2']=0
\end{array}
\right.\ .
\end{equation}
Suppose $a=0$, which is equivalent to $a'=0$. 
Then, $X,Y$ are independent, and we cannot identify the order. Thus, we remove such a case from the beginning.
We claim the order can be identified under $a,a'\not=0$ if and only if either of $e_1, e_2$ is not  Gaussian.

In order to show the claim, suppose that $e_1,e_2$ are Gaussian. Then, from (\ref{eq10a}) and (\ref{eq10b}), we observe that $X,Y,e_1',e_2'$ are Gaussian.
Moreover, from (\ref{eq3}), we have 
\begin{equation}\label{eq8}
\left\{
\begin{array}{l}
{\rm Cov}[e_1,e_2]=0 \Longleftrightarrow e_1\ci e_2\\
{\rm Cov}[e_1',e_2']=0\Longleftrightarrow e_1'\ci e_2'
\end{array}
\right.\ .
\end{equation}
From (\ref{eq7})(\ref{eq8}), we conclude that both of $e_1\ci e_2$ and $e_1'\ci e_2'$ occur.

On the other hand, suppose that $e_1\ci e_2$ and $e_1'\ci e_2'$ occur simultaneously. 
Noting that (\ref{eq10a}) and (\ref{eq10b}) imply
\begin{equation}\label{eq11}
\left\{
\begin{array}{l}
e_1'=ae_1+e_2\\
e_2'=(1-aa')e_1-a'e_2
\end{array}
\right.\ ,
\end{equation}
we consider applying the following proposition.
\begin{prop}[\cite{darmois,skito}]\rm\label{prop1}
Let $z,w,u,v$ be variables related by 
$$
\left\{
\begin{array}{l}
z=pu+qv\\
w=ru+sv
\end{array}
\right.$$
with $p,q,r,s\in {\mathbb R}$, and suppose $u\ci v$ and $z\ci w$.
Then, if $pr\not=0$, $u$ is Gaussian, and if $qs\not=0$, $v$ is Gaussian.
\end{prop}
Suppose $\displaystyle 1=aa'=\frac{{\rm Cov}[X,Y]^2}{V[X]V[Y]}$. Then, the relation between $X,Y$ is deterministic, which is excluded in our discussion.
Thus, under $a,a'\not=0$, we have $a(1-aa'), -a'\not=0$ in (\ref{eq11}). From Proposition \ref{prop1}, we conclude that $e_1.e_2$ are Gaussian.
The claim can be summarized as follows:
\begin{prop}[\cite{shimizu06}]\label{prop2}\rm
Under $a,a'\not=0$, the order can be identified if and only if either of $e_1,e_2$ is non-Gaussian.
\end{prop}
\cite{shimizu06} considered procedures ({\it LiNGAM}, linear non-Gaussian models) that identify the causal order based on Proposition \ref{prop2}.

\subsection{LiNGAM from data}
Suppose that we observe i.i.d. (independent and identically distributed) data
$x^n=(x_1,\ldots,x_n)$, $y^n=(y_1,\ldots,y_n)\in {\mathbb R}^n$ for variables $X,Y$ 
of sample size $n$.
Then, 
We perform statistical independence testing to evaluate whether $x^n \ci y_x^n$ or $y^n \ci x_y^n$, 
examining which of $e_1\ci e_2$ and $e_1'\ci e_2'$ is more likely, where
$$
\begin{array}{l}
\displaystyle y_x^n:=y^n-\frac{c(x^n,y^n)}{v(x^n)}x^n\\
\displaystyle x_y^n:=x^n-\frac{c(x^n,y^n)}{v(y^n)}y^n
\end{array}$$
with $v(\cdot)$ and $c(\cdot,\cdot)$ the variance and covariance based on the samples $x^n,y^n$.
For each of the two possibilities, we obtain the causal order as follows:
$$
\left\{
\begin{array}{lll}
x^n\ci y_x^n&\Longrightarrow&X\rightarrow Y\\
y^n\ci x_y^n&\Longrightarrow&Y\rightarrow X
\end{array}
\right.
$$

Next, suppose that we observe i.i.d. data
$z^n=(z_1,\ldots,z_n)\in {\mathbb R}^n$ for variable $Z$ as well as the $x^n,y^n$.
Then, we compare 
$$x^n\ci \{y_x^n, z_x^n\}\ ,\ y^n\ci \{z_y^n,x_y^n\}\ ,\ z^n\ci \{x_z^n,y_z^n\}\ ,$$
where
the quantities $z_x^n$, $z_y^n$, $x_z^n$, $y_z^n$ are defined similarly to $x_y^n$ and $y_x^n$.

Then, suppose that $x^n\ci \{y_x^n,z_x^n\}$ is the most likely among the three possibilities.
We define the quantity
\begin{eqnarray}
\displaystyle y_{x{z}}^n&:=&y_x^n-\frac{c(y_x^n,z_x^n)}{v(z_x^n)}z_x^n\nonumber\\
\displaystyle z_{x{y}}^n&:=&z_x^n-\frac{c(y_x^n,z_x^n)}{v(y_x^n)}y_x^n\label{eq116}
\end{eqnarray}
to compare $\displaystyle y_x^n \ci z_{x{y}}^n$ and $\displaystyle z_x^n \ci y_{x{z}}^n$.
If $\displaystyle y_x^n \ci z_{x{y}}^n$ as well as $x^n\ci \{y_x^n,z_x^n\}$ are the most likely, then we conclude that 
the order $X\rightarrow Y\rightarrow Z$ is the most likely among the six. Thus, we obtain the residue sequence $(x^n,y^n_x,z^n_{xy})$
by reducing the effects of the upper variables from the original samples $x^n,y^n,z^n$.

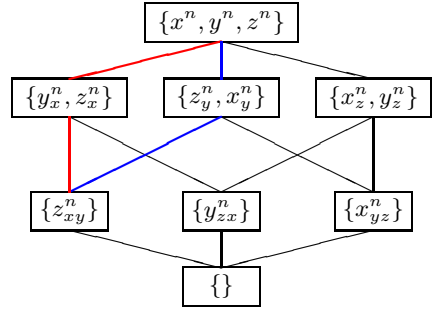
\begin{figure}
\input{fig030}
\caption{\label{fig091}In order to obtain the residue $z_{xy}^n$($=z_{yx}^n$), compute $\{y_x^n,z_x^n\}$ first and then obtain $\{z_{xy}^n\}$ via (\ref{eq116}), as 
shown in red, or compute $\{z_y^n, x_y^n\}$  first and then obtain $\{z_{yx}^n\}$ via (\ref{eq112}), as shown in blue.}
\end{figure}

Note that we can show the equality $z_{xy}^n=z_{yx}^n$ for 
\begin{equation}\label{eq112}
z_{yx}^n:=z_y^n-\frac{c(z_y^n,x_y^n)}{v(x_y^n)}x_y^n\\
\end{equation}
(see Appendix A). We obtain the same value in two ways (Figure \ref{fig091}).
Similarly, we have $x^n_{yz}=x^n_{zy}$ and $y^n_{zx}=y_{xz}$.
In general, the order of suffices in the residues does not matter, although, in this paper, we do not provide its derivation. 

For each of the six possibilities, we obtain the causal order as follows:
$$
\left\{
\begin{array}{ll}
x^n\ci \{y_x^n,z_x^n\}&
\left\{
\displaystyle
\begin{array}{lll}
y_x^n\ci z_{xy}^n&\Longrightarrow&X\rightarrow Y\rightarrow Z\\
z_x^n\ci y_{xz}^n&\Longrightarrow&X\rightarrow Z\rightarrow Y\\
\end{array}\right.\\
y^n\ci \{z_y^n,x_y^n\}&
\left\{
\begin{array}{lll}
z_y^n\ci x_{yz}^n&\Longrightarrow&Y\rightarrow Z\rightarrow X\\
x_y^n\ci z_{yx}^n&\Longrightarrow&X\rightarrow Z\rightarrow Y\\
\end{array}\right.\\
z^n\ci \{x_z^n,y_z^n\}&
\left\{\begin{array}{lll}
x_z^n\ci y_{zx}^n&\Longrightarrow&Z\rightarrow X\rightarrow Y\\
y_z^n\ci x_{zy}^n&\Longrightarrow&Z\rightarrow Y\rightarrow X\\
\end{array}
\right.
\end{array}
\right.
$$

The same procedure can be applied to any $p\geq 2$ variables rather than $p=2,3$.

\subsection{Statistical Testing using HSIC}
Let \({\cal X}\) be a set, and \(k: {\cal X} \times {\cal X} \rightarrow \mathbb{R}\) be a positive definite kernel.
It is known \citep{kernel} that there exists a unique reproducing kernel Hilbert space (RKHS) \(H\), the closure $\overline{\{k(x,\cdot)\}_{x\in {\cal X}}}$ of 
the linear space generated by \(\{k(x,\cdot)\}_{x\in {\cal X}}\).
Let \((k_{\cal X}, H_{\cal X})\) and \((k_{\cal Y}, H_{\cal Y})\) be pairs of positive definite kernel and RKHS for sets \({\cal X}\) and \({\cal Y}\).
We execute the test of independence between \(k_{\cal X}(X,\cdot) \in H_{\cal X}\) and
\(k_{\cal Y}(Y,\cdot) \in H_{\cal Y}\) rather than between \(X \in {\cal X}\) and \(Y \in {\cal Y}\),
and define the quantity called Hilbert-Schmidt information criterion (HSIC) as follows:
\[{\rm HSIC}(X,Y) := \|{E}_{XY}[k_{\cal X}(X,\cdot)k_{\cal Y}(Y,\cdot)]-{E}_X[k_{\cal X}(X,\cdot)]{E}_Y[k_{\cal Y}(Y,\cdot)]\|^2_H\ ,\]
where \(E_X[\cdot]\), \(E_Y[\cdot]\), and \(E_{XY}[\cdot]\) denote the expectations with respect to \(X\), \(Y\), and \((X,Y)\), respectively,
and 
\(\|\cdot\|_{H}\) is the norm defined in the tensored Hilbert space \(H := H_{\cal X} \otimes H_{\cal Y}\).
It is known that 
\begin{equation}\label{hsic}
HSIC(X,Y) = 0  \Longleftrightarrow X \ci Y
\end{equation}
if the kernels are characteristic \citep{kernel}.
However, the $HSIC(X,Y)$ value is not known, 
and we infer whether $HSIC(X,Y)=0$ or not by examining its estimate ${HSIC}_n(x^n,y^n)$ given realizations $\{(x_i,y_i)\}_{i=1}^n$ of $(X,Y)$ with 
$x^n=(x_1,\ldots,x_n)$, $y^n=(y_1,\ldots,n_n)$.

In the test of independence, we see whether statistics \(T := {\rm HSIC}_n(x^n,y^n)\) follows the distribution \(f_{X \ci Y}(t)\) under
the null hypothesis \(X \ci Y\). 
Given a significance level \(\alpha > 0\), 
if \(T\) is significantly large, i.e., \(T > T_\alpha\), where 
\[\alpha = \int_{T_\alpha}^\infty f_{X \ci Y}(t) dt\ , \]
we reject \(X \ci Y\). HSIC is well-known for its strong power in detecting non-independence and is often used in the LiNGAM procedure.

In the actual LiNGAM applications, to minimize computational costs, 
we compare \(T = {HSIC}_n(x^n, y^n_x)\) and \(T' = {HSIC}_n(y^n, x_y^n)\) rather than the $p$-values 
\(\int_{T}^\infty f_{e_1 \ci e_2}(t) \, dt\) and
\(\int_{T'}^\infty f_{e_1' \ci e_2'}(t) \, dt\)
to choose between\footnote{Since \(f_{e_1 \ci e_2}\) and \(f_{e_1' \ci e_2'}\) are different, it is not appropriate to compare
\(\widehat{HSIC}(e_1, e_2)\) and \(\widehat{HSIC}(e_1', e_2')\) for determining which pair is more likely to be independent.} \(e_1 \ci e_2\) and \(e_1' \ci e_2'\).

\subsection{Confounding}\label{sec2.5}
We say that \textit{confounding} exists for $X,Y$ if neither $e_1 \ci e_2$ nor $e_1' \ci e_2'$ holds.
Even if {confounding} exists, we may decide that $e_1 \ci e_2$ is more likely than $e_1' \ci e_2'$ 
if ${HSIC}_n(x^n, y_x^n)$ is smaller than ${HSIC}_n(y^n, x_y^n)$.
However, this paper's main issue is identifying the order when more than two variables exist. For example, if the noises $e_1, e_2, e_3$ in 
\[\left\{
\begin{array}{l}
X=e_1\\
Y=aX+e_2\\
Z=bX+cY+e_3
\end{array}
\right.\]
are independent for some $a, b, c \in \mathbb{R}$, we identify the order as $X \rightarrow Y \rightarrow Z$.
However, what if no independent noises exist for any order of $X, Y, Z$?

One might claim that the same strategy illustrated in Section 2.3 can be applied, such as comparing
\[{HSIC}_n(x^n, \{y^n_x, z^n_x\}), {HSIC}_n(y^n, \{z_{y}^n, x_y^n\}), {HSIC}_n(z^n, \{x_z^n, y_z^n\})\]
and, if the first value is the smallest, further compare ${HSIC}_n(y_x^n, z_{xy}^n)$ and ${HSIC}_n(z_x^n, y_{zx}^n)$, etc.
But, in that case, what merits can be gained by following such a strategy?
For example, what if both of 
${HSIC}_n(y_x^n, z_{xy}^n)$ and ${HSIC}_n(z_x^n, y_{zx}^n)$ have large values, which suggests that both of 
$y_x^n \ci z_{xy}^n$ and $z_x^n\ci y_{zx}^n$ are unlikely?
Then, we would admit that the decision either
$X\rightarrow Y\rightarrow Z$ or 
$X\rightarrow Z\rightarrow Y$ would be wrong.

When a confounder exists, estimating the variable order becomes more challenging.
Most of the previous works \citep{entner13,tashiro14} obtain the order for each maximal subset of variables that are not affected by any confounder and estimate the order among the whole variables by combining the orders among variables that are not affected by any confounder.

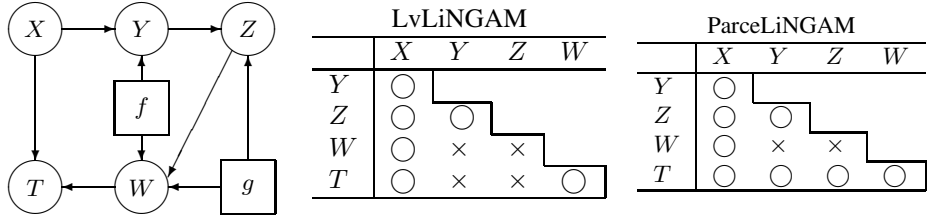
\begin{figure}
\input{fig032}
\caption{\label{fig12}Suppose we have five variables $X,Y,Z,W,T$ are related by $Y=aX+f$, $Z=bY+g$, $W=cZ+f+g$, and $T=cX+dW$
for some $a,b,c,d\in {\mathbb R}$,
and that the confounder $f$ and $g$ affect $Y,W$ and $Z,W$, respectively.
The table on the right shows what pair of variables the order can be recovered for LvLiNGAM and ParceLiNGAM.
For example, in LvLiNGAM, the order between $Y, Z$ can be recovered (denoted as ``$\bigcirc$") even if $Y$ is affected by $f$.
}
\end{figure}

For the details of LvLiNGAM and ParceLiNGAM, see Appendix B.

These methods require us either to know a priori that the same type of confounding always exists or to spend exponential time with $p$ (the number of variables) {\it even when no confounder exists}.

\section{Extended LiNGAM}
In this section, we propose to quantify the amount of confounding in terms of the noises $e_1,\ldots, e_p$, and 
extend the LiNGAM in the framework. 

\subsection{Quantification of Confounding}

We define the amount of confounding by
\begin{equation}\label{eq12}
K(e_1,\ldots,e_p):=E[\log\frac{P(e_1,\ldots,e_p)}{P(e_1)\cdots P(e_p)}]
\end{equation}
when the variables $X_1,\ldots,X_p$ are related by
$$\left\{
\begin{array}{lll}
X_1&=&e_1\\
X_2&=&b_{2,1}X_1+e_2\\
\multicolumn{1}{c}{\vdots}&\multicolumn{1}{c}{\vdots}&\multicolumn{1}{c}{\vdots}\\
X_p&=&b_{p,1}X_1+\cdots+b_{p,p-1}X_{p-1}+e_p
\end{array}\right.$$
for some $b_{i,j}\in {\mathbb R}$, $i=2,\ldots,p-1$, $j=1,\ldots,i-1$.
We interpret confounding as the divergence from the independence among the noises $e_1,\ldots.e_p$,

If there is no confunding, we have $P(e_1,\ldots,e_p)=P(e_1)\cdots P(e_p)$ and 
$K(e_1,\ldots,e_p)=0$, which the original LiNGAM assumes for some order among $X_1,\ldots,X_p$.

We define the
mutual information (MI) between $X,Y$
\begin{eqnarray*}
I(X,Y):=E[\log\frac{P(X,Y)}{P(X)P(Y)}]\ ,
\end{eqnarray*}
which takes non-negative values and satisfies 
\begin{eqnarray}\label{mi}
I(X,Y)=0 \Longleftrightarrow X\ci Y\ ,
\end{eqnarray}
which is similar to (\ref{hsic}).
Then, the confounding can be expressed by the sum of the mutual information values as below:
\begin{eqnarray}\label{eq21}
&&K(e_1,\ldots,e_p)\nonumber\\
&=&E[\log\frac{P(e_1,\ldots,e_p)}{P(e_1)P(e_2,\ldots, e_p)}]+E[\log\frac{P(e_2,\ldots,e_p)}{P(e_2)\cdots P(e_3,\ldots,e_p)}]
+\cdots+E[\log\frac{P(e_{p-1},e_p)}{P(e_{p-1})P(e_p)}]\nonumber
\\
&=&
I(e_1,\{e_2,\ldots,e_p\})+I(e_2,\{e_3,\ldots,e_p\})+\cdots+
I(e_{p-1},e_p)
\end{eqnarray}
Moreover, since mutual information takes non-negative values, we have the equivalences
\begin{eqnarray*}
{\rm no\ confounding}& \Longleftrightarrow &K(e_1,\ldots,e_p)=0\\ &\Longleftrightarrow &I(e_1,\{e_2,\ldots,e_p\})\cdots =I(e_{p-1},e_p)=0 \\&\Longleftrightarrow &e_1,\ldots,e_p\ {\rm are\ independent}
\end{eqnarray*}

One might wonder why we do not minimize 
$$HSIC(e_1,\{e_2,\ldots,e_p\})+
HSIC(e_2,\{e_3,\ldots,e_p\})+\cdots+
HSIC(e_{p-1},e_p)
$$
instead of minimizing (\ref{eq21}). However, each term $HSIC(e_i,\{e_{i+1},\ldots,e_p\})$ has a different deviation and the sum over $i=1,\ldots,p-1$
does not provide any information about confounding if $p\geq 3$.

The additive noise model assumes that exactly one of $K(e_1,\ldots,e_p)$ is zero among the $p!$ orders.
The extended framework assumes that exactly one order of $p$ variables should minimize $K(e_1,\ldots,e_p)$ rather than diminish it.

We compare the identifiability conditions of the original and extended LiNGAM. 
The original LiNGAM requires that $e_1\ci e_2$ and $e_1'\ci e_2'$
 should not happen at the same time. More precisely, it assumes the restrictive condition
$$I(e_1,e_2)=0\ ,\ I(e_1',e_2')>0\hspace{5mm}{\rm or}\hspace{5mm} I(e_1,e_2)>0\ ,\ I(e_1',e_2')=0\ .$$
Note that $I(e_1,e_2)=I(e_1',e_2')=0$ is equivalent to that both of $e_1,e_2$ are Gaussian
(Proposition \ref{prop2}), In this case, order identification is impossible for the original and extended LiNGAM. 
On the other hand, the extended LiNGAM  only requires
$$I(e_1,e_2)\not=I(e_1',e_2')\ .$$
For the $p$ variable case, the original LiNGAM requires $K(e_1,\ldots,e_p)=0$ for exactly  one order while 
the extended requires no tie-breaking occurs for the minimization of $K(e_1,\ldots,e_p)$, which does not seeme to be any constraint.

\begin{figure}
\input{fig033}
\caption{
\label{fig003} There are six paths corresponding to the six orders. We compare the sum of the distances from the top $\{x^n,y^n,z^n\}$ to 
the bottom $\{\}$ according to each path (order). }
\end{figure}
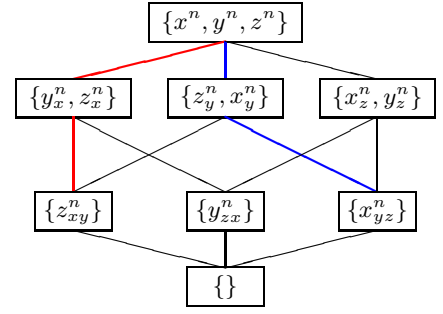

\subsection{Compution of the Order that minimizes the confounding}

This section considers minimizing MI values' sum (\ref{eq21}). To this end, we prepare an ordered graph to obtain the shortest path.
Given $x^n,y^n,z^n$, we can compare the six path as in Figure~\ref{fig003} for variables $X,Y,Z$ ($p=3$).
For example, if we compare orders $X\rightarrow Y\rightarrow Z$ and $Y\rightarrow Z\rightarrow X$, we 
compute the sums over the paths in red and blue. 
Because the residues such as $x^n, y^n_z$ are data rather than variables, we denote the estimated mutual information $I(\cdot,\cdot)$ and Kullback-Leibler (KL) divergence $K(\cdots)$ as $I_n(\cdot,\cdot)$ and $K_n(\cdots)$, respectively.
Suppose we have DATA=$\{x^n,y^n,z^n\}$ as input. 
Then, we can compute the residues as in Figure~\ref{fig003} and an MI estimate value for each of the twelve edges (we assume $I_n(\{x^n_{yz}\},\{\})=I_n(\{y^n_{zx}\},\{\})=I_n(\{z^n_{xy}\},\{\})=0$).

We regard the MI estimates as distances.
Then, for each node $v$, we can compute the length $d(v)$ of the path from the top $\{X, Y, Z\}$ to $v$ and the sum of the distances of the edges along the path.
If multiple paths exist to a node, we choose the shortest path and store it in the node.
For example, for the path $\{x^n,y^n,z^n\}\rightarrow \{y^n_x,z^n_x\}\rightarrow \{z^n_{xy}\}\rightarrow \{\}$,
the sum of the distances is 
\begin{eqnarray*}
I_n(x^n,\{y_x^n,z_x^n\})+I_n(y_x^n,z_{xy}^n)+0=I_n(x^n,\{y_x^n,z_{xy}^n\})+I_n(y_x^n,z_{xy}^n)
=K_n(x^n,y_x^n,z_{xy}^n)\ ,
\end{eqnarray*}
which is the estimated KL divergence of $e_1,e_2,e_3$ such that $X=e_1,Y=aX+e_2,Z=bX+cY+e_3$ for some constants $a,b,c$.
We aim to find the shortest path from the top $\{x^n, y^n, z^n\}$ to the bottom $\{\}$.

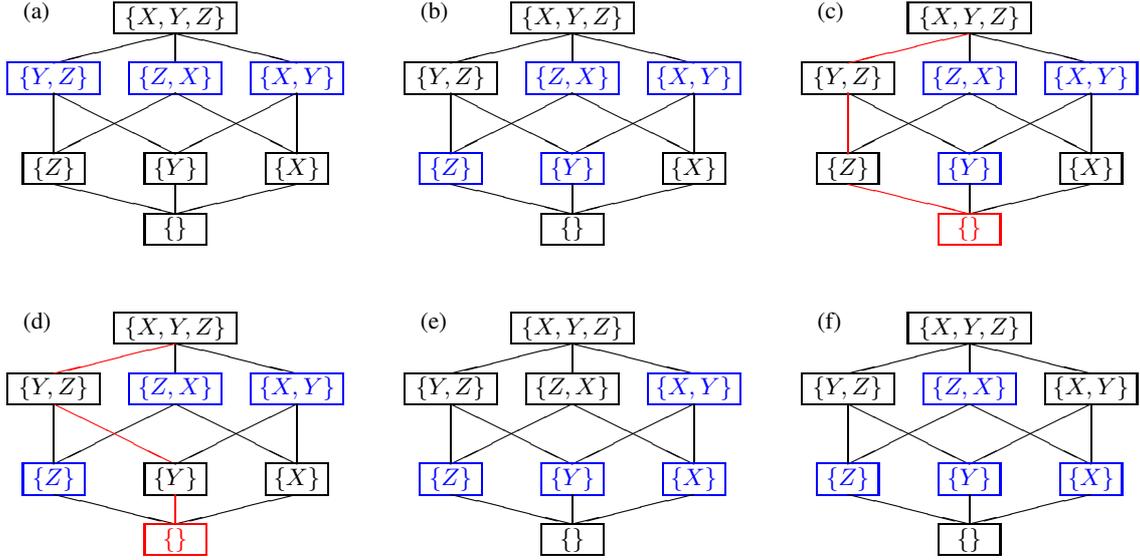
\begin{figure}
\input{figure007d}
\vspace{1em}
\caption{\label{fig03}
The ordered graph consists of the subsets of $V$, where the blue rectangles and red paths are the opened nodes and solutions, respectively.
}
\end{figure}

In Figure \ref{fig03},
we first compute the lengths of the edges from the top $\{X,Y,Z\}$ to $\{Y,Z\},\{Z,X\},\{X,Y\}$:
\begin{eqnarray*}
d(\{Y,Z\})&:=&I_n(x^n,\{y_x^n,z_x^n\})\ ,\\
d(\{Z,X\})&:=&I_n(y^n,\{z_y^n,x_y^n\})\ ,\ {\rm and}\\
d(\{X,Y\})&:=&I_n(z^n,\{x_z^n,y_z^n\})\ .
\end{eqnarray*}
We close the top node $\{X,Y,Z\}$ and open $\{Y,Z\}, \{Z,X\}, \{X,Y\}$ (Figure \ref{fig03}(a)). Suppose that $d(\{Y, Z\})$ is the smallest among the three nodes. 
Then, we compute $I_n(y_x^n,z_{xy}^n)$ and $I_n(z_{x}^n,y_{zx}^n)$ and obtain 
\begin{equation}\label{eq17}
d(\{Z\}):=d(\{Y,Z\})+I_n(y_x^n,z_{xy}^n)
\end{equation}
and
$d(\{Y\}):=d(\{Y,Z\})+I_n(z_{x}^n,y_{zx}^n)$,
respectively. We close $\{Y,Z\}$ and open $\{Z\}$ and $\{Y\}$
(Figure \ref{fig03} (b)).

If $d(\{Z\}\})$ is the smallest in Figure \ref{fig03} (b), then $X\rightarrow Y\rightarrow Z$ is the shortest path 
(Figure \ref{fig03} (c));  if $d(\{Y\})$ is the smallest in Figure \ref{fig03} (b), then $X\rightarrow Z\rightarrow Y$ is the shortest path (Figure \ref{fig03} (d)).
On the other hand, if $d(\{Z,X\})$ is the smallest in Figure \ref{fig03} (b), 
we compute $I_n(z_y^n,x_{yz}^n)$ and $I_n(x_{y}^n,z_{xy}^n)$, and we obtain 
$d(\{X\}):=d(\{Z,X\})+I_n(z_y^n,x_{yz}^n)$
and
\begin{equation}\label{eq18}
d(\{Z\}):=d(\{Z,X\})+I_n(x_{y}^n,z_{xy}^n)\ .
\end{equation}
We close $\{Z,X\}$ and open $\{X\}$ and $\{Z\}$.
However, the values of (\ref{eq17}) and (\ref{eq18}) conflict; thus, we replace (\ref{eq17}) by (\ref{eq18}) if (\ref{eq18}) is smaller (Figure \ref{fig03} (e)).
Finally, if $d(\{X,Y\})$ is the smallest in Figure~\ref{fig03} (b),
we obtain the state as depicted in Figure~\ref{fig03} (f), in which the values of $d(\{Y\})$ conflict, and the shorter path is chosen from $\{X,Y,Z\}$ to $\{Y\}$.

We continue this procedure to obtain the distance $d(\{\})$ and the shortest path
from the top $\{X,Y,Z\}$ to the bottom $\{\}$.
The same procedure can be applied to any number $p$ of variables, not just three.

The procedure (LiNGAM-MMI) is summarized below as Algorithm \ref{algo1}  with input DATA and output SHORTEST\_PATH.
Let TOP and BOTTOM be the top and bottom nodes, and we define $append((u_1,\cdots,u_s), u_{s+1}):=(u_1,\cdots,u_s,u_{s+1})$
for the nodes $u_1,u_2,\cdots,u_{s+1}$.

\begin{algo}\rm \label{algo1}
Let $OPEN:=\{TOP\}$, CLOSE:=$\{\}$, ${\rm path}({\rm TOP}):=()$, $r({\rm TOP}):=$DATA, and repeat:
\begin{enumerate}
\item Suppose $d(v)$ is the smallest among $v\in $ OPEN and that $v_1,\ldots,v_m$ are connected to the $v$.
Then, move $v\in $ OPEN to CLOSE;
\item For each $i=1,\cdots,m$:
\begin{enumerate}
\item If $v_i \not\in$ OPEN, compute the residue $r(v_i)$ of $v_i$ from $r(v)$;
\item Compute the estimated MI $mi$ via $r(v)$ and $r(v_i)$.
\item If either $v_i \not\in$ OPEN  or \{$v_i \in$ OPEN, and $d(v)+mi<d(v_i)$\}, then $d(v_i)=d(v)+mi$ and 
${\rm path}(v_i)=append({\rm path}(v),v_i)$
\item join $v_i$ to OPEN if $v_i\notin$ OPEN for $j=1,\dots,m$.
\end{enumerate}
\item If BOTTOM $\in$ OPEN, SHORTEST\_PATH = $append({\rm path}(v),\{\})$ and terminate.
\end{enumerate}
\end{algo}

Note that Algorithm \ref{algo1} does not compute the residues and MI estimates at the beginning; instead, it calculates each step by step when necessary to reduce the computational complexity. 
In addition, the SHORTEST\_PATH is expressed by a sequence of nodes such as $(\{X,Y,Z\},\{Y,Z\},\{Z\},\{\})$ rather than variables separated by arrows, as in $X\rightarrow Y\rightarrow Z$.

\begin{teiri}\rm
Algorithm \ref{algo1} computes the causal order of random variables that minimizes the estimated KL divergence of the corresponding noise set.
\end{teiri}

Then, one might think that the proposed procedure takes an exponential time with the number $p$ of variables.
In fact, for the worst case, we cannot avoid such computation that LvLiNGAM \citep{entner13} and ParceLiNGAM \citep{tashiro14} require.
However, we have a significant merit over them (Theorem 2 below).

Let $I_n(\cdot,\cdot)$ be an MI estimator  such that $I_n(x^n,y^n)=0 \Longleftrightarrow X\ci Y$ with probability one for the $\{(x_i,y_i)\}_{i=1}^n$, independent realizations of $(X,Y)$ for $x^n=(x_1,\ldots,x_n)$ and $y^n=(y_1,\ldots,y_n)$.
Such an estimator $I_n(\cdot,\cdot)$ is said to be consistent.
\begin{teiri}\rm\label{teiri2}
Suppose we estimate the MI between variables in Algorithm \ref{algo1} via a consistent $I_n(\cdot,\cdot)$,
When no confounding exists, the original and proposed LiNGAM take the same number 
$$\displaystyle \frac{p(p+1)}{2}-1$$
of computing the MI estimates $I_n(\cdot,\cdot)$ with probability one.
\end{teiri}
Proof. Suppose that there exists an order among the $p$ variables such that the total distance
$$I(e_1,\{e_2,\ldots,e_p\})+I(e_2,\{e_3,\ldots,e_p\})+\cdots+I(e_{p-1},e_p)$$
is zero, which is equivalent to
$$I(e_1,\{e_2,\ldots,e_p\})=I(e_2,\{e_3,\ldots,e_p\})=\cdots=I(e_{p-1},e_p)=0\ .$$
Then, we have the residues $e_1^n, \ldots, e_p^n \in {\mathbb R}^n$ of the associated order such that each of
$$I_n(e_1^n,\{e_2^n,\ldots,e_p^n\})\ ,\ I(e_2^n,\{e_3^n,\ldots,e_p^n\})\ ,\ \cdots\ ,\ I(e_{p-1}^n,e_p^n)$$
almost surely converges to zero. Then, among the OPEN variables of Algorithm \ref{algo1}, the one with 
$$\sum_{j=1}^kI_n(e_j^n,\{e_{j+1}^n,\ldots,e_p^n\})=0$$ is chosen for iteration $k=1,\ldots,p$.
\hfill {$\blacksquare$}

For example, suppose that in Figure \ref{fig03}, $X\rightarrow Y\rightarrow Z$ is the correct order and that no confounding exists. Then, we first compute and compare $d(\{Y,Z\})$, $d(\{Z,X\})$, and $d(\{X,Y\})$ to find that $d(\{Y,Z\})=0$ is the smallest. Then, we compute $d(\{Y\})$ and $d(\{Z\})$ and compare with $d(\{Z,X\})$ and $d(\{X,Y\})$ to find that $d(\{Z\})=0$ is the smallets. Thus, we computed 3+2=5 MI estimates for $p=3$. On the other hand, the original LiNGAM procedure in Section 2.3 requires five estimates of the MI or HSIC estimates before obtaining the order.

The proposed procedure finds the best order among the $p!$ candidates. On the other hand, the original LiNGAM procedure in Section 2.3 searches the solution in a greedy (topdown) manner and successfully finds the correct order when no confounding exists. However, even when no confounding exists, the estimates of HSIC and MI show positive values if starting from data, which means that no one can see the border between whether confounding exists. Thus, we cannot assume that no confounding exists in general situations. 

In reality, if $p$ is large, assuming that no confounding exists among the $p$ variables is hopeless.

Moreover, if we follow the greedy search in Section 2.3 while the procedure is efficient, we will face fatal situations.
For example, suppose we wish to identify the order among $X,Y,Z$ from $x^n,y^n,z^n$, and that $I_n(x^n,\{y^n_x,z^n_x\})$ is smaller than the other MI estimates. Then, we cannot decide that $X$ is the top variable because both of $I_n(y^n_x,z^n_{xy}\})$ and $I_n(y^n_{xz},z^n_{x}\})$ may be large. In general, $e_1\ci \{e_2,e_3\}$ does not mean $e_2\ci e_3$.

\subsection{MI Estimation}
Many MI estimation procedures
exist, such as \citep{Kraskov}, \citep{mine}.
The precision of the proposed procedure depends on the choice of the MI estimation method.
Among them, we choose copula entropy \citep{copula}, which shows the best performance in our preliminary experiments.

Copulas \citep{Sklar} provide a framework to separate the dependence structure from the marginal distributions. 
Let
$$
F_X(x):=\int_{-\infty}^x f_X(x)dx\ ,\ 
F_Y(y):=\int_{-\infty}^y f_Y(y)dy\ ,\ {\rm and}\ 
F_{XY}(x,y):=\int_{-\infty}^x\int_{-\infty}^y f_{XY}(x,y)dx$$
with $x,y\in {\mathbb R}$
be the distribution functions of variables $X$, $Y$, and $(X,Y)$.
\cite{Sklar} proves that the existence of the function $C$
$$F_{XY}(x,y)=C(F_X(x),F_Y(y))$$
with $x,y\in {\mathbb R}$ for any variables $X,Y$.
Then, for $u_X=F_X(x)$ and $u_Y=F_Y(y)$, we have
$$\frac{du_X}{dx}=f_X(x)\ ,\ 
\frac{du_Y}{dy}=f_Y(y)
$$
and 
$$c(u_X,u_Y):=\frac{\partial^2C(u_X.u_Y)}{\partial u_X\partial u_Y}=\frac{f_{XY}(x,y)}{f_{X}(x)f_{Y}(y)}\ .$$
Then, we have two expressions for the MI:
\begin{equation}\label{eq1112}
\int_{-\infty}^\infty\int_{-\infty}^\infty f_{XY}(x,y)\log \frac{f_{XY}(x,y)}{f_{X}(y)f_{Y}(y)}dxdy=
\int_{0}^1\int_{0}^1 c(u_X,u_Y)\log c(u_X,u_Y)du_Xdu_Y\ .
\end{equation}
We estimate the right-hand side (negated copula entropy) of (\ref{eq1112}) rather than the left, which we claim is much easier to compute.
Even when $X, Y$ are in high dimensions, the estimation of the right-hand side is always in the two-dimensional space.
For the details, see \cite{ma2021}.
For $n$ data points, we take the following steps:
\begin{enumerate}
\item Computing 
$$u_{X}^{i}=\frac{1}{n} \sum_{j=1}^{n} 1_{\left\{x^{j} \leq x^{i}\right\}} \quad {\rm and} \quad u_{Y}^{i}=\frac{1}{n} \sum_{j=1}^{n} 1_{\left\{y^{j} \leq y^{i}\right\}}\ ,$$
for $i=1, \ldots, n$ from samples $\left\{x^{i}\right\}_{i=1}^{n},\left\{y^{i}\right\}_{i=1}^{n}$
\item Using whatever entropy estimation method to estimate the entropy of samples $\left\{u^{i}\right\}_{i=1}^{n}$ where $u^{i}=\left\{(u_{X}^{i}, u_{Y}^{i})\right\}$ (concatenation and yields variables in two dimensions).
\end{enumerate}
In the second step, we use \cite{Kraskov} based on the KNN method. 

One might think that an independence test based on the HSIC (Hilbert Schmidt Information Criterion \citep{hsic})
achieves better performance than the one based on MI.
In the case $p=2$, this may hold. However, the independence test requires us to execute 
a greedy search for $p\geq 3$, which may result in poor performance unless $n$ is infinitely large and no confounder exists. In the next section, we can examine the performances of the original LiNGAM using the HSIC and the proposed LiNGAM.

\section{Experiments}

This section presents the experimental results.

We compare the proposed method to variants of the LiNGAM as specified in Table \ref{tabtab1}, focusing on causal order identification.
For the implementation, we utilize Python packages: \verb@copent@ for copula entropy \citep{ma2021}, and \verb@lingam@, \verb@gCastle@ \citep{gcastle}. 
When confounders are present, we also substitute RESIT with ParceLiNGAM \citep{tashiro14}. 

For assessing causal order, we employ two criteria for performance comparison: Criterion A counts an error when the whole order of $p$ variables is incorrect. Criterion B counts the pairwise errors and divides the count by $p(p-1)/2$. 
Criterion A evaluates the error rate more severely, particularly when $p$ is large.

\begin{table}
\caption{\label{tabtab1} The procedures used in the experiments and Figures 7-10 }
\begin{center}
\begin{tabular}{l|l|l}
\hline
Pairwise&DirectLiNGAM using pairwise likelihood& \cite{apo13}\\
Kernel&kernel-based DirectLiNGAM &\cite{shimizu11}\\
HSIC&DirectLiNGAM using HSIC&\cite{apo13}\\
ICA&ICA-LiNGAM&\cite{shimizu06}\\
MMI&LiNGAM-MMI using copula entropy&Proposed\\
\hline
\end{tabular}
\end{center}
\end{table}

\clearpage
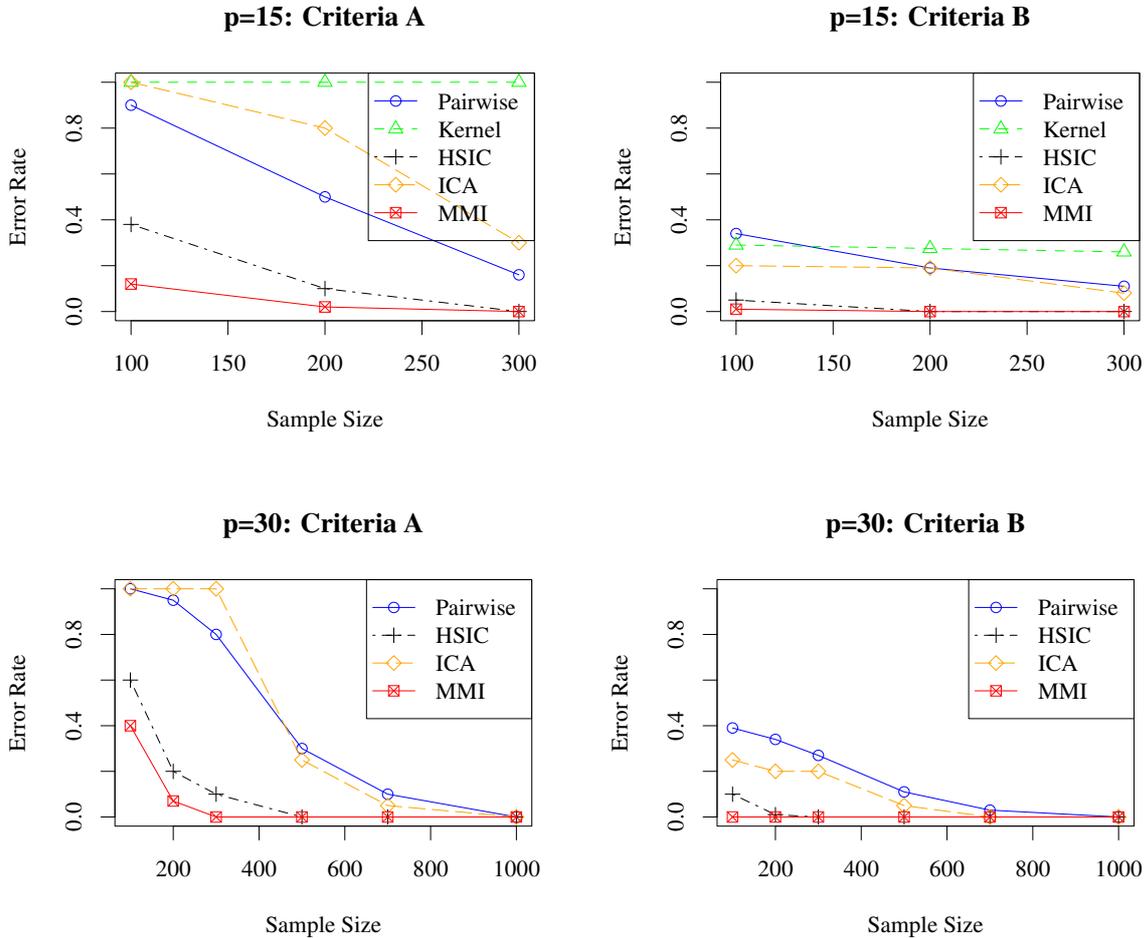
\begin{figure}
    \centering
    \resizebox{\columnwidth}{!}{ 

    \input{result_11}
\input{result_21}}
\resizebox{\columnwidth}{!}{ 
\input{result_31}
\input{result_41}
}
    \caption{\textbf{No Confounder}: Error ratio of Criteria A and B with sample size $n$ (the lower, the better). We observe that the proposed LiNGAM-MMI performs better than the conventional methods.}
    \label{1530no}
\end{figure}

\begin{figure}
    \centering
    \resizebox{\columnwidth}{!}{ 
    \input{result_51}
\input{result_61}}
\resizebox{\columnwidth}{!}{ 
\input{result_71}
\input{result_81}
}
    \caption{\textbf{With Confounder}: Error ratio of Criteria A and B with sample size $n$ (the lower, the better). We observe that the proposed LiNGAM-MMI performs better than the conventional methods.}
    \label{1530with}
\end{figure}
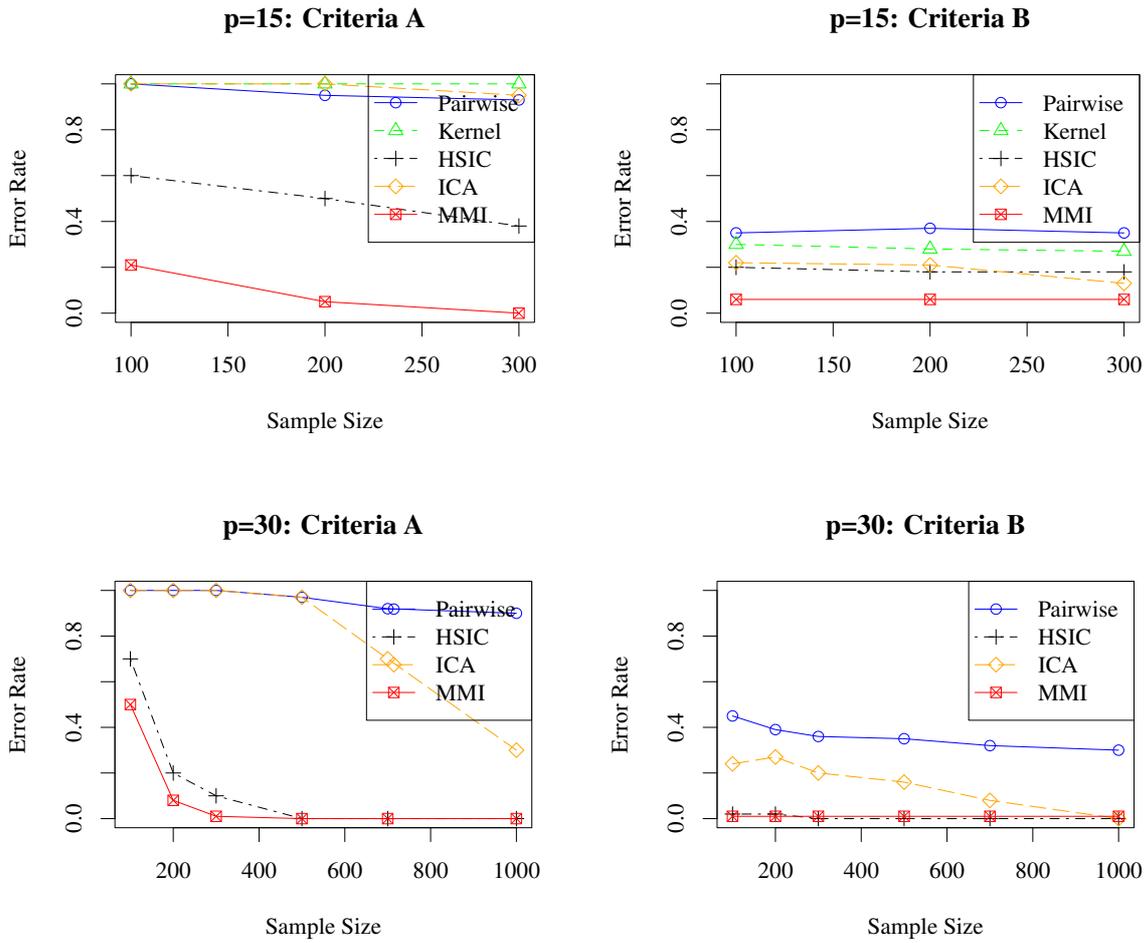

\begin{figure}
    \centering
    \resizebox{\columnwidth}{!}{ 

    \input{result_91}
\input{result_101}}
\caption{\label{15no}\textbf{No Confounder}: Error ratio of Criterion A and B with sample size $n$ (the lower, the better) when copula entropy was applied to the proposed and existing methods.}
\end{figure}
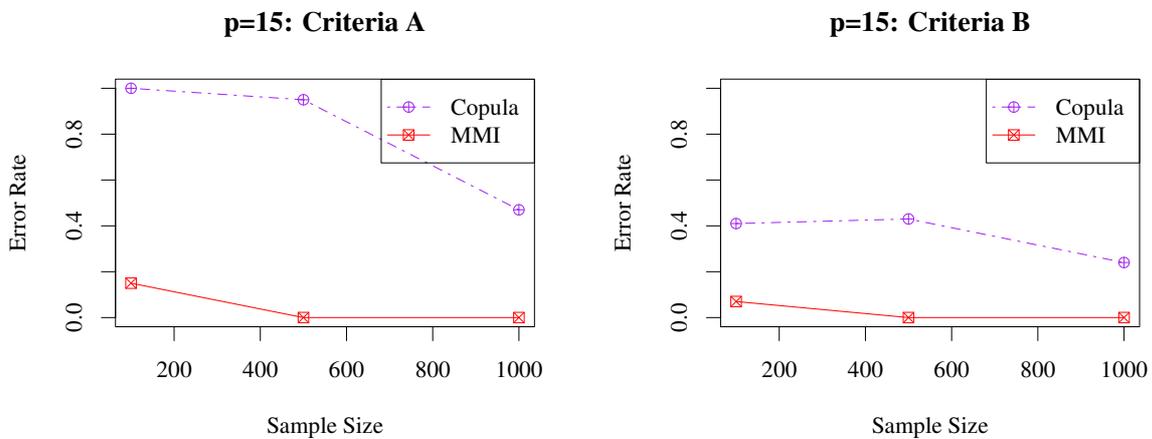

\begin{figure}
    \centering
\resizebox{\columnwidth}{!}{ 
\input{result_1111}
\input{result_121}
}
\caption{\label{15co}\textbf{With Confounder}: Error ratio of Criterion A and B with sample size $n$ (the lower, the better) when copula entropy was applied to the proposed and existing methods.}
\end{figure}
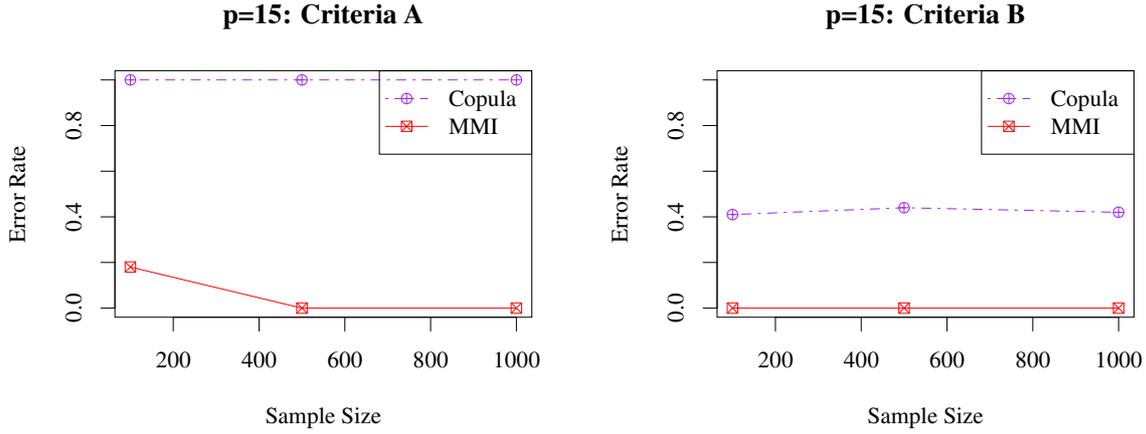
\subsection{Simulations}

We estimate the causal order for $p=15,30$. The true causal order is $X_0\rightarrow X_1,\dots,\rightarrow X_{p-1}\rightarrow X_p$ with $X_i=X_{i-1}+e_i,i=1,\dots,p$. 
When considering confounders, we assume they are $f_k\sim 2U(0,1)-1, k=1,\dots,8$ at different positions. For $p=30$, we set the confounder with $f_1$ on $\{X_1,X_2\}$, $f_2$ on $\{X_5,X_6\}$, $f_3$ on $\{X_8,X_9\}$, $f_4$ on $\{X_{10},X_{11}\}$, $f_5$ on $\{X_{13},X_{14}\}$, $f_6$ on $\{X_{15},X_{16}\}$, $f_7$ on $\{X_{20},X_{21}\}$, $f_8$ on $\{X_{25},X_{26}\}$, respectively. For  $p=15$, we set the confounder as $f_1$ on $\{X_2,X_3\}$, $f_2$ on $\{X_5,X_6\}$, $f_3$ on $\{X_8,X_9\}$, $f_4$ on $\{X_{12},X_{13}\}$, respectively. 

We execute the existing and proposed procedures  for 
$n=100, 200, 300, 500, 700, 1000$, $p=30$ and $n=100, 200, 300$, $p=15$, we compare the performances of Criteria A and B. 
For the whole procedure, we take the arithmetic average over 50 trials. The results in Figure~\ref{1530no} and Figure~\ref{1530with} show that our proposed method performs best whether the sample size $n$ is small or large. 
We observe that the DirectLiNGAM with HSIC shows competitive performances with the LiNGAM-MMI only when confounding is absent and the sample size is large. In particular, even when no confounding exists, the DirectLiNGAM with HSIC often failed for $p=30$ and a small sample size. The reason seems to be that when the sample size is small, the situation is far from confounding-free.

In addition, we compare our LiNGAM-MMI with the DirectLiNGAM with copula entropy (the same MI measure as our method). We show the results in Figure~\ref{15no} and Figure~\ref{15co}, which suggests that our method still performs better even though the DirectLiNGAM uses copula entropy, which suggests that the greedy search fails to give correct orders often.

\subsection{Real data}
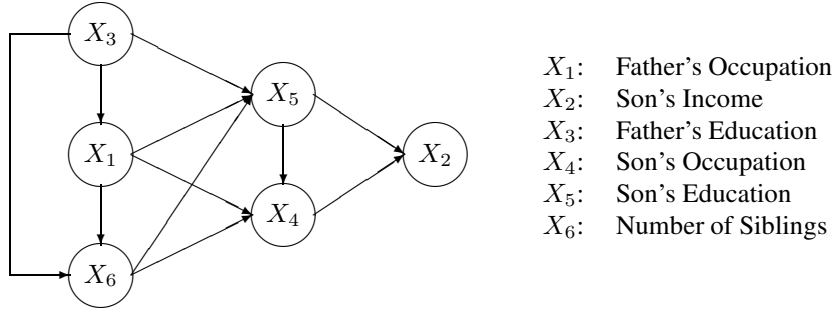
\begin{figure}[!htbp]
    \centering
    {
	\input{fig_new_new}
    }
    \caption{
    Causal relations by LiNGAM-MMI in GSS dataset. Red arrows: correct; dashed ones: wrong; other: appropriate.}
    \label{gss}
\end{figure}

We examined the order of the variables in
the General Social Survey data set\footnote{http://www.norc.org/GSS+Website/
}, taken from a sociological data  repository, used in \cite{shimizu11, RCD}. See the actual causal structure in Figure \ref{gss}. In general, we do not know the true order among variables. However, in this example, from the meanings of the six variables, we may assume that 
$$X_3\rightarrow X_1 \rightarrow X_6\rightarrow 
X_5\rightarrow X_4\rightarrow X_2\ .$$
We observed that the LiNGAM-MMI obtains the same order.
We cannot claim that the LiNGAM-MMI always shows better performances only from this example but rather illustrate how to apply it to real data. 
For seeking the causal structure besides the order, we need to obtain the parent sets of each variable using structure learning such as the PC algorithm \citep{pc}.

\section{Concluding Remarks}

This paper extends LiNGAM itself without explicitly searching for variables influenced by confounding. Specifically, it proposes a method (LiNGAM-MMI) that quantifies the magnitude of confounding using KL divergence and determines the order of variables to minimize it. This approach formulates the problem using the shortest path problem and successfully finds globally optimal solutions efficiently.
The proposed LiNGAM-MMI completes processing in the same amount of time as the existing LiNGAM when there is no confounding, as it does not address confounding.
The experiments support the merits of the LiNGAM-MMI. 

Although this paper has extended the estimation of MI using copulas, applying estimators with higher accuracy is expected to improve performance further.

\section*{Appendix A: Proof of $z_{xy}^n=z_{yx}^n$}

From definitions of $v(\cdot)$ and $c(\cdot,\cdot)$, we have 
\begin{eqnarray*}
c(y_x^n,z_x^n)&=&c(y^n-\frac{c(x^n,y^n)}{v(x^n)}x^n, z^n-\frac{c(x^n,z^n)}{v(x^n)}x^n)=c(y^n,z^n)-\frac{c(x^n,y^n)c(x^n,z^n)}{v(x^n)}\\
v(y_x^n)&=&v(y^n-\frac{c(x^n,y^n)}{v(x^n)}x^n)=v(y^n)+\frac{c(x^n,y^n)^2}{v(x^n)}-2\frac{c(x^n,y^n)^2}{v(x^n)}=v(y^n)-\frac{c(x^n,y^n)^2}{v(x^n)}\\
\frac{c(y_x^n,z_x^n)}{v(y_x^n)}&=&\frac{v(x^n)c(y^n,z^n)-c(x^n,y^n)c(x^n,z^n)}{v(x^n)v(y^n)-c(x^n,y^n)^2}\ .
\end{eqnarray*}

Thus, we can express (\ref{eq116}) as 
\begin{eqnarray*}\label{eq104}
z_{xy}^n&=&z^n-\frac{c(x^n,z^n)}{v(x^n)}x^n -\frac{v(x^n)c(y^n,z^n)-c(x^n,y^n)c(x^n,z^n)}{v(x^n)v(y^n)-c(x^n,y^n)^2}\{y^n-\frac{c(x^n,y^n)}{v(x^n)}x^n \}
\nonumber\\
&=&z^n-\frac{c(x^n,z^n)\{v(x^n)v(y^n)-c(x^n,y^n)^2\}-c(x^n,y^n)\{v(x^n)c(y^n,z^n)-c(x^n,y^n)c(x^n,z^n)\}}{v(x^n)\{v(x^n)v(y^n)-c(x^n,y^n)^2\}}x^n\nonumber\\
&&-
\frac{v(x^n)c(y^n,z^n)-c(x^n,y^n)c(x^n,z^n)}{v(x^n)v(y^n)-c(x^n,y^n)^2}y^n\\
&=&z^n-\frac{v(y^n)c(x^n,z^n)-c(x^n,y^n)c(y^n,z^n)}{v(x^n)v(y^n)-c(x^n,y^n)^2}x^n-
\frac{v(x^n)c(y^n,z^n)-c(x^n,y^n)c(x^n,z^n)}{v(x^n)v(y^n)-c(x^n,y^n)^2}y^n
\end{eqnarray*}
which coincides with (\ref{eq112}) since they are symmetric between $x^n$ and $y^n$.

\section*{Appendix B: LvLiNGAM and ParceLiNGAM}

LvLiNGAM \citep{entner13} 
considers $e_{12}, e_{13}$, etc., as well as $e_1, e_2, e_3$ as noise variables, and obtains the orders of variable pairs that are not affected by any confounder such as $e_{12}, e_{13}$ (Figure \ref{fig09}), and estimates the order of the whole variables by combining those pairwise orders.

ParceLiNGAM \citep{tashiro14} 
divides the variable set $V$ into upper, middle, and lower variable sets 
(we denote them as $U, M, L$, respectively, such that $V = U \cup M \cup L$)
by top-down and bottom-up causal searches, where the upper and lower variable sets are the maximal subsets that contain no confounder but the top and bottom variables, respectively. 
Let $F_U(u)$ and $F_{L}(l)$ be the Fisher estimations between $u\in U$ and its lower variables 
and between $l\in L$ and its upper variables, respectively. 
Then, the following quantity can be evaluated for $V$:
\[
F := \sum_{u\in U} F_{U}(u) + \sum_{l\in L} F_{L}(l)\ .
\]
For each subset $S$ of $V$, ParceLiNGAM estimates $U, M, L$ and computes $F$ for $S$ rather than $V$.
Then, the order between variables $v, v' \in V$ can be determined based on 
$S$ that determines the order between $v$ and $v'$ and maximizes $F$.

However, ParceLiNGAM has several drawbacks. First of all, it is possible that 
the order cannot be determined based on any $S \subseteq V$ for some $v, v' \in V$.
For example, if $V$ consists of $X, Y$ and a confounder exist for them, ParceLiNGAM does not work at all  (Figure \ref{fig12}).
Secondly, the evaluation is for each pair of variables rather than for each path from the top to the bottom,
so we are not sure that the estimation is correct.
Moreover, Fisher estimation between variable $r$ and variable set $R$ evaluates the sum of pairwise independences between 
$r$ and each element in $R$, and using it is not appropriate unless the variables in $R$ are mutually independent.
Finally, the computation is huge: $2^p$ subsets should be considered for $p$ variables,
so that ParceLiNGAM can be used only when $p$ is small.

\bibliography{2018-2-7}

\end{document}

%% file: fig031.tex
\setlength\unitlength{0.6mm}
\small
\begin{center}
\begin{picture}(180,45)(0,-5)
\put(10,10){\makebox(0,0)[c]{$X_1$}}
\put(10,10){\circle{10}}
\put(40,10){\makebox(0,0)[c]{$X_2$}}
\put(40,10){\circle{10}}
\put(70,10){\makebox(0,0)[c]{$X_3$}}
\put(70,10){\circle{10}}
\put(10,40){\makebox(0,0)[c]{$e_1$}}
\put(10,40){\circle{10}}
\put(40,40){\makebox(0,0)[c]{$e_2$}}
\put(40,40){\circle{10}}
\put(70,40){\makebox(0,0)[c]{$e_3$}}
\put(70,40){\circle{10}}
\put(10,35){\vector(0,-1){20}}
\put(40,35){\vector(0,-1){20}}
\put(70,35){\vector(0,-1){20}}
\put(40,-5){\makebox(0,0)[c]{\color{blue}Without Confounding}}

{\thicklines
\put(15,10){\vector(1,0){20}}
\put(45,10){\vector(1,0){20}}
\put(117,10){\vector(1,0){20}}
\put(147,10){\vector(1,0){20}}
}

{\color{red}
\put(125,30){\makebox(0,0)[c]{$e_{12}$}}
\put(125,30){\circle{10}}
\put(155,30){\makebox(0,0)[c]{$e_{23}$}}
\put(155,30){\circle{10}}
\put(140,-5){\makebox(0,0)[c]{$e_{13}$}}
\put(140,-5){\circle{10}}
\put(122,26){\vector(-1,-1){11}}
\put(128,26){\vector(1,-1){11}}
\put(152,26){\vector(-1,-1){11}}
\put(158,26){\vector(1,-1){11}}
\put(135,-5){\vector(-2,1){22}}
\put(145,-5){\vector(2,1){22}}
}
\put(110,10){\makebox(0,0)[c]{$X_1$}}
\put(110,10){\circle{10}}
\put(140,10){\makebox(0,0)[c]{$X_2$}}
\put(140,10){\circle{10}}
\put(170,10){\makebox(0,0)[c]{$X_3$}}
\put(170,10){\circle{10}}
\put(110,40){\makebox(0,0)[c]{$e_1$}}
\put(110,40){\circle{10}}
\put(140,40){\makebox(0,0)[c]{$e_2$}}
\put(140,40){\circle{10}}
\put(170,40){\makebox(0,0)[c]{$e_3$}}
\put(170,40){\circle{10}}
\put(110,35){\vector(0,-1){20}}
\put(140,35){\vector(0,-1){20}}
\put(170,35){\vector(0,-1){20}}
\end{picture}
\end{center}

%% file: ancestral.tex
\setlength\unitlength{0.6mm}
\begin{center}
\begin{picture}(200,35)
\put(10,15){\circle{10}}
\put(30,15){\circle{10}}
\put(50,15){\circle{10}}
\put(10,15){\makebox(0,0)[c]{$1$}}
\put(30,15){\makebox(0,0)[c]{$2$}}
\put(50,15){\makebox(0,0)[c]{$3$}}
{\color{blue}
\put(25,15){\vector(-1,0){10}}
\put(35,15){\vector(1,0){10}}
}
{\color{red}
\put(10,30){\vector(0,-1){10}}
\put(50,30){\vector(0,-1){10}}
\put(10,30){\line(1,0){40}}
}
\put(17,0){(a) Ancestral}

\put(80,15){\circle{10}}
\put(100,15){\circle{10}}
\put(120,15){\circle{10}}
\put(80,15){\makebox(0,0)[c]{$1$}}
\put(100,15){\makebox(0,0)[c]{$2$}}
\put(120,15){\makebox(0,0)[c]{$3$}}
{\color{blue}
\put(85,15){\vector(1,0){10}}
\put(105,15){\vector(1,0){10}}
}
{\color{red}
\put(80,30){\vector(0,-1){10}}
\put(120,30){\vector(0,-1){10}}
\put(80,30){\line(1,0){40}}
}
\put(70,0){(b) Non-Ancestral, Bow-free}

\put(150,15){\circle{10}}
\put(170,15){\circle{10}}
\put(190,15){\circle{10}}
\put(150,15){\makebox(0,0)[c]{$1$}}
\put(170,15){\makebox(0,0)[c]{$2$}}
\put(190,15){\makebox(0,0)[c]{$3$}}
{\color{blue}
\put(155,15){\vector(1,0){10}}
\put(175,15){\vector(1,0){10}}
}
{\color{red}
\put(150,30){\vector(0,-1){10}}
\put(170,30){\vector(0,-1){10}}
\put(150,30){\line(1,0){20}}
}
\put(149,0){(c) Non-Bow-free}

\end{picture}
\end{center}

%% file: fig029.tex
\begin{center}
{\setlength{\unitlength}{0.8mm}
\begin{picture}(150,40)(0,5)
\put(50,10){\color{red}\vector(0,1){20}}
\put(10,10){\color{blue}\vector(1,0){50}}
\put(20,25){\makebox(0,0)[c]{$Y=aX+e_2$}}
\put(30,5){\makebox(0,0)[c]{\color{blue}$X=e_1$}}
\put(55,20){\makebox(0,0)[c]{\color{red}$e_2$}}
\put(10,10){\vector(2,1){40}}
\put(120,10){\color{red}\vector(0,1){20}}
\put(80,10){\color{blue}\vector(1,0){50}}
\put(90,25){\makebox(0,0)[c]{$X=a'Y+e_2'$}}
\put(100,5){\makebox(0,0)[c]{\color{blue}$Y=e_1'$}}
\put(125,20){\makebox(0,0)[c]{\color{red}$e_2'$}}
\put(80,10){\vector(2,1){40}}
\end{picture}
}
\end{center}

%% file: fig030.tex
{\setlength\unitlength{0.5mm}
\small
\begin{center}
\begin{picture}(80,80)(20,20)
\put(40,90){\framebox(40,10){$\{x^n,y^n,z^n\}$}}
\put(60,90){\line(4,-1){40}}
\put(5,70){\framebox(30,10){$\{y^n_x,z^n_x\}$}}
\put(45,70){\framebox(30,10){$\{z_y^n,x_y^n\}$}}
\put(85,70){\framebox(30,10){$\{x_z^n,y_z^n\}$}}
\put(60,70){\line(2,-1){40}}
\put(20,70){\line(2,-1){40}}
\put(100,70){\line(-2,-1){40}}
\put(100,70){\line(0,-1){20}}
\put(10,40){\framebox(20,10){$\{z_{xy}^n\}$}}
\put(50,40){\framebox(20,10){$\{y_{zx}^n\}$}}
\put(90,40){\framebox(20,10){$\{x_{yz}^n\}$}}
\put(20,40){\line(4,-1){40}}
\put(100,40){\line(-4,-1){40}}
\put(60,40){\line(0,-1){10}}
\put(50,20){\framebox(20,10){$\{\}$}}
\thicklines
\put(60,90){\color{red}\line(-4,-1){40}}
\put(60,90){\color{blue}\line(0,-1){10}}
\put(60,70){\color{blue}\line(-2,-1){40}}
\put(20,70){\color{red}\line(0,-1){20}}
\end{picture}
\end{center}
}

%% file: fig032.tex
\begin{center}
\begin{tabular}{llll}
{
\setlength\unitlength{0.7mm}
\small
\begin{picture}(50,30)(0,20)
\put(0,30){\makebox(10,10){$X$}}
\put(5,35){\circle{10}}
\put(20,30){\makebox(10,10){$Y$}}
\put(25,35){\circle{10}}
\put(20,15){\framebox(10,10){$f$}}
\put(40,0){\framebox(10,10){$g$}}
\put(20,0){\makebox(10,10){$W$}}
\put(0,0){\makebox(10,10){$T$}}
\put(5,5){\circle{10}}
\put(20,5){\vector(-1,0){10}}
\put(25,5){\circle{10}}
\put(40,5){\vector(-1,0){10}}
\put(45,10){\vector(0,1){20}}
\put(42,31){\vector(-1,-2){12}}
\put(5,30){\vector(0,-1){20}}
\put(10,35){\vector(1,0){10}}
\put(40,30){\makebox(10,10){$Z$}}
\put(45,35){\circle{10}}
\put(30,35){\vector(1,0){10}}
\put(25,15){\vector(0,-11){5}}
\put(25,25){\vector(0,1){5}}
\end{picture}
}
&
{\begin{tabular}{l|lllll}
\multicolumn{5}{c}{LvLiNGAM}\\
\hline
&$X$
&$Y$
&$Z$
&$W$
\\
\hline
$Y$&\multicolumn{1}{l|}{$\bigcirc$}\\
\cline{3-3}
$Z$&$\bigcirc$&\multicolumn{1}{l|}{$\bigcirc$}\\
\cline{4-4}
$W$&$\bigcirc$&$\times$&\multicolumn{1}{l|}{$\times$}\\
\cline{5-5}
$T$&$\bigcirc$&$\times$&$\times$&\multicolumn{1}{l|}{$\bigcirc$}\\
\hline
\end{tabular}}&
{\small\begin{tabular}{l|lllll}
\multicolumn{5}{c}{ParceLiNGAM}\\
\hline
&$X$
&$Y$
&$Z$
&$W$
\\
\hline
$Y$&\multicolumn{1}{l|}{$\bigcirc$}\\
\cline{3-3}
$Z$&$\bigcirc$&\multicolumn{1}{l|}{$\bigcirc$}\\
\cline{4-4}
$W$&$\bigcirc$&$\times$&\multicolumn{1}{l|}{$\times$}\\
\cline{5-5}
$T$&$\bigcirc$&$\bigcirc$&$\bigcirc$&\multicolumn{1}{l|}{$\bigcirc$}\\
\hline
\end{tabular}}
\end{tabular}
\end{center}

%% file: fig033.tex
\begin{wrapfigure}{r}[10pt]{0.4\textwidth}
{\setlength\unitlength{0.5mm}
\small
\begin{center}
\begin{picture}(80,70)(20,30)
\put(40,90){\framebox(40,10){$\{x^n,y^n,z^n\}$}}
\put(60,90){\thicklines\color{red}\line(-4,-1){40}}
\put(60,90){\thicklines\color{blue}\line(0,-1){10}}
\put(60,90){\line(4,-1){40}}
\put(5,70){\framebox(30,10){$\{y^n_x,z^n_x\}$}}
\put(45,70){\framebox(30,10){$\{z_y^n,x_y^n\}$}}
\put(85,70){\framebox(30,10){$\{x_z^n,y_z^n\}$}}
\put(60,70){\line(-2,-1){40}}
\put(60,70){\thicklines\color{blue}\line(2,-1){40}}
\put(20,70){\line(2,-1){40}}
\put(100,70){\line(-2,-1){40}}
\put(20,70){\thicklines\color{red}\line(0,-1){20}}
\put(100,70){\line(0,-1){20}}
\put(10,40){\framebox(20,10){$\{z_{xy}^n\}$}}
\put(50,40){\framebox(20,10){$\{y_{zx}^n\}$}}
\put(90,40){\framebox(20,10){$\{x_{yz}^n\}$}}
\put(20,40){\line(4,-1){40}}
\put(100,40){\line(-4,-1){40}}
\put(60,40){\line(0,-1){10}}
\put(50,20){\framebox(20,10){$\{\}$}}
\end{picture}
\end{center}}
\end{wrapfigure}
\begin{eqnarray*}
K_n(x^n,y^n_x,z^n_{xy})&=&{\color{red}I_n(x^n,\{y_x^n,z_x^n\})+I_n(y_x^n,z_{xy}^n)}\\
K_n(x^n,z_x^n,y_{xz}^n)&=&I_n(x^n,\{y^n_x,z_x^n\})+I_n(z_{x}^n,y_{xz}^n)\\
K_n(y^n,z^n_y,x^n_{yz})&=&{\color{blue}I_n(y^n,\{z^n_y,x_y^n\})+I_n(z_y^n,x_{yz}^n)}\\
K_n(y^n,z_x^n,y_{xz}^n)&=&I_n(y^n,\{z^n_y,x_y^n\})+I_n(x_{y}^n,z_{xy}^n)\\
K_n(z^n,x^n_z,y^n_{zx})&=&I_n(z^n,\{x^n_z,y_z^n\})+I_n(x_z^n,y_{zx}^n)\\
K_n(z^n,z_x^n,y_{xz}^n)&=&I_n(z^n,\{x^n_z,y_z^n\})+I_n(y_{z}^n,x_{yz}^n)
\end{eqnarray*}
\vspace{1em}

%% file: figure007d.tex
\setlength\unitlength{0.4mm}
\small
\begin{center}
\begin{tabular}{cccc}
\begin{picture}(120,90)(0,20)
\put(10,95){(a)}
\put(40,90){\framebox(40,10){$\{X,Y,Z\}$}}
\put(60,90){\line(-4,-1){40}}
\put(60,90){\line(0,-1){10}}
\put(60,90){\line(4,-1){40}}
\put(5,70){\color{blue}\framebox(30,10){$\{Y,Z\}$}}
\put(45,70){\color{blue}\framebox(30,10){$\{Z,X\}$}}
\put(85,70){\color{blue}\framebox(30,10){$\{X,Y\}$}}
\put(60,70){\line(-2,-1){40}}
\put(60,70){\line(2,-1){40}}
\put(20,70){\line(2,-1){40}}
\put(100,70){\line(-2,-1){40}}
\put(20,70){\line(0,-1){20}}
\put(100,70){\line(0,-1){20}}
\put(10,40){\framebox(20,10){$\{Z\}$}}
\put(50,40){\framebox(20,10){$\{Y\}$}}
\put(90,40){\framebox(20,10){$\{X\}$}}
\put(20,40){\line(4,-1){40}}
\put(100,40){\line(-4,-1){40}}
\put(60,40){\line(0,-1){10}}
\put(50,20){\framebox(20,10){$\{\}$}}
\end{picture}&
\begin{picture}(120,90)(0,20)
\put(10,95){(b)}
\put(40,90){\framebox(40,10){$\{X,Y,Z\}$}}
\put(60,90){\line(-4,-1){40}}
\put(60,90){\line(0,-1){10}}
\put(60,90){\line(4,-1){40}}
\put(5,70){\framebox(30,10){$\{Y,Z\}$}}
\put(45,70){\color{blue}\framebox(30,10){$\{Z,X\}$}}
\put(85,70){\color{blue}\framebox(30,10){$\{X,Y\}$}}
\put(60,70){\line(-2,-1){40}}
\put(60,70){\line(2,-1){40}}
\put(20,70){\line(2,-1){40}}
\put(100,70){\line(-2,-1){40}}
\put(20,70){\line(0,-1){20}}
\put(100,70){\line(0,-1){20}}
\put(10,40){\color{blue}\framebox(20,10){$\{Z\}$}}
\put(50,40){\color{blue}\framebox(20,10){$\{Y\}$}}
\put(90,40){\framebox(20,10){$\{X\}$}}
\put(20,40){\line(4,-1){40}}
\put(100,40){\line(-4,-1){40}}
\put(60,40){\line(0,-1){10}}
\put(50,20){\framebox(20,10){$\{\}$}}\end{picture}&
\begin{picture}(120,90)(0,20)
\put(10,95){(c)}
\put(40,90){\framebox(40,10){$\{X,Y,Z\}$}}
\put(60,90){\line(0,-1){10}}
\put(60,90){\line(4,-1){40}}
\put(5,70){\framebox(30,10){$\{Y,Z\}$}}
\put(45,70){\color{blue}\framebox(30,10){$\{Z,X\}$}}
\put(85,70){\color{blue}\framebox(30,10){$\{X,Y\}$}}
\put(60,70){\line(-2,-1){40}}
\put(60,70){\line(2,-1){40}}
\put(20,70){\line(2,-1){40}}
\put(100,70){\line(-2,-1){40}}
\put(100,70){\line(0,-1){20}}
\put(10,40){\framebox(20,10){$\{Z\}$}}
\put(50,40){\color{blue}\framebox(20,10){$\{Y\}$}}
\put(90,40){\framebox(20,10){$\{X\}$}}
\put(100,40){\line(-4,-1){40}}
\put(60,40){\line(0,-1){10}}
\put(50,20){\color{red}\framebox(20,10){$\{\}$}}
\put(20,40){\color{red}\line(4,-1){40}}
\put(20,70){\color{red}\line(0,-1){20}}
\put(60,90){\color{red}\line(-4,-1){40}}
\end{picture}\\
\begin{picture}(120,90)(0,30)
\put(10,95){(d)}
\put(40,90){\framebox(40,10){$\{X,Y,Z\}$}}
\put(60,90){\line(0,-1){10}}
\put(60,90){\line(4,-1){40}}
\put(5,70){\framebox(30,10){$\{Y,Z\}$}}
\put(45,70){\color{blue}\framebox(30,10){$\{Z,X\}$}}
\put(85,70){\color{blue}\framebox(30,10){$\{X,Y\}$}}
\put(60,70){\line(-2,-1){40}}
\put(60,70){\line(2,-1){40}}
\put(20,70){\color{red}\line(2,-1){40}}
\put(100,70){\line(-2,-1){40}}
\put(100,70){\line(0,-1){20}}
\put(10,40){\color{blue}\framebox(20,10){$\{Z\}$}}
\put(50,40){\framebox(20,10){$\{Y\}$}}
\put(90,40){\framebox(20,10){$\{X\}$}}
\put(100,40){\line(-4,-1){40}}
\put(20,40){\line(4,-1){40}}
\put(60,40){\color{red}\line(0,-1){10}}
\put(50,20){\color{red}\framebox(20,10){$\{\}$}}
\put(20,70){\line(0,-1){20}}
\put(60,90){\color{red}\line(-4,-1){40}}
\end{picture}&
\begin{picture}(120,90)(0,30)
\put(10,95){(e)}
\put(40,90){\framebox(40,10){$\{X,Y,Z\}$}}
\put(60,90){\line(0,-1){10}}
\put(60,90){\line(4,-1){40}}
\put(5,70){\framebox(30,10){$\{Y,Z\}$}}
\put(45,70){\framebox(30,10){$\{Z,X\}$}}
\put(85,70){\color{blue}\framebox(30,10){$\{X,Y\}$}}
\put(60,70){\line(-2,-1){40}}
\put(60,70){\line(2,-1){40}}
\put(20,70){\line(2,-1){40}}
\put(100,70){\line(-2,-1){40}}
\put(100,70){\line(0,-1){20}}
\put(10,40){\color{blue}\framebox(20,10){$\{Z\}$}}
\put(50,40){\color{blue}\framebox(20,10){$\{Y\}$}}
\put(90,40){\color{blue}\framebox(20,10){$\{X\}$}}
\put(100,40){\line(-4,-1){40}}
\put(20,40){\line(4,-1){40}}
\put(60,40){\line(0,-1){10}}
\put(50,20){\framebox(20,10){$\{\}$}}\put(20,70){\line(0,-1){20}}
\put(60,90){\line(-4,-1){40}}
\end{picture}&
\begin{picture}(120,90)(0,30)
\put(10,95){(f)}
\put(40,90){\framebox(40,10){$\{X,Y,Z\}$}}
\put(60,90){\line(0,-1){10}}
\put(60,90){\line(4,-1){40}}
\put(5,70){\framebox(30,10){$\{Y,Z\}$}}
\put(45,70){\color{blue}\framebox(30,10){$\{Z,X\}$}}
\put(85,70){\framebox(30,10){$\{X,Y\}$}}
\put(60,70){\line(-2,-1){40}}
\put(60,70){\line(2,-1){40}}
\put(20,70){\line(2,-1){40}}
\put(100,70){\line(-2,-1){40}}
\put(100,70){\line(0,-1){20}}
\put(10,40){\color{blue}\framebox(20,10){$\{Z\}$}}
\put(50,40){\color{blue}\framebox(20,10){$\{Y\}$}}
\put(90,40){\color{blue}\framebox(20,10){$\{X\}$}}
\put(100,40){\line(-4,-1){40}}
\put(20,40){\line(4,-1){40}}
\put(60,40){\line(0,-1){10}}
\put(50,20){\framebox(20,10){$\{\}$}}\put(20,70){\line(0,-1){20}}
\put(60,90){\line(-4,-1){40}}
\end{picture}
\end{tabular}
\end{center}

%% file: result_11.tex
\begin{tikzpicture}[x=1pt,y=1pt]
\definecolor{fillColor}{RGB}{255,255,255}
\path[use as bounding box,fill=fillColor,fill opacity=0.00] (0,0) rectangle (252.94,216.81);
\begin{scope}
\path[clip] (  0.00,  0.00) rectangle (252.94,216.81);
\definecolor{drawColor}{RGB}{0,0,0}

\path[draw=drawColor,line width= 0.4pt,line join=round,line cap=round] ( 55.81, 61.20) -- (221.13, 61.20);

\path[draw=drawColor,line width= 0.4pt,line join=round,line cap=round] ( 55.81, 61.20) -- ( 55.81, 55.20);

\path[draw=drawColor,line width= 0.4pt,line join=round,line cap=round] ( 97.14, 61.20) -- ( 97.14, 55.20);

\path[draw=drawColor,line width= 0.4pt,line join=round,line cap=round] (138.47, 61.20) -- (138.47, 55.20);

\path[draw=drawColor,line width= 0.4pt,line join=round,line cap=round] (179.80, 61.20) -- (179.80, 55.20);

\path[draw=drawColor,line width= 0.4pt,line join=round,line cap=round] (221.13, 61.20) -- (221.13, 55.20);

\node[text=drawColor,anchor=base,inner sep=0pt, outer sep=0pt, scale=  1.00] at ( 55.81, 39.60) {100};

\node[text=drawColor,anchor=base,inner sep=0pt, outer sep=0pt, scale=  1.00] at ( 97.14, 39.60) {150};

\node[text=drawColor,anchor=base,inner sep=0pt, outer sep=0pt, scale=  1.00] at (138.47, 39.60) {200};

\node[text=drawColor,anchor=base,inner sep=0pt, outer sep=0pt, scale=  1.00] at (179.80, 39.60) {250};

\node[text=drawColor,anchor=base,inner sep=0pt, outer sep=0pt, scale=  1.00] at (221.13, 39.60) {300};

\path[draw=drawColor,line width= 0.4pt,line join=round,line cap=round] ( 49.20, 65.14) -- ( 49.20,163.67);

\path[draw=drawColor,line width= 0.4pt,line join=round,line cap=round] ( 49.20, 65.14) -- ( 43.20, 65.14);

\path[draw=drawColor,line width= 0.4pt,line join=round,line cap=round] ( 49.20, 84.85) -- ( 43.20, 84.85);

\path[draw=drawColor,line width= 0.4pt,line join=round,line cap=round] ( 49.20,104.55) -- ( 43.20,104.55);

\path[draw=drawColor,line width= 0.4pt,line join=round,line cap=round] ( 49.20,124.26) -- ( 43.20,124.26);

\path[draw=drawColor,line width= 0.4pt,line join=round,line cap=round] ( 49.20,143.96) -- ( 43.20,143.96);

\path[draw=drawColor,line width= 0.4pt,line join=round,line cap=round] ( 49.20,163.67) -- ( 43.20,163.67);

\node[text=drawColor,rotate= 90.00,anchor=base,inner sep=0pt, outer sep=0pt, scale=  1.00] at ( 34.80, 65.14) {0.0};

\node[text=drawColor,rotate= 90.00,anchor=base,inner sep=0pt, outer sep=0pt, scale=  1.00] at ( 34.80,104.55) {0.4};

\node[text=drawColor,rotate= 90.00,anchor=base,inner sep=0pt, outer sep=0pt, scale=  1.00] at ( 34.80,143.96) {0.8};

\path[draw=drawColor,line width= 0.4pt,line join=round,line cap=round] ( 49.20, 61.20) --
	(227.75, 61.20) --
	(227.75,167.61) --
	( 49.20,167.61) --
	cycle;
\end{scope}
\begin{scope}
\path[clip] (  0.00,  0.00) rectangle (252.94,216.81);
\definecolor{drawColor}{RGB}{0,0,0}

\node[text=drawColor,anchor=base,inner sep=0pt, outer sep=0pt, scale=  1.20] at (138.47,188.07) {\bfseries p=15: Criteria A};

\node[text=drawColor,anchor=base,inner sep=0pt, outer sep=0pt, scale=  1.00] at (138.47, 15.60) {Sample Size};

\node[text=drawColor,rotate= 90.00,anchor=base,inner sep=0pt, outer sep=0pt, scale=  1.00] at ( 10.80,114.41) {Error Rate};
\end{scope}
\begin{scope}
\path[clip] ( 49.20, 61.20) rectangle (227.75,167.61);
\definecolor{drawColor}{RGB}{0,0,255}

\path[draw=drawColor,line width= 0.4pt,line join=round,line cap=round] ( 55.81,153.82) circle (  2.25);

\path[draw=drawColor,line width= 0.4pt,line join=round,line cap=round] (138.47,114.40) circle (  2.25);

\path[draw=drawColor,line width= 0.4pt,line join=round,line cap=round] (221.13, 80.91) circle (  2.25);

\path[draw=drawColor,line width= 0.4pt,line join=round,line cap=round] ( 55.81,153.82) --
	(138.47,114.40) --
	(221.13, 80.91);
\definecolor{drawColor}{RGB}{0,255,0}

\path[draw=drawColor,line width= 0.4pt,line join=round,line cap=round] ( 55.81,167.17) --
	( 58.84,161.92) --
	( 52.78,161.92) --
	cycle;

\path[draw=drawColor,line width= 0.4pt,line join=round,line cap=round] (138.47,167.17) --
	(141.50,161.92) --
	(135.44,161.92) --
	cycle;

\path[draw=drawColor,line width= 0.4pt,line join=round,line cap=round] (221.13,167.17) --
	(224.16,161.92) --
	(218.10,161.92) --
	cycle;

\path[draw=drawColor,line width= 0.4pt,dash pattern=on 4pt off 4pt ,line join=round,line cap=round] ( 55.81,163.67) --
	(138.47,163.67) --
	(221.13,163.67);
\definecolor{drawColor}{RGB}{0,0,0}

\path[draw=drawColor,line width= 0.4pt,line join=round,line cap=round] ( 52.63,102.58) -- ( 58.99,102.58);

\path[draw=drawColor,line width= 0.4pt,line join=round,line cap=round] ( 55.81, 99.40) -- ( 55.81,105.76);

\path[draw=drawColor,line width= 0.4pt,line join=round,line cap=round] (135.29, 74.99) -- (141.65, 74.99);

\path[draw=drawColor,line width= 0.4pt,line join=round,line cap=round] (138.47, 71.81) -- (138.47, 78.18);

\path[draw=drawColor,line width= 0.4pt,line join=round,line cap=round] (217.95, 65.14) -- (224.31, 65.14);

\path[draw=drawColor,line width= 0.4pt,line join=round,line cap=round] (221.13, 61.96) -- (221.13, 68.32);

\path[draw=drawColor,line width= 0.4pt,dash pattern=on 1pt off 3pt on 4pt off 3pt ,line join=round,line cap=round] ( 55.81,102.58) --
	(138.47, 74.99) --
	(221.13, 65.14);
\definecolor{drawColor}{RGB}{255,165,0}

\path[draw=drawColor,line width= 0.4pt,line join=round,line cap=round] ( 52.63,163.67) --
	( 55.81,166.85) --
	( 58.99,163.67) --
	( 55.81,160.49) --
	cycle;

\path[draw=drawColor,line width= 0.4pt,line join=round,line cap=round] (135.29,143.96) --
	(138.47,147.15) --
	(141.65,143.96) --
	(138.47,140.78) --
	cycle;

\path[draw=drawColor,line width= 0.4pt,line join=round,line cap=round] (217.95, 94.70) --
	(221.13, 97.88) --
	(224.31, 94.70) --
	(221.13, 91.52) --
	cycle;

\path[draw=drawColor,line width= 0.4pt,dash pattern=on 7pt off 3pt ,line join=round,line cap=round] ( 55.81,163.67) --
	(138.47,143.96) --
	(221.13, 94.70);
\definecolor{drawColor}{RGB}{255,0,0}

\path[draw=drawColor,line width= 0.4pt,line join=round,line cap=round] ( 53.56, 74.71) rectangle ( 58.06, 79.21);

\path[draw=drawColor,line width= 0.4pt,line join=round,line cap=round] ( 53.56, 74.71) -- ( 58.06, 79.21);

\path[draw=drawColor,line width= 0.4pt,line join=round,line cap=round] ( 53.56, 79.21) -- ( 58.06, 74.71);

\path[draw=drawColor,line width= 0.4pt,line join=round,line cap=round] (136.22, 64.86) rectangle (140.72, 69.36);

\path[draw=drawColor,line width= 0.4pt,line join=round,line cap=round] (136.22, 64.86) -- (140.72, 69.36);

\path[draw=drawColor,line width= 0.4pt,line join=round,line cap=round] (136.22, 69.36) -- (140.72, 64.86);

\path[draw=drawColor,line width= 0.4pt,line join=round,line cap=round] (218.88, 62.89) rectangle (223.38, 67.39);

\path[draw=drawColor,line width= 0.4pt,line join=round,line cap=round] (218.88, 62.89) -- (223.38, 67.39);

\path[draw=drawColor,line width= 0.4pt,line join=round,line cap=round] (218.88, 67.39) -- (223.38, 62.89);

\path[draw=drawColor,line width= 0.4pt,line join=round,line cap=round] ( 55.81, 76.96) --
	(138.47, 67.11) --
	(221.13, 65.14);
\definecolor{drawColor}{RGB}{0,0,0}

\path[draw=drawColor,line width= 0.4pt,line join=round,line cap=round] (156.94,167.61) rectangle (227.75, 95.61);
\definecolor{drawColor}{RGB}{0,0,255}

\path[draw=drawColor,line width= 0.4pt,line join=round,line cap=round] (159.64,155.61) -- (177.64,155.61);
\definecolor{drawColor}{RGB}{0,255,0}

\path[draw=drawColor,line width= 0.4pt,dash pattern=on 4pt off 4pt ,line join=round,line cap=round] (159.64,143.61) -- (177.64,143.61);
\definecolor{drawColor}{RGB}{0,0,0}

\path[draw=drawColor,line width= 0.4pt,dash pattern=on 1pt off 3pt on 4pt off 3pt ,line join=round,line cap=round] (159.64,131.61) -- (177.64,131.61);
\definecolor{drawColor}{RGB}{255,165,0}

\path[draw=drawColor,line width= 0.4pt,dash pattern=on 7pt off 3pt ,line join=round,line cap=round] (159.64,119.61) -- (177.64,119.61);
\definecolor{drawColor}{RGB}{255,0,0}

\path[draw=drawColor,line width= 0.4pt,line join=round,line cap=round] (159.64,107.61) -- (177.64,107.61);
\definecolor{drawColor}{RGB}{0,0,255}

\path[draw=drawColor,line width= 0.4pt,line join=round,line cap=round] (168.64,155.61) circle (  2.25);
\definecolor{drawColor}{RGB}{0,255,0}

\path[draw=drawColor,line width= 0.4pt,line join=round,line cap=round] (168.64,147.11) --
	(171.67,141.86) --
	(165.61,141.86) --
	cycle;
\definecolor{drawColor}{RGB}{0,0,0}

\path[draw=drawColor,line width= 0.4pt,line join=round,line cap=round] (165.46,131.61) -- (171.82,131.61);

\path[draw=drawColor,line width= 0.4pt,line join=round,line cap=round] (168.64,128.43) -- (168.64,134.79);
\definecolor{drawColor}{RGB}{255,165,0}

\path[draw=drawColor,line width= 0.4pt,line join=round,line cap=round] (165.46,119.61) --
	(168.64,122.79) --
	(171.82,119.61) --
	(168.64,116.43) --
	cycle;
\definecolor{drawColor}{RGB}{255,0,0}

\path[draw=drawColor,line width= 0.4pt,line join=round,line cap=round] (166.39,105.36) rectangle (170.89,109.86);

\path[draw=drawColor,line width= 0.4pt,line join=round,line cap=round] (166.39,105.36) -- (170.89,109.86);

\path[draw=drawColor,line width= 0.4pt,line join=round,line cap=round] (166.39,109.86) -- (170.89,105.36);
\definecolor{drawColor}{RGB}{0,0,0}

\node[text=drawColor,anchor=base west,inner sep=0pt, outer sep=0pt, scale=  1.00] at (186.64,152.17) {Pairwise};

\node[text=drawColor,anchor=base west,inner sep=0pt, outer sep=0pt, scale=  1.00] at (186.64,140.17) {Kernel};

\node[text=drawColor,anchor=base west,inner sep=0pt, outer sep=0pt, scale=  1.00] at (186.64,128.17) {HSIC};

\node[text=drawColor,anchor=base west,inner sep=0pt, outer sep=0pt, scale=  1.00] at (186.64,116.17) {ICA};

\node[text=drawColor,anchor=base west,inner sep=0pt, outer sep=0pt, scale=  1.00] at (186.64,104.17) {MMI};
\end{scope}
\end{tikzpicture}

%% file: result_21.tex
\begin{tikzpicture}[x=1pt,y=1pt]
\definecolor{fillColor}{RGB}{255,255,255}
\path[use as bounding box,fill=fillColor,fill opacity=0.00] (0,0) rectangle (252.94,216.81);
\begin{scope}
\path[clip] (  0.00,  0.00) rectangle (252.94,216.81);
\definecolor{drawColor}{RGB}{0,0,0}

\path[draw=drawColor,line width= 0.4pt,line join=round,line cap=round] ( 55.81, 61.20) -- (221.13, 61.20);

\path[draw=drawColor,line width= 0.4pt,line join=round,line cap=round] ( 55.81, 61.20) -- ( 55.81, 55.20);

\path[draw=drawColor,line width= 0.4pt,line join=round,line cap=round] ( 97.14, 61.20) -- ( 97.14, 55.20);

\path[draw=drawColor,line width= 0.4pt,line join=round,line cap=round] (138.47, 61.20) -- (138.47, 55.20);

\path[draw=drawColor,line width= 0.4pt,line join=round,line cap=round] (179.80, 61.20) -- (179.80, 55.20);

\path[draw=drawColor,line width= 0.4pt,line join=round,line cap=round] (221.13, 61.20) -- (221.13, 55.20);

\node[text=drawColor,anchor=base,inner sep=0pt, outer sep=0pt, scale=  1.00] at ( 55.81, 39.60) {100};

\node[text=drawColor,anchor=base,inner sep=0pt, outer sep=0pt, scale=  1.00] at ( 97.14, 39.60) {150};

\node[text=drawColor,anchor=base,inner sep=0pt, outer sep=0pt, scale=  1.00] at (138.47, 39.60) {200};

\node[text=drawColor,anchor=base,inner sep=0pt, outer sep=0pt, scale=  1.00] at (179.80, 39.60) {250};

\node[text=drawColor,anchor=base,inner sep=0pt, outer sep=0pt, scale=  1.00] at (221.13, 39.60) {300};

\path[draw=drawColor,line width= 0.4pt,line join=round,line cap=round] ( 49.20, 65.14) -- ( 49.20,163.67);

\path[draw=drawColor,line width= 0.4pt,line join=round,line cap=round] ( 49.20, 65.14) -- ( 43.20, 65.14);

\path[draw=drawColor,line width= 0.4pt,line join=round,line cap=round] ( 49.20, 84.85) -- ( 43.20, 84.85);

\path[draw=drawColor,line width= 0.4pt,line join=round,line cap=round] ( 49.20,104.55) -- ( 43.20,104.55);

\path[draw=drawColor,line width= 0.4pt,line join=round,line cap=round] ( 49.20,124.26) -- ( 43.20,124.26);

\path[draw=drawColor,line width= 0.4pt,line join=round,line cap=round] ( 49.20,143.96) -- ( 43.20,143.96);

\path[draw=drawColor,line width= 0.4pt,line join=round,line cap=round] ( 49.20,163.67) -- ( 43.20,163.67);

\node[text=drawColor,rotate= 90.00,anchor=base,inner sep=0pt, outer sep=0pt, scale=  1.00] at ( 34.80, 65.14) {0.0};

\node[text=drawColor,rotate= 90.00,anchor=base,inner sep=0pt, outer sep=0pt, scale=  1.00] at ( 34.80,104.55) {0.4};

\node[text=drawColor,rotate= 90.00,anchor=base,inner sep=0pt, outer sep=0pt, scale=  1.00] at ( 34.80,143.96) {0.8};

\path[draw=drawColor,line width= 0.4pt,line join=round,line cap=round] ( 49.20, 61.20) --
	(227.75, 61.20) --
	(227.75,167.61) --
	( 49.20,167.61) --
	cycle;
\end{scope}
\begin{scope}
\path[clip] (  0.00,  0.00) rectangle (252.94,216.81);
\definecolor{drawColor}{RGB}{0,0,0}

\node[text=drawColor,anchor=base,inner sep=0pt, outer sep=0pt, scale=  1.20] at (138.47,188.07) {\bfseries p=15: Criteria B};

\node[text=drawColor,anchor=base,inner sep=0pt, outer sep=0pt, scale=  1.00] at (138.47, 15.60) {Sample Size};

\node[text=drawColor,rotate= 90.00,anchor=base,inner sep=0pt, outer sep=0pt, scale=  1.00] at ( 10.80,114.41) {Error Rate};
\end{scope}
\begin{scope}
\path[clip] ( 49.20, 61.20) rectangle (227.75,167.61);
\definecolor{drawColor}{RGB}{0,0,255}

\path[draw=drawColor,line width= 0.4pt,line join=round,line cap=round] ( 55.81, 98.64) circle (  2.25);

\path[draw=drawColor,line width= 0.4pt,line join=round,line cap=round] (138.47, 83.86) circle (  2.25);

\path[draw=drawColor,line width= 0.4pt,line join=round,line cap=round] (221.13, 75.98) circle (  2.25);

\path[draw=drawColor,line width= 0.4pt,line join=round,line cap=round] ( 55.81, 98.64) --
	(138.47, 83.86) --
	(221.13, 75.98);
\definecolor{drawColor}{RGB}{0,255,0}

\path[draw=drawColor,line width= 0.4pt,line join=round,line cap=round] ( 55.81, 97.21) --
	( 58.84, 91.96) --
	( 52.78, 91.96) --
	cycle;

\path[draw=drawColor,line width= 0.4pt,line join=round,line cap=round] (138.47, 95.74) --
	(141.50, 90.49) --
	(135.44, 90.49) --
	cycle;

\path[draw=drawColor,line width= 0.4pt,line join=round,line cap=round] (221.13, 94.26) --
	(224.16, 89.01) --
	(218.10, 89.01) --
	cycle;

\path[draw=drawColor,line width= 0.4pt,dash pattern=on 4pt off 4pt ,line join=round,line cap=round] ( 55.81, 93.71) --
	(138.47, 92.24) --
	(221.13, 90.76);
\definecolor{drawColor}{RGB}{0,0,0}

\path[draw=drawColor,line width= 0.4pt,line join=round,line cap=round] ( 52.63, 70.07) -- ( 58.99, 70.07);

\path[draw=drawColor,line width= 0.4pt,line join=round,line cap=round] ( 55.81, 66.89) -- ( 55.81, 73.25);

\path[draw=drawColor,line width= 0.4pt,line join=round,line cap=round] (135.29, 65.14) -- (141.65, 65.14);

\path[draw=drawColor,line width= 0.4pt,line join=round,line cap=round] (138.47, 61.96) -- (138.47, 68.32);

\path[draw=drawColor,line width= 0.4pt,line join=round,line cap=round] (217.95, 65.14) -- (224.31, 65.14);

\path[draw=drawColor,line width= 0.4pt,line join=round,line cap=round] (221.13, 61.96) -- (221.13, 68.32);

\path[draw=drawColor,line width= 0.4pt,dash pattern=on 1pt off 3pt on 4pt off 3pt ,line join=round,line cap=round] ( 55.81, 70.07) --
	(138.47, 65.14) --
	(221.13, 65.14);
\definecolor{drawColor}{RGB}{255,165,0}

\path[draw=drawColor,line width= 0.4pt,line join=round,line cap=round] ( 52.63, 84.85) --
	( 55.81, 88.03) --
	( 58.99, 84.85) --
	( 55.81, 81.66) --
	cycle;

\path[draw=drawColor,line width= 0.4pt,line join=round,line cap=round] (135.29, 83.86) --
	(138.47, 87.04) --
	(141.65, 83.86) --
	(138.47, 80.68) --
	cycle;

\path[draw=drawColor,line width= 0.4pt,line join=round,line cap=round] (217.95, 73.02) --
	(221.13, 76.21) --
	(224.31, 73.02) --
	(221.13, 69.84) --
	cycle;

\path[draw=drawColor,line width= 0.4pt,dash pattern=on 7pt off 3pt ,line join=round,line cap=round] ( 55.81, 84.85) --
	(138.47, 83.86) --
	(221.13, 73.02);
\definecolor{drawColor}{RGB}{255,0,0}

\path[draw=drawColor,line width= 0.4pt,line join=round,line cap=round] ( 53.56, 63.88) rectangle ( 58.06, 68.38);

\path[draw=drawColor,line width= 0.4pt,line join=round,line cap=round] ( 53.56, 63.88) -- ( 58.06, 68.38);

\path[draw=drawColor,line width= 0.4pt,line join=round,line cap=round] ( 53.56, 68.38) -- ( 58.06, 63.88);

\path[draw=drawColor,line width= 0.4pt,line join=round,line cap=round] (136.22, 62.89) rectangle (140.72, 67.39);

\path[draw=drawColor,line width= 0.4pt,line join=round,line cap=round] (136.22, 62.89) -- (140.72, 67.39);

\path[draw=drawColor,line width= 0.4pt,line join=round,line cap=round] (136.22, 67.39) -- (140.72, 62.89);

\path[draw=drawColor,line width= 0.4pt,line join=round,line cap=round] (218.88, 62.89) rectangle (223.38, 67.39);

\path[draw=drawColor,line width= 0.4pt,line join=round,line cap=round] (218.88, 62.89) -- (223.38, 67.39);

\path[draw=drawColor,line width= 0.4pt,line join=round,line cap=round] (218.88, 67.39) -- (223.38, 62.89);

\path[draw=drawColor,line width= 0.4pt,line join=round,line cap=round] ( 55.81, 66.13) --
	(138.47, 65.14) --
	(221.13, 65.14);
\definecolor{drawColor}{RGB}{0,0,0}

\path[draw=drawColor,line width= 0.4pt,line join=round,line cap=round] (156.94,167.61) rectangle (227.75, 95.61);
\definecolor{drawColor}{RGB}{0,0,255}

\path[draw=drawColor,line width= 0.4pt,line join=round,line cap=round] (159.64,155.61) -- (177.64,155.61);
\definecolor{drawColor}{RGB}{0,255,0}

\path[draw=drawColor,line width= 0.4pt,dash pattern=on 4pt off 4pt ,line join=round,line cap=round] (159.64,143.61) -- (177.64,143.61);
\definecolor{drawColor}{RGB}{0,0,0}

\path[draw=drawColor,line width= 0.4pt,dash pattern=on 1pt off 3pt on 4pt off 3pt ,line join=round,line cap=round] (159.64,131.61) -- (177.64,131.61);
\definecolor{drawColor}{RGB}{255,165,0}

\path[draw=drawColor,line width= 0.4pt,dash pattern=on 7pt off 3pt ,line join=round,line cap=round] (159.64,119.61) -- (177.64,119.61);
\definecolor{drawColor}{RGB}{255,0,0}

\path[draw=drawColor,line width= 0.4pt,line join=round,line cap=round] (159.64,107.61) -- (177.64,107.61);
\definecolor{drawColor}{RGB}{0,0,255}

\path[draw=drawColor,line width= 0.4pt,line join=round,line cap=round] (168.64,155.61) circle (  2.25);
\definecolor{drawColor}{RGB}{0,255,0}

\path[draw=drawColor,line width= 0.4pt,line join=round,line cap=round] (168.64,147.11) --
	(171.67,141.86) --
	(165.61,141.86) --
	cycle;
\definecolor{drawColor}{RGB}{0,0,0}

\path[draw=drawColor,line width= 0.4pt,line join=round,line cap=round] (165.46,131.61) -- (171.82,131.61);

\path[draw=drawColor,line width= 0.4pt,line join=round,line cap=round] (168.64,128.43) -- (168.64,134.79);
\definecolor{drawColor}{RGB}{255,165,0}

\path[draw=drawColor,line width= 0.4pt,line join=round,line cap=round] (165.46,119.61) --
	(168.64,122.79) --
	(171.82,119.61) --
	(168.64,116.43) --
	cycle;
\definecolor{drawColor}{RGB}{255,0,0}

\path[draw=drawColor,line width= 0.4pt,line join=round,line cap=round] (166.39,105.36) rectangle (170.89,109.86);

\path[draw=drawColor,line width= 0.4pt,line join=round,line cap=round] (166.39,105.36) -- (170.89,109.86);

\path[draw=drawColor,line width= 0.4pt,line join=round,line cap=round] (166.39,109.86) -- (170.89,105.36);
\definecolor{drawColor}{RGB}{0,0,0}

\node[text=drawColor,anchor=base west,inner sep=0pt, outer sep=0pt, scale=  1.00] at (186.64,152.17) {Pairwise};

\node[text=drawColor,anchor=base west,inner sep=0pt, outer sep=0pt, scale=  1.00] at (186.64,140.17) {Kernel};

\node[text=drawColor,anchor=base west,inner sep=0pt, outer sep=0pt, scale=  1.00] at (186.64,128.17) {HSIC};

\node[text=drawColor,anchor=base west,inner sep=0pt, outer sep=0pt, scale=  1.00] at (186.64,116.17) {ICA};

\node[text=drawColor,anchor=base west,inner sep=0pt, outer sep=0pt, scale=  1.00] at (186.64,104.17) {MMI};
\end{scope}
\end{tikzpicture}

%% file: result_31.tex
\begin{tikzpicture}[x=1pt,y=1pt]
\definecolor{fillColor}{RGB}{255,255,255}
\path[use as bounding box,fill=fillColor,fill opacity=0.00] (0,0) rectangle (252.94,216.81);
\begin{scope}
\path[clip] (  0.00,  0.00) rectangle (252.94,216.81);
\definecolor{drawColor}{RGB}{0,0,0}

\path[draw=drawColor,line width= 0.4pt,line join=round,line cap=round] ( 74.18, 61.20) -- (221.13, 61.20);

\path[draw=drawColor,line width= 0.4pt,line join=round,line cap=round] ( 74.18, 61.20) -- ( 74.18, 55.20);

\path[draw=drawColor,line width= 0.4pt,line join=round,line cap=round] (110.92, 61.20) -- (110.92, 55.20);

\path[draw=drawColor,line width= 0.4pt,line join=round,line cap=round] (147.66, 61.20) -- (147.66, 55.20);

\path[draw=drawColor,line width= 0.4pt,line join=round,line cap=round] (184.39, 61.20) -- (184.39, 55.20);

\path[draw=drawColor,line width= 0.4pt,line join=round,line cap=round] (221.13, 61.20) -- (221.13, 55.20);

\node[text=drawColor,anchor=base,inner sep=0pt, outer sep=0pt, scale=  1.00] at ( 74.18, 39.60) {200};

\node[text=drawColor,anchor=base,inner sep=0pt, outer sep=0pt, scale=  1.00] at (110.92, 39.60) {400};

\node[text=drawColor,anchor=base,inner sep=0pt, outer sep=0pt, scale=  1.00] at (147.66, 39.60) {600};

\node[text=drawColor,anchor=base,inner sep=0pt, outer sep=0pt, scale=  1.00] at (184.39, 39.60) {800};

\node[text=drawColor,anchor=base,inner sep=0pt, outer sep=0pt, scale=  1.00] at (221.13, 39.60) {1000};

\path[draw=drawColor,line width= 0.4pt,line join=round,line cap=round] ( 49.20, 65.14) -- ( 49.20,163.67);

\path[draw=drawColor,line width= 0.4pt,line join=round,line cap=round] ( 49.20, 65.14) -- ( 43.20, 65.14);

\path[draw=drawColor,line width= 0.4pt,line join=round,line cap=round] ( 49.20, 84.85) -- ( 43.20, 84.85);

\path[draw=drawColor,line width= 0.4pt,line join=round,line cap=round] ( 49.20,104.55) -- ( 43.20,104.55);

\path[draw=drawColor,line width= 0.4pt,line join=round,line cap=round] ( 49.20,124.26) -- ( 43.20,124.26);

\path[draw=drawColor,line width= 0.4pt,line join=round,line cap=round] ( 49.20,143.96) -- ( 43.20,143.96);

\path[draw=drawColor,line width= 0.4pt,line join=round,line cap=round] ( 49.20,163.67) -- ( 43.20,163.67);

\node[text=drawColor,rotate= 90.00,anchor=base,inner sep=0pt, outer sep=0pt, scale=  1.00] at ( 34.80, 65.14) {0.0};

\node[text=drawColor,rotate= 90.00,anchor=base,inner sep=0pt, outer sep=0pt, scale=  1.00] at ( 34.80,104.55) {0.4};

\node[text=drawColor,rotate= 90.00,anchor=base,inner sep=0pt, outer sep=0pt, scale=  1.00] at ( 34.80,143.96) {0.8};

\path[draw=drawColor,line width= 0.4pt,line join=round,line cap=round] ( 49.20, 61.20) --
	(227.75, 61.20) --
	(227.75,167.61) --
	( 49.20,167.61) --
	cycle;
\end{scope}
\begin{scope}
\path[clip] (  0.00,  0.00) rectangle (252.94,216.81);
\definecolor{drawColor}{RGB}{0,0,0}

\node[text=drawColor,anchor=base,inner sep=0pt, outer sep=0pt, scale=  1.20] at (138.47,188.07) {\bfseries p=30: Criteria A};

\node[text=drawColor,anchor=base,inner sep=0pt, outer sep=0pt, scale=  1.00] at (138.47, 15.60) {Sample Size};

\node[text=drawColor,rotate= 90.00,anchor=base,inner sep=0pt, outer sep=0pt, scale=  1.00] at ( 10.80,114.41) {Error Rate};
\end{scope}
\begin{scope}
\path[clip] ( 49.20, 61.20) rectangle (227.75,167.61);
\definecolor{drawColor}{RGB}{0,0,255}

\path[draw=drawColor,line width= 0.4pt,line join=round,line cap=round] ( 55.81,163.67) circle (  2.25);

\path[draw=drawColor,line width= 0.4pt,line join=round,line cap=round] ( 74.18,158.74) circle (  2.25);

\path[draw=drawColor,line width= 0.4pt,line join=round,line cap=round] ( 92.55,143.96) circle (  2.25);

\path[draw=drawColor,line width= 0.4pt,line join=round,line cap=round] (129.29, 94.70) circle (  2.25);

\path[draw=drawColor,line width= 0.4pt,line join=round,line cap=round] (166.03, 74.99) circle (  2.25);

\path[draw=drawColor,line width= 0.4pt,line join=round,line cap=round] (221.13, 65.14) circle (  2.25);

\path[draw=drawColor,line width= 0.4pt,line join=round,line cap=round] ( 55.81,163.67) --
	( 74.18,158.74) --
	( 92.55,143.96) --
	(129.29, 94.70) --
	(166.03, 74.99) --
	(221.13, 65.14);
\definecolor{drawColor}{RGB}{0,0,0}

\path[draw=drawColor,line width= 0.4pt,line join=round,line cap=round] ( 52.63,124.26) -- ( 58.99,124.26);

\path[draw=drawColor,line width= 0.4pt,line join=round,line cap=round] ( 55.81,121.08) -- ( 55.81,127.44);

\path[draw=drawColor,line width= 0.4pt,line join=round,line cap=round] ( 71.00, 84.85) -- ( 77.36, 84.85);

\path[draw=drawColor,line width= 0.4pt,line join=round,line cap=round] ( 74.18, 81.66) -- ( 74.18, 88.03);

\path[draw=drawColor,line width= 0.4pt,line join=round,line cap=round] ( 89.37, 74.99) -- ( 95.73, 74.99);

\path[draw=drawColor,line width= 0.4pt,line join=round,line cap=round] ( 92.55, 71.81) -- ( 92.55, 78.18);

\path[draw=drawColor,line width= 0.4pt,line join=round,line cap=round] (126.11, 65.14) -- (132.47, 65.14);

\path[draw=drawColor,line width= 0.4pt,line join=round,line cap=round] (129.29, 61.96) -- (129.29, 68.32);

\path[draw=drawColor,line width= 0.4pt,line join=round,line cap=round] (162.84, 65.14) -- (169.21, 65.14);

\path[draw=drawColor,line width= 0.4pt,line join=round,line cap=round] (166.03, 61.96) -- (166.03, 68.32);

\path[draw=drawColor,line width= 0.4pt,line join=round,line cap=round] (217.95, 65.14) -- (224.31, 65.14);

\path[draw=drawColor,line width= 0.4pt,line join=round,line cap=round] (221.13, 61.96) -- (221.13, 68.32);

\path[draw=drawColor,line width= 0.4pt,dash pattern=on 1pt off 3pt on 4pt off 3pt ,line join=round,line cap=round] ( 55.81,124.26) --
	( 74.18, 84.85) --
	( 92.55, 74.99) --
	(129.29, 65.14) --
	(166.03, 65.14) --
	(221.13, 65.14);
\definecolor{drawColor}{RGB}{255,165,0}

\path[draw=drawColor,line width= 0.4pt,line join=round,line cap=round] ( 52.63,163.67) --
	( 55.81,166.85) --
	( 58.99,163.67) --
	( 55.81,160.49) --
	cycle;

\path[draw=drawColor,line width= 0.4pt,line join=round,line cap=round] ( 71.00,163.67) --
	( 74.18,166.85) --
	( 77.36,163.67) --
	( 74.18,160.49) --
	cycle;

\path[draw=drawColor,line width= 0.4pt,line join=round,line cap=round] ( 89.37,163.67) --
	( 92.55,166.85) --
	( 95.73,163.67) --
	( 92.55,160.49) --
	cycle;

\path[draw=drawColor,line width= 0.4pt,line join=round,line cap=round] (126.11, 89.77) --
	(129.29, 92.96) --
	(132.47, 89.77) --
	(129.29, 86.59) --
	cycle;

\path[draw=drawColor,line width= 0.4pt,line join=round,line cap=round] (162.84, 70.07) --
	(166.03, 73.25) --
	(169.21, 70.07) --
	(166.03, 66.89) --
	cycle;

\path[draw=drawColor,line width= 0.4pt,line join=round,line cap=round] (217.95, 65.14) --
	(221.13, 68.32) --
	(224.31, 65.14) --
	(221.13, 61.96) --
	cycle;

\path[draw=drawColor,line width= 0.4pt,dash pattern=on 7pt off 3pt ,line join=round,line cap=round] ( 55.81,163.67) --
	( 74.18,163.67) --
	( 92.55,163.67) --
	(129.29, 89.77) --
	(166.03, 70.07) --
	(221.13, 65.14);
\definecolor{drawColor}{RGB}{255,0,0}

\path[draw=drawColor,line width= 0.4pt,line join=round,line cap=round] ( 53.56,102.30) rectangle ( 58.06,106.80);

\path[draw=drawColor,line width= 0.4pt,line join=round,line cap=round] ( 53.56,102.30) -- ( 58.06,106.80);

\path[draw=drawColor,line width= 0.4pt,line join=round,line cap=round] ( 53.56,106.80) -- ( 58.06,102.30);

\path[draw=drawColor,line width= 0.4pt,line join=round,line cap=round] ( 71.93, 69.79) rectangle ( 76.43, 74.29);

\path[draw=drawColor,line width= 0.4pt,line join=round,line cap=round] ( 71.93, 69.79) -- ( 76.43, 74.29);

\path[draw=drawColor,line width= 0.4pt,line join=round,line cap=round] ( 71.93, 74.29) -- ( 76.43, 69.79);

\path[draw=drawColor,line width= 0.4pt,line join=round,line cap=round] ( 90.30, 62.89) rectangle ( 94.80, 67.39);

\path[draw=drawColor,line width= 0.4pt,line join=round,line cap=round] ( 90.30, 62.89) -- ( 94.80, 67.39);

\path[draw=drawColor,line width= 0.4pt,line join=round,line cap=round] ( 90.30, 67.39) -- ( 94.80, 62.89);

\path[draw=drawColor,line width= 0.4pt,line join=round,line cap=round] (127.04, 62.89) rectangle (131.54, 67.39);

\path[draw=drawColor,line width= 0.4pt,line join=round,line cap=round] (127.04, 62.89) -- (131.54, 67.39);

\path[draw=drawColor,line width= 0.4pt,line join=round,line cap=round] (127.04, 67.39) -- (131.54, 62.89);

\path[draw=drawColor,line width= 0.4pt,line join=round,line cap=round] (163.78, 62.89) rectangle (168.28, 67.39);

\path[draw=drawColor,line width= 0.4pt,line join=round,line cap=round] (163.78, 62.89) -- (168.28, 67.39);

\path[draw=drawColor,line width= 0.4pt,line join=round,line cap=round] (163.78, 67.39) -- (168.28, 62.89);

\path[draw=drawColor,line width= 0.4pt,line join=round,line cap=round] (218.88, 62.89) rectangle (223.38, 67.39);

\path[draw=drawColor,line width= 0.4pt,line join=round,line cap=round] (218.88, 62.89) -- (223.38, 67.39);

\path[draw=drawColor,line width= 0.4pt,line join=round,line cap=round] (218.88, 67.39) -- (223.38, 62.89);

\path[draw=drawColor,line width= 0.4pt,line join=round,line cap=round] ( 55.81,104.55) --
	( 74.18, 72.04) --
	( 92.55, 65.14) --
	(129.29, 65.14) --
	(166.03, 65.14) --
	(221.13, 65.14);
\definecolor{drawColor}{RGB}{0,0,0}

\path[draw=drawColor,line width= 0.4pt,line join=round,line cap=round] (156.94,167.61) rectangle (227.75,107.61);
\definecolor{drawColor}{RGB}{0,0,255}

\path[draw=drawColor,line width= 0.4pt,line join=round,line cap=round] (159.64,155.61) -- (177.64,155.61);
\definecolor{drawColor}{RGB}{0,0,0}

\path[draw=drawColor,line width= 0.4pt,dash pattern=on 1pt off 3pt on 4pt off 3pt ,line join=round,line cap=round] (159.64,143.61) -- (177.64,143.61);
\definecolor{drawColor}{RGB}{255,165,0}

\path[draw=drawColor,line width= 0.4pt,dash pattern=on 7pt off 3pt ,line join=round,line cap=round] (159.64,131.61) -- (177.64,131.61);
\definecolor{drawColor}{RGB}{255,0,0}

\path[draw=drawColor,line width= 0.4pt,line join=round,line cap=round] (159.64,119.61) -- (177.64,119.61);
\definecolor{drawColor}{RGB}{0,0,255}

\path[draw=drawColor,line width= 0.4pt,line join=round,line cap=round] (168.64,155.61) circle (  2.25);
\definecolor{drawColor}{RGB}{0,0,0}

\path[draw=drawColor,line width= 0.4pt,line join=round,line cap=round] (165.46,143.61) -- (171.82,143.61);

\path[draw=drawColor,line width= 0.4pt,line join=round,line cap=round] (168.64,140.43) -- (168.64,146.79);
\definecolor{drawColor}{RGB}{255,165,0}

\path[draw=drawColor,line width= 0.4pt,line join=round,line cap=round] (165.46,131.61) --
	(168.64,134.79) --
	(171.82,131.61) --
	(168.64,128.43) --
	cycle;
\definecolor{drawColor}{RGB}{255,0,0}

\path[draw=drawColor,line width= 0.4pt,line join=round,line cap=round] (166.39,117.36) rectangle (170.89,121.86);

\path[draw=drawColor,line width= 0.4pt,line join=round,line cap=round] (166.39,117.36) -- (170.89,121.86);

\path[draw=drawColor,line width= 0.4pt,line join=round,line cap=round] (166.39,121.86) -- (170.89,117.36);
\definecolor{drawColor}{RGB}{0,0,0}

\node[text=drawColor,anchor=base west,inner sep=0pt, outer sep=0pt, scale=  1.00] at (186.64,152.17) {Pairwise};

\node[text=drawColor,anchor=base west,inner sep=0pt, outer sep=0pt, scale=  1.00] at (186.64,140.17) {HSIC};

\node[text=drawColor,anchor=base west,inner sep=0pt, outer sep=0pt, scale=  1.00] at (186.64,128.17) {ICA};

\node[text=drawColor,anchor=base west,inner sep=0pt, outer sep=0pt, scale=  1.00] at (186.64,116.17) {MMI};
\end{scope}
\end{tikzpicture}

%% file: result_41.tex
\begin{tikzpicture}[x=1pt,y=1pt]
\definecolor{fillColor}{RGB}{255,255,255}
\path[use as bounding box,fill=fillColor,fill opacity=0.00] (0,0) rectangle (252.94,216.81);
\begin{scope}
\path[clip] (  0.00,  0.00) rectangle (252.94,216.81);
\definecolor{drawColor}{RGB}{0,0,0}

\path[draw=drawColor,line width= 0.4pt,line join=round,line cap=round] ( 74.18, 61.20) -- (221.13, 61.20);

\path[draw=drawColor,line width= 0.4pt,line join=round,line cap=round] ( 74.18, 61.20) -- ( 74.18, 55.20);

\path[draw=drawColor,line width= 0.4pt,line join=round,line cap=round] (110.92, 61.20) -- (110.92, 55.20);

\path[draw=drawColor,line width= 0.4pt,line join=round,line cap=round] (147.66, 61.20) -- (147.66, 55.20);

\path[draw=drawColor,line width= 0.4pt,line join=round,line cap=round] (184.39, 61.20) -- (184.39, 55.20);

\path[draw=drawColor,line width= 0.4pt,line join=round,line cap=round] (221.13, 61.20) -- (221.13, 55.20);

\node[text=drawColor,anchor=base,inner sep=0pt, outer sep=0pt, scale=  1.00] at ( 74.18, 39.60) {200};

\node[text=drawColor,anchor=base,inner sep=0pt, outer sep=0pt, scale=  1.00] at (110.92, 39.60) {400};

\node[text=drawColor,anchor=base,inner sep=0pt, outer sep=0pt, scale=  1.00] at (147.66, 39.60) {600};

\node[text=drawColor,anchor=base,inner sep=0pt, outer sep=0pt, scale=  1.00] at (184.39, 39.60) {800};

\node[text=drawColor,anchor=base,inner sep=0pt, outer sep=0pt, scale=  1.00] at (221.13, 39.60) {1000};

\path[draw=drawColor,line width= 0.4pt,line join=round,line cap=round] ( 49.20, 65.14) -- ( 49.20,163.67);

\path[draw=drawColor,line width= 0.4pt,line join=round,line cap=round] ( 49.20, 65.14) -- ( 43.20, 65.14);

\path[draw=drawColor,line width= 0.4pt,line join=round,line cap=round] ( 49.20, 84.85) -- ( 43.20, 84.85);

\path[draw=drawColor,line width= 0.4pt,line join=round,line cap=round] ( 49.20,104.55) -- ( 43.20,104.55);

\path[draw=drawColor,line width= 0.4pt,line join=round,line cap=round] ( 49.20,124.26) -- ( 43.20,124.26);

\path[draw=drawColor,line width= 0.4pt,line join=round,line cap=round] ( 49.20,143.96) -- ( 43.20,143.96);

\path[draw=drawColor,line width= 0.4pt,line join=round,line cap=round] ( 49.20,163.67) -- ( 43.20,163.67);

\node[text=drawColor,rotate= 90.00,anchor=base,inner sep=0pt, outer sep=0pt, scale=  1.00] at ( 34.80, 65.14) {0.0};

\node[text=drawColor,rotate= 90.00,anchor=base,inner sep=0pt, outer sep=0pt, scale=  1.00] at ( 34.80,104.55) {0.4};

\node[text=drawColor,rotate= 90.00,anchor=base,inner sep=0pt, outer sep=0pt, scale=  1.00] at ( 34.80,143.96) {0.8};

\path[draw=drawColor,line width= 0.4pt,line join=round,line cap=round] ( 49.20, 61.20) --
	(227.75, 61.20) --
	(227.75,167.61) --
	( 49.20,167.61) --
	cycle;
\end{scope}
\begin{scope}
\path[clip] (  0.00,  0.00) rectangle (252.94,216.81);
\definecolor{drawColor}{RGB}{0,0,0}

\node[text=drawColor,anchor=base,inner sep=0pt, outer sep=0pt, scale=  1.20] at (138.47,188.07) {\bfseries p=30: Criteria B};

\node[text=drawColor,anchor=base,inner sep=0pt, outer sep=0pt, scale=  1.00] at (138.47, 15.60) {Sample Size};

\node[text=drawColor,rotate= 90.00,anchor=base,inner sep=0pt, outer sep=0pt, scale=  1.00] at ( 10.80,114.41) {Error Rate};
\end{scope}
\begin{scope}
\path[clip] ( 49.20, 61.20) rectangle (227.75,167.61);
\definecolor{drawColor}{RGB}{0,0,255}

\path[draw=drawColor,line width= 0.4pt,line join=round,line cap=round] ( 55.81,103.57) circle (  2.25);

\path[draw=drawColor,line width= 0.4pt,line join=round,line cap=round] ( 74.18, 98.64) circle (  2.25);

\path[draw=drawColor,line width= 0.4pt,line join=round,line cap=round] ( 92.55, 91.74) circle (  2.25);

\path[draw=drawColor,line width= 0.4pt,line join=round,line cap=round] (129.29, 75.98) circle (  2.25);

\path[draw=drawColor,line width= 0.4pt,line join=round,line cap=round] (166.03, 68.10) circle (  2.25);

\path[draw=drawColor,line width= 0.4pt,line join=round,line cap=round] (221.13, 65.14) circle (  2.25);

\path[draw=drawColor,line width= 0.4pt,line join=round,line cap=round] ( 55.81,103.57) --
	( 74.18, 98.64) --
	( 92.55, 91.74) --
	(129.29, 75.98) --
	(166.03, 68.10) --
	(221.13, 65.14);
\definecolor{drawColor}{RGB}{0,0,0}

\path[draw=drawColor,line width= 0.4pt,line join=round,line cap=round] ( 52.63, 74.99) -- ( 58.99, 74.99);

\path[draw=drawColor,line width= 0.4pt,line join=round,line cap=round] ( 55.81, 71.81) -- ( 55.81, 78.18);

\path[draw=drawColor,line width= 0.4pt,line join=round,line cap=round] ( 71.00, 66.13) -- ( 77.36, 66.13);

\path[draw=drawColor,line width= 0.4pt,line join=round,line cap=round] ( 74.18, 62.94) -- ( 74.18, 69.31);

\path[draw=drawColor,line width= 0.4pt,line join=round,line cap=round] ( 89.37, 65.14) -- ( 95.73, 65.14);

\path[draw=drawColor,line width= 0.4pt,line join=round,line cap=round] ( 92.55, 61.96) -- ( 92.55, 68.32);

\path[draw=drawColor,line width= 0.4pt,line join=round,line cap=round] (126.11, 65.14) -- (132.47, 65.14);

\path[draw=drawColor,line width= 0.4pt,line join=round,line cap=round] (129.29, 61.96) -- (129.29, 68.32);

\path[draw=drawColor,line width= 0.4pt,line join=round,line cap=round] (162.84, 65.14) -- (169.21, 65.14);

\path[draw=drawColor,line width= 0.4pt,line join=round,line cap=round] (166.03, 61.96) -- (166.03, 68.32);

\path[draw=drawColor,line width= 0.4pt,line join=round,line cap=round] (217.95, 65.14) -- (224.31, 65.14);

\path[draw=drawColor,line width= 0.4pt,line join=round,line cap=round] (221.13, 61.96) -- (221.13, 68.32);

\path[draw=drawColor,line width= 0.4pt,dash pattern=on 1pt off 3pt on 4pt off 3pt ,line join=round,line cap=round] ( 55.81, 74.99) --
	( 74.18, 66.13) --
	( 92.55, 65.14) --
	(129.29, 65.14) --
	(166.03, 65.14) --
	(221.13, 65.14);
\definecolor{drawColor}{RGB}{255,165,0}

\path[draw=drawColor,line width= 0.4pt,line join=round,line cap=round] ( 52.63, 89.77) --
	( 55.81, 92.96) --
	( 58.99, 89.77) --
	( 55.81, 86.59) --
	cycle;

\path[draw=drawColor,line width= 0.4pt,line join=round,line cap=round] ( 71.00, 84.85) --
	( 74.18, 88.03) --
	( 77.36, 84.85) --
	( 74.18, 81.66) --
	cycle;

\path[draw=drawColor,line width= 0.4pt,line join=round,line cap=round] ( 89.37, 84.85) --
	( 92.55, 88.03) --
	( 95.73, 84.85) --
	( 92.55, 81.66) --
	cycle;

\path[draw=drawColor,line width= 0.4pt,line join=round,line cap=round] (126.11, 70.07) --
	(129.29, 73.25) --
	(132.47, 70.07) --
	(129.29, 66.89) --
	cycle;

\path[draw=drawColor,line width= 0.4pt,line join=round,line cap=round] (162.84, 65.14) --
	(166.03, 68.32) --
	(169.21, 65.14) --
	(166.03, 61.96) --
	cycle;

\path[draw=drawColor,line width= 0.4pt,line join=round,line cap=round] (217.95, 65.14) --
	(221.13, 68.32) --
	(224.31, 65.14) --
	(221.13, 61.96) --
	cycle;

\path[draw=drawColor,line width= 0.4pt,dash pattern=on 7pt off 3pt ,line join=round,line cap=round] ( 55.81, 89.77) --
	( 74.18, 84.85) --
	( 92.55, 84.85) --
	(129.29, 70.07) --
	(166.03, 65.14) --
	(221.13, 65.14);
\definecolor{drawColor}{RGB}{255,0,0}

\path[draw=drawColor,line width= 0.4pt,line join=round,line cap=round] ( 53.56, 62.89) rectangle ( 58.06, 67.39);

\path[draw=drawColor,line width= 0.4pt,line join=round,line cap=round] ( 53.56, 62.89) -- ( 58.06, 67.39);

\path[draw=drawColor,line width= 0.4pt,line join=round,line cap=round] ( 53.56, 67.39) -- ( 58.06, 62.89);

\path[draw=drawColor,line width= 0.4pt,line join=round,line cap=round] ( 71.93, 62.89) rectangle ( 76.43, 67.39);

\path[draw=drawColor,line width= 0.4pt,line join=round,line cap=round] ( 71.93, 62.89) -- ( 76.43, 67.39);

\path[draw=drawColor,line width= 0.4pt,line join=round,line cap=round] ( 71.93, 67.39) -- ( 76.43, 62.89);

\path[draw=drawColor,line width= 0.4pt,line join=round,line cap=round] ( 90.30, 62.89) rectangle ( 94.80, 67.39);

\path[draw=drawColor,line width= 0.4pt,line join=round,line cap=round] ( 90.30, 62.89) -- ( 94.80, 67.39);

\path[draw=drawColor,line width= 0.4pt,line join=round,line cap=round] ( 90.30, 67.39) -- ( 94.80, 62.89);

\path[draw=drawColor,line width= 0.4pt,line join=round,line cap=round] (127.04, 62.89) rectangle (131.54, 67.39);

\path[draw=drawColor,line width= 0.4pt,line join=round,line cap=round] (127.04, 62.89) -- (131.54, 67.39);

\path[draw=drawColor,line width= 0.4pt,line join=round,line cap=round] (127.04, 67.39) -- (131.54, 62.89);

\path[draw=drawColor,line width= 0.4pt,line join=round,line cap=round] (163.78, 62.89) rectangle (168.28, 67.39);

\path[draw=drawColor,line width= 0.4pt,line join=round,line cap=round] (163.78, 62.89) -- (168.28, 67.39);

\path[draw=drawColor,line width= 0.4pt,line join=round,line cap=round] (163.78, 67.39) -- (168.28, 62.89);

\path[draw=drawColor,line width= 0.4pt,line join=round,line cap=round] (218.88, 62.89) rectangle (223.38, 67.39);

\path[draw=drawColor,line width= 0.4pt,line join=round,line cap=round] (218.88, 62.89) -- (223.38, 67.39);

\path[draw=drawColor,line width= 0.4pt,line join=round,line cap=round] (218.88, 67.39) -- (223.38, 62.89);

\path[draw=drawColor,line width= 0.4pt,line join=round,line cap=round] ( 55.81, 65.14) --
	( 74.18, 65.14) --
	( 92.55, 65.14) --
	(129.29, 65.14) --
	(166.03, 65.14) --
	(221.13, 65.14);
\definecolor{drawColor}{RGB}{0,0,0}

\path[draw=drawColor,line width= 0.4pt,line join=round,line cap=round] (156.94,167.61) rectangle (227.75,107.61);
\definecolor{drawColor}{RGB}{0,0,255}

\path[draw=drawColor,line width= 0.4pt,line join=round,line cap=round] (159.64,155.61) -- (177.64,155.61);
\definecolor{drawColor}{RGB}{0,0,0}

\path[draw=drawColor,line width= 0.4pt,dash pattern=on 1pt off 3pt on 4pt off 3pt ,line join=round,line cap=round] (159.64,143.61) -- (177.64,143.61);
\definecolor{drawColor}{RGB}{255,165,0}

\path[draw=drawColor,line width= 0.4pt,dash pattern=on 7pt off 3pt ,line join=round,line cap=round] (159.64,131.61) -- (177.64,131.61);
\definecolor{drawColor}{RGB}{255,0,0}

\path[draw=drawColor,line width= 0.4pt,line join=round,line cap=round] (159.64,119.61) -- (177.64,119.61);
\definecolor{drawColor}{RGB}{0,0,255}

\path[draw=drawColor,line width= 0.4pt,line join=round,line cap=round] (168.64,155.61) circle (  2.25);
\definecolor{drawColor}{RGB}{0,0,0}

\path[draw=drawColor,line width= 0.4pt,line join=round,line cap=round] (165.46,143.61) -- (171.82,143.61);

\path[draw=drawColor,line width= 0.4pt,line join=round,line cap=round] (168.64,140.43) -- (168.64,146.79);
\definecolor{drawColor}{RGB}{255,165,0}

\path[draw=drawColor,line width= 0.4pt,line join=round,line cap=round] (165.46,131.61) --
	(168.64,134.79) --
	(171.82,131.61) --
	(168.64,128.43) --
	cycle;
\definecolor{drawColor}{RGB}{255,0,0}

\path[draw=drawColor,line width= 0.4pt,line join=round,line cap=round] (166.39,117.36) rectangle (170.89,121.86);

\path[draw=drawColor,line width= 0.4pt,line join=round,line cap=round] (166.39,117.36) -- (170.89,121.86);

\path[draw=drawColor,line width= 0.4pt,line join=round,line cap=round] (166.39,121.86) -- (170.89,117.36);
\definecolor{drawColor}{RGB}{0,0,0}

\node[text=drawColor,anchor=base west,inner sep=0pt, outer sep=0pt, scale=  1.00] at (186.64,152.17) {Pairwise};

\node[text=drawColor,anchor=base west,inner sep=0pt, outer sep=0pt, scale=  1.00] at (186.64,140.17) {HSIC};

\node[text=drawColor,anchor=base west,inner sep=0pt, outer sep=0pt, scale=  1.00] at (186.64,128.17) {ICA};

\node[text=drawColor,anchor=base west,inner sep=0pt, outer sep=0pt, scale=  1.00] at (186.64,116.17) {MMI};
\end{scope}
\end{tikzpicture}

%% file: result_51.tex
\begin{tikzpicture}[x=1pt,y=1pt]
\definecolor{fillColor}{RGB}{255,255,255}
\path[use as bounding box,fill=fillColor,fill opacity=0.00] (0,0) rectangle (252.94,216.81);
\begin{scope}
\path[clip] (  0.00,  0.00) rectangle (252.94,216.81);
\definecolor{drawColor}{RGB}{0,0,0}

\path[draw=drawColor,line width= 0.4pt,line join=round,line cap=round] ( 55.81, 61.20) -- (221.13, 61.20);

\path[draw=drawColor,line width= 0.4pt,line join=round,line cap=round] ( 55.81, 61.20) -- ( 55.81, 55.20);

\path[draw=drawColor,line width= 0.4pt,line join=round,line cap=round] ( 97.14, 61.20) -- ( 97.14, 55.20);

\path[draw=drawColor,line width= 0.4pt,line join=round,line cap=round] (138.47, 61.20) -- (138.47, 55.20);

\path[draw=drawColor,line width= 0.4pt,line join=round,line cap=round] (179.80, 61.20) -- (179.80, 55.20);

\path[draw=drawColor,line width= 0.4pt,line join=round,line cap=round] (221.13, 61.20) -- (221.13, 55.20);

\node[text=drawColor,anchor=base,inner sep=0pt, outer sep=0pt, scale=  1.00] at ( 55.81, 39.60) {100};

\node[text=drawColor,anchor=base,inner sep=0pt, outer sep=0pt, scale=  1.00] at ( 97.14, 39.60) {150};

\node[text=drawColor,anchor=base,inner sep=0pt, outer sep=0pt, scale=  1.00] at (138.47, 39.60) {200};

\node[text=drawColor,anchor=base,inner sep=0pt, outer sep=0pt, scale=  1.00] at (179.80, 39.60) {250};

\node[text=drawColor,anchor=base,inner sep=0pt, outer sep=0pt, scale=  1.00] at (221.13, 39.60) {300};

\path[draw=drawColor,line width= 0.4pt,line join=round,line cap=round] ( 49.20, 65.14) -- ( 49.20,163.67);

\path[draw=drawColor,line width= 0.4pt,line join=round,line cap=round] ( 49.20, 65.14) -- ( 43.20, 65.14);

\path[draw=drawColor,line width= 0.4pt,line join=round,line cap=round] ( 49.20, 84.85) -- ( 43.20, 84.85);

\path[draw=drawColor,line width= 0.4pt,line join=round,line cap=round] ( 49.20,104.55) -- ( 43.20,104.55);

\path[draw=drawColor,line width= 0.4pt,line join=round,line cap=round] ( 49.20,124.26) -- ( 43.20,124.26);

\path[draw=drawColor,line width= 0.4pt,line join=round,line cap=round] ( 49.20,143.96) -- ( 43.20,143.96);

\path[draw=drawColor,line width= 0.4pt,line join=round,line cap=round] ( 49.20,163.67) -- ( 43.20,163.67);

\node[text=drawColor,rotate= 90.00,anchor=base,inner sep=0pt, outer sep=0pt, scale=  1.00] at ( 34.80, 65.14) {0.0};

\node[text=drawColor,rotate= 90.00,anchor=base,inner sep=0pt, outer sep=0pt, scale=  1.00] at ( 34.80,104.55) {0.4};

\node[text=drawColor,rotate= 90.00,anchor=base,inner sep=0pt, outer sep=0pt, scale=  1.00] at ( 34.80,143.96) {0.8};

\path[draw=drawColor,line width= 0.4pt,line join=round,line cap=round] ( 49.20, 61.20) --
	(227.75, 61.20) --
	(227.75,167.61) --
	( 49.20,167.61) --
	cycle;
\end{scope}
\begin{scope}
\path[clip] (  0.00,  0.00) rectangle (252.94,216.81);
\definecolor{drawColor}{RGB}{0,0,0}

\node[text=drawColor,anchor=base,inner sep=0pt, outer sep=0pt, scale=  1.20] at (138.47,188.07) {\bfseries p=15: Criteria A};

\node[text=drawColor,anchor=base,inner sep=0pt, outer sep=0pt, scale=  1.00] at (138.47, 15.60) {Sample Size};

\node[text=drawColor,rotate= 90.00,anchor=base,inner sep=0pt, outer sep=0pt, scale=  1.00] at ( 10.80,114.41) {Error Rate};
\end{scope}
\begin{scope}
\path[clip] ( 49.20, 61.20) rectangle (227.75,167.61);
\definecolor{drawColor}{RGB}{0,0,255}

\path[draw=drawColor,line width= 0.4pt,line join=round,line cap=round] ( 55.81,163.67) circle (  2.25);

\path[draw=drawColor,line width= 0.4pt,line join=round,line cap=round] (138.47,158.74) circle (  2.25);

\path[draw=drawColor,line width= 0.4pt,line join=round,line cap=round] (221.13,156.77) circle (  2.25);

\path[draw=drawColor,line width= 0.4pt,line join=round,line cap=round] ( 55.81,163.67) --
	(138.47,158.74) --
	(221.13,156.77);
\definecolor{drawColor}{RGB}{0,255,0}

\path[draw=drawColor,line width= 0.4pt,line join=round,line cap=round] ( 55.81,167.17) --
	( 58.84,161.92) --
	( 52.78,161.92) --
	cycle;

\path[draw=drawColor,line width= 0.4pt,line join=round,line cap=round] (138.47,167.17) --
	(141.50,161.92) --
	(135.44,161.92) --
	cycle;

\path[draw=drawColor,line width= 0.4pt,line join=round,line cap=round] (221.13,167.17) --
	(224.16,161.92) --
	(218.10,161.92) --
	cycle;

\path[draw=drawColor,line width= 0.4pt,dash pattern=on 4pt off 4pt ,line join=round,line cap=round] ( 55.81,163.67) --
	(138.47,163.67) --
	(221.13,163.67);
\definecolor{drawColor}{RGB}{0,0,0}

\path[draw=drawColor,line width= 0.4pt,line join=round,line cap=round] ( 52.63,124.26) -- ( 58.99,124.26);

\path[draw=drawColor,line width= 0.4pt,line join=round,line cap=round] ( 55.81,121.08) -- ( 55.81,127.44);

\path[draw=drawColor,line width= 0.4pt,line join=round,line cap=round] (135.29,114.40) -- (141.65,114.40);

\path[draw=drawColor,line width= 0.4pt,line join=round,line cap=round] (138.47,111.22) -- (138.47,117.59);

\path[draw=drawColor,line width= 0.4pt,line join=round,line cap=round] (217.95,102.58) -- (224.31,102.58);

\path[draw=drawColor,line width= 0.4pt,line join=round,line cap=round] (221.13, 99.40) -- (221.13,105.76);

\path[draw=drawColor,line width= 0.4pt,dash pattern=on 1pt off 3pt on 4pt off 3pt ,line join=round,line cap=round] ( 55.81,124.26) --
	(138.47,114.40) --
	(221.13,102.58);
\definecolor{drawColor}{RGB}{255,165,0}

\path[draw=drawColor,line width= 0.4pt,line join=round,line cap=round] ( 52.63,163.67) --
	( 55.81,166.85) --
	( 58.99,163.67) --
	( 55.81,160.49) --
	cycle;

\path[draw=drawColor,line width= 0.4pt,line join=round,line cap=round] (135.29,163.67) --
	(138.47,166.85) --
	(141.65,163.67) --
	(138.47,160.49) --
	cycle;

\path[draw=drawColor,line width= 0.4pt,line join=round,line cap=round] (217.95,158.74) --
	(221.13,161.92) --
	(224.31,158.74) --
	(221.13,155.56) --
	cycle;

\path[draw=drawColor,line width= 0.4pt,dash pattern=on 7pt off 3pt ,line join=round,line cap=round] ( 55.81,163.67) --
	(138.47,163.67) --
	(221.13,158.74);
\definecolor{drawColor}{RGB}{255,0,0}

\path[draw=drawColor,line width= 0.4pt,line join=round,line cap=round] ( 53.56, 83.58) rectangle ( 58.06, 88.08);

\path[draw=drawColor,line width= 0.4pt,line join=round,line cap=round] ( 53.56, 83.58) -- ( 58.06, 88.08);

\path[draw=drawColor,line width= 0.4pt,line join=round,line cap=round] ( 53.56, 88.08) -- ( 58.06, 83.58);

\path[draw=drawColor,line width= 0.4pt,line join=round,line cap=round] (136.22, 67.82) rectangle (140.72, 72.32);

\path[draw=drawColor,line width= 0.4pt,line join=round,line cap=round] (136.22, 67.82) -- (140.72, 72.32);

\path[draw=drawColor,line width= 0.4pt,line join=round,line cap=round] (136.22, 72.32) -- (140.72, 67.82);

\path[draw=drawColor,line width= 0.4pt,line join=round,line cap=round] (218.88, 62.89) rectangle (223.38, 67.39);

\path[draw=drawColor,line width= 0.4pt,line join=round,line cap=round] (218.88, 62.89) -- (223.38, 67.39);

\path[draw=drawColor,line width= 0.4pt,line join=round,line cap=round] (218.88, 67.39) -- (223.38, 62.89);

\path[draw=drawColor,line width= 0.4pt,line join=round,line cap=round] ( 55.81, 85.83) --
	(138.47, 70.07) --
	(221.13, 65.14);
\definecolor{drawColor}{RGB}{0,0,0}

\path[draw=drawColor,line width= 0.4pt,line join=round,line cap=round] (156.94,167.61) rectangle (227.75, 95.61);
\definecolor{drawColor}{RGB}{0,0,255}

\path[draw=drawColor,line width= 0.4pt,line join=round,line cap=round] (159.64,155.61) -- (177.64,155.61);
\definecolor{drawColor}{RGB}{0,255,0}

\path[draw=drawColor,line width= 0.4pt,dash pattern=on 4pt off 4pt ,line join=round,line cap=round] (159.64,143.61) -- (177.64,143.61);
\definecolor{drawColor}{RGB}{0,0,0}

\path[draw=drawColor,line width= 0.4pt,dash pattern=on 1pt off 3pt on 4pt off 3pt ,line join=round,line cap=round] (159.64,131.61) -- (177.64,131.61);
\definecolor{drawColor}{RGB}{255,165,0}

\path[draw=drawColor,line width= 0.4pt,dash pattern=on 7pt off 3pt ,line join=round,line cap=round] (159.64,119.61) -- (177.64,119.61);
\definecolor{drawColor}{RGB}{255,0,0}

\path[draw=drawColor,line width= 0.4pt,line join=round,line cap=round] (159.64,107.61) -- (177.64,107.61);
\definecolor{drawColor}{RGB}{0,0,255}

\path[draw=drawColor,line width= 0.4pt,line join=round,line cap=round] (168.64,155.61) circle (  2.25);
\definecolor{drawColor}{RGB}{0,255,0}

\path[draw=drawColor,line width= 0.4pt,line join=round,line cap=round] (168.64,147.11) --
	(171.67,141.86) --
	(165.61,141.86) --
	cycle;
\definecolor{drawColor}{RGB}{0,0,0}

\path[draw=drawColor,line width= 0.4pt,line join=round,line cap=round] (165.46,131.61) -- (171.82,131.61);

\path[draw=drawColor,line width= 0.4pt,line join=round,line cap=round] (168.64,128.43) -- (168.64,134.79);
\definecolor{drawColor}{RGB}{255,165,0}

\path[draw=drawColor,line width= 0.4pt,line join=round,line cap=round] (165.46,119.61) --
	(168.64,122.79) --
	(171.82,119.61) --
	(168.64,116.43) --
	cycle;
\definecolor{drawColor}{RGB}{255,0,0}

\path[draw=drawColor,line width= 0.4pt,line join=round,line cap=round] (166.39,105.36) rectangle (170.89,109.86);

\path[draw=drawColor,line width= 0.4pt,line join=round,line cap=round] (166.39,105.36) -- (170.89,109.86);

\path[draw=drawColor,line width= 0.4pt,line join=round,line cap=round] (166.39,109.86) -- (170.89,105.36);
\definecolor{drawColor}{RGB}{0,0,0}

\node[text=drawColor,anchor=base west,inner sep=0pt, outer sep=0pt, scale=  1.00] at (186.64,152.17) {Pairwise};

\node[text=drawColor,anchor=base west,inner sep=0pt, outer sep=0pt, scale=  1.00] at (186.64,140.17) {Kernel};

\node[text=drawColor,anchor=base west,inner sep=0pt, outer sep=0pt, scale=  1.00] at (186.64,128.17) {HSIC};

\node[text=drawColor,anchor=base west,inner sep=0pt, outer sep=0pt, scale=  1.00] at (186.64,116.17) {ICA};

\node[text=drawColor,anchor=base west,inner sep=0pt, outer sep=0pt, scale=  1.00] at (186.64,104.17) {MMI};
\end{scope}
\end{tikzpicture}

%% file: result_61.tex
\begin{tikzpicture}[x=1pt,y=1pt]
\definecolor{fillColor}{RGB}{255,255,255}
\path[use as bounding box,fill=fillColor,fill opacity=0.00] (0,0) rectangle (252.94,216.81);
\begin{scope}
\path[clip] (  0.00,  0.00) rectangle (252.94,216.81);
\definecolor{drawColor}{RGB}{0,0,0}

\path[draw=drawColor,line width= 0.4pt,line join=round,line cap=round] ( 55.81, 61.20) -- (221.13, 61.20);

\path[draw=drawColor,line width= 0.4pt,line join=round,line cap=round] ( 55.81, 61.20) -- ( 55.81, 55.20);

\path[draw=drawColor,line width= 0.4pt,line join=round,line cap=round] ( 97.14, 61.20) -- ( 97.14, 55.20);

\path[draw=drawColor,line width= 0.4pt,line join=round,line cap=round] (138.47, 61.20) -- (138.47, 55.20);

\path[draw=drawColor,line width= 0.4pt,line join=round,line cap=round] (179.80, 61.20) -- (179.80, 55.20);

\path[draw=drawColor,line width= 0.4pt,line join=round,line cap=round] (221.13, 61.20) -- (221.13, 55.20);

\node[text=drawColor,anchor=base,inner sep=0pt, outer sep=0pt, scale=  1.00] at ( 55.81, 39.60) {100};

\node[text=drawColor,anchor=base,inner sep=0pt, outer sep=0pt, scale=  1.00] at ( 97.14, 39.60) {150};

\node[text=drawColor,anchor=base,inner sep=0pt, outer sep=0pt, scale=  1.00] at (138.47, 39.60) {200};

\node[text=drawColor,anchor=base,inner sep=0pt, outer sep=0pt, scale=  1.00] at (179.80, 39.60) {250};

\node[text=drawColor,anchor=base,inner sep=0pt, outer sep=0pt, scale=  1.00] at (221.13, 39.60) {300};

\path[draw=drawColor,line width= 0.4pt,line join=round,line cap=round] ( 49.20, 65.14) -- ( 49.20,163.67);

\path[draw=drawColor,line width= 0.4pt,line join=round,line cap=round] ( 49.20, 65.14) -- ( 43.20, 65.14);

\path[draw=drawColor,line width= 0.4pt,line join=round,line cap=round] ( 49.20, 84.85) -- ( 43.20, 84.85);

\path[draw=drawColor,line width= 0.4pt,line join=round,line cap=round] ( 49.20,104.55) -- ( 43.20,104.55);

\path[draw=drawColor,line width= 0.4pt,line join=round,line cap=round] ( 49.20,124.26) -- ( 43.20,124.26);

\path[draw=drawColor,line width= 0.4pt,line join=round,line cap=round] ( 49.20,143.96) -- ( 43.20,143.96);

\path[draw=drawColor,line width= 0.4pt,line join=round,line cap=round] ( 49.20,163.67) -- ( 43.20,163.67);

\node[text=drawColor,rotate= 90.00,anchor=base,inner sep=0pt, outer sep=0pt, scale=  1.00] at ( 34.80, 65.14) {0.0};

\node[text=drawColor,rotate= 90.00,anchor=base,inner sep=0pt, outer sep=0pt, scale=  1.00] at ( 34.80,104.55) {0.4};

\node[text=drawColor,rotate= 90.00,anchor=base,inner sep=0pt, outer sep=0pt, scale=  1.00] at ( 34.80,143.96) {0.8};

\path[draw=drawColor,line width= 0.4pt,line join=round,line cap=round] ( 49.20, 61.20) --
	(227.75, 61.20) --
	(227.75,167.61) --
	( 49.20,167.61) --
	cycle;
\end{scope}
\begin{scope}
\path[clip] (  0.00,  0.00) rectangle (252.94,216.81);
\definecolor{drawColor}{RGB}{0,0,0}

\node[text=drawColor,anchor=base,inner sep=0pt, outer sep=0pt, scale=  1.20] at (138.47,188.07) {\bfseries p=15: Criteria B};

\node[text=drawColor,anchor=base,inner sep=0pt, outer sep=0pt, scale=  1.00] at (138.47, 15.60) {Sample Size};

\node[text=drawColor,rotate= 90.00,anchor=base,inner sep=0pt, outer sep=0pt, scale=  1.00] at ( 10.80,114.41) {Error Rate};
\end{scope}
\begin{scope}
\path[clip] ( 49.20, 61.20) rectangle (227.75,167.61);
\definecolor{drawColor}{RGB}{0,0,255}

\path[draw=drawColor,line width= 0.4pt,line join=round,line cap=round] ( 55.81, 99.63) circle (  2.25);

\path[draw=drawColor,line width= 0.4pt,line join=round,line cap=round] (138.47,101.60) circle (  2.25);

\path[draw=drawColor,line width= 0.4pt,line join=round,line cap=round] (221.13, 99.63) circle (  2.25);

\path[draw=drawColor,line width= 0.4pt,line join=round,line cap=round] ( 55.81, 99.63) --
	(138.47,101.60) --
	(221.13, 99.63);
\definecolor{drawColor}{RGB}{0,255,0}

\path[draw=drawColor,line width= 0.4pt,line join=round,line cap=round] ( 55.81, 98.20) --
	( 58.84, 92.95) --
	( 52.78, 92.95) --
	cycle;

\path[draw=drawColor,line width= 0.4pt,line join=round,line cap=round] (138.47, 96.23) --
	(141.50, 90.98) --
	(135.44, 90.98) --
	cycle;

\path[draw=drawColor,line width= 0.4pt,line join=round,line cap=round] (221.13, 95.24) --
	(224.16, 89.99) --
	(218.10, 89.99) --
	cycle;

\path[draw=drawColor,line width= 0.4pt,dash pattern=on 4pt off 4pt ,line join=round,line cap=round] ( 55.81, 94.70) --
	(138.47, 92.73) --
	(221.13, 91.74);
\definecolor{drawColor}{RGB}{0,0,0}

\path[draw=drawColor,line width= 0.4pt,line join=round,line cap=round] ( 52.63, 84.85) -- ( 58.99, 84.85);

\path[draw=drawColor,line width= 0.4pt,line join=round,line cap=round] ( 55.81, 81.66) -- ( 55.81, 88.03);

\path[draw=drawColor,line width= 0.4pt,line join=round,line cap=round] (135.29, 82.88) -- (141.65, 82.88);

\path[draw=drawColor,line width= 0.4pt,line join=round,line cap=round] (138.47, 79.69) -- (138.47, 86.06);

\path[draw=drawColor,line width= 0.4pt,line join=round,line cap=round] (217.95, 82.88) -- (224.31, 82.88);

\path[draw=drawColor,line width= 0.4pt,line join=round,line cap=round] (221.13, 79.69) -- (221.13, 86.06);

\path[draw=drawColor,line width= 0.4pt,dash pattern=on 1pt off 3pt on 4pt off 3pt ,line join=round,line cap=round] ( 55.81, 84.85) --
	(138.47, 82.88) --
	(221.13, 82.88);
\definecolor{drawColor}{RGB}{255,165,0}

\path[draw=drawColor,line width= 0.4pt,line join=round,line cap=round] ( 52.63, 86.82) --
	( 55.81, 90.00) --
	( 58.99, 86.82) --
	( 55.81, 83.64) --
	cycle;

\path[draw=drawColor,line width= 0.4pt,line join=round,line cap=round] (135.29, 85.83) --
	(138.47, 89.01) --
	(141.65, 85.83) --
	(138.47, 82.65) --
	cycle;

\path[draw=drawColor,line width= 0.4pt,line join=round,line cap=round] (217.95, 77.95) --
	(221.13, 81.13) --
	(224.31, 77.95) --
	(221.13, 74.77) --
	cycle;

\path[draw=drawColor,line width= 0.4pt,dash pattern=on 7pt off 3pt ,line join=round,line cap=round] ( 55.81, 86.82) --
	(138.47, 85.83) --
	(221.13, 77.95);
\definecolor{drawColor}{RGB}{255,0,0}

\path[draw=drawColor,line width= 0.4pt,line join=round,line cap=round] ( 53.56, 68.80) rectangle ( 58.06, 73.30);

\path[draw=drawColor,line width= 0.4pt,line join=round,line cap=round] ( 53.56, 68.80) -- ( 58.06, 73.30);

\path[draw=drawColor,line width= 0.4pt,line join=round,line cap=round] ( 53.56, 73.30) -- ( 58.06, 68.80);

\path[draw=drawColor,line width= 0.4pt,line join=round,line cap=round] (136.22, 68.80) rectangle (140.72, 73.30);

\path[draw=drawColor,line width= 0.4pt,line join=round,line cap=round] (136.22, 68.80) -- (140.72, 73.30);

\path[draw=drawColor,line width= 0.4pt,line join=round,line cap=round] (136.22, 73.30) -- (140.72, 68.80);

\path[draw=drawColor,line width= 0.4pt,line join=round,line cap=round] (218.88, 68.80) rectangle (223.38, 73.30);

\path[draw=drawColor,line width= 0.4pt,line join=round,line cap=round] (218.88, 68.80) -- (223.38, 73.30);

\path[draw=drawColor,line width= 0.4pt,line join=round,line cap=round] (218.88, 73.30) -- (223.38, 68.80);

\path[draw=drawColor,line width= 0.4pt,line join=round,line cap=round] ( 55.81, 71.05) --
	(138.47, 71.05) --
	(221.13, 71.05);
\definecolor{drawColor}{RGB}{0,0,0}

\path[draw=drawColor,line width= 0.4pt,line join=round,line cap=round] (156.94,167.61) rectangle (227.75, 95.61);
\definecolor{drawColor}{RGB}{0,0,255}

\path[draw=drawColor,line width= 0.4pt,line join=round,line cap=round] (159.64,155.61) -- (177.64,155.61);
\definecolor{drawColor}{RGB}{0,255,0}

\path[draw=drawColor,line width= 0.4pt,dash pattern=on 4pt off 4pt ,line join=round,line cap=round] (159.64,143.61) -- (177.64,143.61);
\definecolor{drawColor}{RGB}{0,0,0}

\path[draw=drawColor,line width= 0.4pt,dash pattern=on 1pt off 3pt on 4pt off 3pt ,line join=round,line cap=round] (159.64,131.61) -- (177.64,131.61);
\definecolor{drawColor}{RGB}{255,165,0}

\path[draw=drawColor,line width= 0.4pt,dash pattern=on 7pt off 3pt ,line join=round,line cap=round] (159.64,119.61) -- (177.64,119.61);
\definecolor{drawColor}{RGB}{255,0,0}

\path[draw=drawColor,line width= 0.4pt,line join=round,line cap=round] (159.64,107.61) -- (177.64,107.61);
\definecolor{drawColor}{RGB}{0,0,255}

\path[draw=drawColor,line width= 0.4pt,line join=round,line cap=round] (168.64,155.61) circle (  2.25);
\definecolor{drawColor}{RGB}{0,255,0}

\path[draw=drawColor,line width= 0.4pt,line join=round,line cap=round] (168.64,147.11) --
	(171.67,141.86) --
	(165.61,141.86) --
	cycle;
\definecolor{drawColor}{RGB}{0,0,0}

\path[draw=drawColor,line width= 0.4pt,line join=round,line cap=round] (165.46,131.61) -- (171.82,131.61);

\path[draw=drawColor,line width= 0.4pt,line join=round,line cap=round] (168.64,128.43) -- (168.64,134.79);
\definecolor{drawColor}{RGB}{255,165,0}

\path[draw=drawColor,line width= 0.4pt,line join=round,line cap=round] (165.46,119.61) --
	(168.64,122.79) --
	(171.82,119.61) --
	(168.64,116.43) --
	cycle;
\definecolor{drawColor}{RGB}{255,0,0}

\path[draw=drawColor,line width= 0.4pt,line join=round,line cap=round] (166.39,105.36) rectangle (170.89,109.86);

\path[draw=drawColor,line width= 0.4pt,line join=round,line cap=round] (166.39,105.36) -- (170.89,109.86);

\path[draw=drawColor,line width= 0.4pt,line join=round,line cap=round] (166.39,109.86) -- (170.89,105.36);
\definecolor{drawColor}{RGB}{0,0,0}

\node[text=drawColor,anchor=base west,inner sep=0pt, outer sep=0pt, scale=  1.00] at (186.64,152.17) {Pairwise};

\node[text=drawColor,anchor=base west,inner sep=0pt, outer sep=0pt, scale=  1.00] at (186.64,140.17) {Kernel};

\node[text=drawColor,anchor=base west,inner sep=0pt, outer sep=0pt, scale=  1.00] at (186.64,128.17) {HSIC};

\node[text=drawColor,anchor=base west,inner sep=0pt, outer sep=0pt, scale=  1.00] at (186.64,116.17) {ICA};

\node[text=drawColor,anchor=base west,inner sep=0pt, outer sep=0pt, scale=  1.00] at (186.64,104.17) {MMI};
\end{scope}
\end{tikzpicture}

%% file: result_71.tex
\begin{tikzpicture}[x=1pt,y=1pt]
\definecolor{fillColor}{RGB}{255,255,255}
\path[use as bounding box,fill=fillColor,fill opacity=0.00] (0,0) rectangle (252.94,216.81);
\begin{scope}
\path[clip] (  0.00,  0.00) rectangle (252.94,216.81);
\definecolor{drawColor}{RGB}{0,0,0}

\path[draw=drawColor,line width= 0.4pt,line join=round,line cap=round] ( 74.18, 61.20) -- (221.13, 61.20);

\path[draw=drawColor,line width= 0.4pt,line join=round,line cap=round] ( 74.18, 61.20) -- ( 74.18, 55.20);

\path[draw=drawColor,line width= 0.4pt,line join=round,line cap=round] (110.92, 61.20) -- (110.92, 55.20);

\path[draw=drawColor,line width= 0.4pt,line join=round,line cap=round] (147.66, 61.20) -- (147.66, 55.20);

\path[draw=drawColor,line width= 0.4pt,line join=round,line cap=round] (184.39, 61.20) -- (184.39, 55.20);

\path[draw=drawColor,line width= 0.4pt,line join=round,line cap=round] (221.13, 61.20) -- (221.13, 55.20);

\node[text=drawColor,anchor=base,inner sep=0pt, outer sep=0pt, scale=  1.00] at ( 74.18, 39.60) {200};

\node[text=drawColor,anchor=base,inner sep=0pt, outer sep=0pt, scale=  1.00] at (110.92, 39.60) {400};

\node[text=drawColor,anchor=base,inner sep=0pt, outer sep=0pt, scale=  1.00] at (147.66, 39.60) {600};

\node[text=drawColor,anchor=base,inner sep=0pt, outer sep=0pt, scale=  1.00] at (184.39, 39.60) {800};

\node[text=drawColor,anchor=base,inner sep=0pt, outer sep=0pt, scale=  1.00] at (221.13, 39.60) {1000};

\path[draw=drawColor,line width= 0.4pt,line join=round,line cap=round] ( 49.20, 65.14) -- ( 49.20,163.67);

\path[draw=drawColor,line width= 0.4pt,line join=round,line cap=round] ( 49.20, 65.14) -- ( 43.20, 65.14);

\path[draw=drawColor,line width= 0.4pt,line join=round,line cap=round] ( 49.20, 84.85) -- ( 43.20, 84.85);

\path[draw=drawColor,line width= 0.4pt,line join=round,line cap=round] ( 49.20,104.55) -- ( 43.20,104.55);

\path[draw=drawColor,line width= 0.4pt,line join=round,line cap=round] ( 49.20,124.26) -- ( 43.20,124.26);

\path[draw=drawColor,line width= 0.4pt,line join=round,line cap=round] ( 49.20,143.96) -- ( 43.20,143.96);

\path[draw=drawColor,line width= 0.4pt,line join=round,line cap=round] ( 49.20,163.67) -- ( 43.20,163.67);

\node[text=drawColor,rotate= 90.00,anchor=base,inner sep=0pt, outer sep=0pt, scale=  1.00] at ( 34.80, 65.14) {0.0};

\node[text=drawColor,rotate= 90.00,anchor=base,inner sep=0pt, outer sep=0pt, scale=  1.00] at ( 34.80,104.55) {0.4};

\node[text=drawColor,rotate= 90.00,anchor=base,inner sep=0pt, outer sep=0pt, scale=  1.00] at ( 34.80,143.96) {0.8};

\path[draw=drawColor,line width= 0.4pt,line join=round,line cap=round] ( 49.20, 61.20) --
	(227.75, 61.20) --
	(227.75,167.61) --
	( 49.20,167.61) --
	cycle;
\end{scope}
\begin{scope}
\path[clip] (  0.00,  0.00) rectangle (252.94,216.81);
\definecolor{drawColor}{RGB}{0,0,0}

\node[text=drawColor,anchor=base,inner sep=0pt, outer sep=0pt, scale=  1.20] at (138.47,188.07) {\bfseries p=30: Criteria A};

\node[text=drawColor,anchor=base,inner sep=0pt, outer sep=0pt, scale=  1.00] at (138.47, 15.60) {Sample Size};

\node[text=drawColor,rotate= 90.00,anchor=base,inner sep=0pt, outer sep=0pt, scale=  1.00] at ( 10.80,114.41) {Error Rate};
\end{scope}
\begin{scope}
\path[clip] ( 49.20, 61.20) rectangle (227.75,167.61);
\definecolor{drawColor}{RGB}{0,0,255}

\path[draw=drawColor,line width= 0.4pt,line join=round,line cap=round] ( 55.81,163.67) circle (  2.25);

\path[draw=drawColor,line width= 0.4pt,line join=round,line cap=round] ( 74.18,163.67) circle (  2.25);

\path[draw=drawColor,line width= 0.4pt,line join=round,line cap=round] ( 92.55,163.67) circle (  2.25);

\path[draw=drawColor,line width= 0.4pt,line join=round,line cap=round] (129.29,160.71) circle (  2.25);

\path[draw=drawColor,line width= 0.4pt,line join=round,line cap=round] (166.03,155.79) circle (  2.25);

\path[draw=drawColor,line width= 0.4pt,line join=round,line cap=round] (221.13,153.82) circle (  2.25);

\path[draw=drawColor,line width= 0.4pt,line join=round,line cap=round] ( 55.81,163.67) --
	( 74.18,163.67) --
	( 92.55,163.67) --
	(129.29,160.71) --
	(166.03,155.79) --
	(221.13,153.82);
\definecolor{drawColor}{RGB}{0,0,0}

\path[draw=drawColor,line width= 0.4pt,line join=round,line cap=round] ( 52.63,134.11) -- ( 58.99,134.11);

\path[draw=drawColor,line width= 0.4pt,line join=round,line cap=round] ( 55.81,130.93) -- ( 55.81,137.29);

\path[draw=drawColor,line width= 0.4pt,line join=round,line cap=round] ( 71.00, 84.85) -- ( 77.36, 84.85);

\path[draw=drawColor,line width= 0.4pt,line join=round,line cap=round] ( 74.18, 81.66) -- ( 74.18, 88.03);

\path[draw=drawColor,line width= 0.4pt,line join=round,line cap=round] ( 89.37, 74.99) -- ( 95.73, 74.99);

\path[draw=drawColor,line width= 0.4pt,line join=round,line cap=round] ( 92.55, 71.81) -- ( 92.55, 78.18);

\path[draw=drawColor,line width= 0.4pt,line join=round,line cap=round] (126.11, 65.14) -- (132.47, 65.14);

\path[draw=drawColor,line width= 0.4pt,line join=round,line cap=round] (129.29, 61.96) -- (129.29, 68.32);

\path[draw=drawColor,line width= 0.4pt,line join=round,line cap=round] (162.84, 65.14) -- (169.21, 65.14);

\path[draw=drawColor,line width= 0.4pt,line join=round,line cap=round] (166.03, 61.96) -- (166.03, 68.32);

\path[draw=drawColor,line width= 0.4pt,line join=round,line cap=round] (217.95, 65.14) -- (224.31, 65.14);

\path[draw=drawColor,line width= 0.4pt,line join=round,line cap=round] (221.13, 61.96) -- (221.13, 68.32);

\path[draw=drawColor,line width= 0.4pt,dash pattern=on 1pt off 3pt on 4pt off 3pt ,line join=round,line cap=round] ( 55.81,134.11) --
	( 74.18, 84.85) --
	( 92.55, 74.99) --
	(129.29, 65.14) --
	(166.03, 65.14) --
	(221.13, 65.14);
\definecolor{drawColor}{RGB}{255,165,0}

\path[draw=drawColor,line width= 0.4pt,line join=round,line cap=round] ( 52.63,163.67) --
	( 55.81,166.85) --
	( 58.99,163.67) --
	( 55.81,160.49) --
	cycle;

\path[draw=drawColor,line width= 0.4pt,line join=round,line cap=round] ( 71.00,163.67) --
	( 74.18,166.85) --
	( 77.36,163.67) --
	( 74.18,160.49) --
	cycle;

\path[draw=drawColor,line width= 0.4pt,line join=round,line cap=round] ( 89.37,163.67) --
	( 92.55,166.85) --
	( 95.73,163.67) --
	( 92.55,160.49) --
	cycle;

\path[draw=drawColor,line width= 0.4pt,line join=round,line cap=round] (126.11,160.71) --
	(129.29,163.90) --
	(132.47,160.71) --
	(129.29,157.53) --
	cycle;

\path[draw=drawColor,line width= 0.4pt,line join=round,line cap=round] (162.84,134.11) --
	(166.03,137.29) --
	(169.21,134.11) --
	(166.03,130.93) --
	cycle;

\path[draw=drawColor,line width= 0.4pt,line join=round,line cap=round] (217.95, 94.70) --
	(221.13, 97.88) --
	(224.31, 94.70) --
	(221.13, 91.52) --
	cycle;

\path[draw=drawColor,line width= 0.4pt,dash pattern=on 7pt off 3pt ,line join=round,line cap=round] ( 55.81,163.67) --
	( 74.18,163.67) --
	( 92.55,163.67) --
	(129.29,160.71) --
	(166.03,134.11) --
	(221.13, 94.70);
\definecolor{drawColor}{RGB}{255,0,0}

\path[draw=drawColor,line width= 0.4pt,line join=round,line cap=round] ( 53.56,112.15) rectangle ( 58.06,116.65);

\path[draw=drawColor,line width= 0.4pt,line join=round,line cap=round] ( 53.56,112.15) -- ( 58.06,116.65);

\path[draw=drawColor,line width= 0.4pt,line join=round,line cap=round] ( 53.56,116.65) -- ( 58.06,112.15);

\path[draw=drawColor,line width= 0.4pt,line join=round,line cap=round] ( 71.93, 70.77) rectangle ( 76.43, 75.27);

\path[draw=drawColor,line width= 0.4pt,line join=round,line cap=round] ( 71.93, 70.77) -- ( 76.43, 75.27);

\path[draw=drawColor,line width= 0.4pt,line join=round,line cap=round] ( 71.93, 75.27) -- ( 76.43, 70.77);

\path[draw=drawColor,line width= 0.4pt,line join=round,line cap=round] ( 90.30, 63.88) rectangle ( 94.80, 68.38);

\path[draw=drawColor,line width= 0.4pt,line join=round,line cap=round] ( 90.30, 63.88) -- ( 94.80, 68.38);

\path[draw=drawColor,line width= 0.4pt,line join=round,line cap=round] ( 90.30, 68.38) -- ( 94.80, 63.88);

\path[draw=drawColor,line width= 0.4pt,line join=round,line cap=round] (127.04, 62.89) rectangle (131.54, 67.39);

\path[draw=drawColor,line width= 0.4pt,line join=round,line cap=round] (127.04, 62.89) -- (131.54, 67.39);

\path[draw=drawColor,line width= 0.4pt,line join=round,line cap=round] (127.04, 67.39) -- (131.54, 62.89);

\path[draw=drawColor,line width= 0.4pt,line join=round,line cap=round] (163.78, 62.89) rectangle (168.28, 67.39);

\path[draw=drawColor,line width= 0.4pt,line join=round,line cap=round] (163.78, 62.89) -- (168.28, 67.39);

\path[draw=drawColor,line width= 0.4pt,line join=round,line cap=round] (163.78, 67.39) -- (168.28, 62.89);

\path[draw=drawColor,line width= 0.4pt,line join=round,line cap=round] (218.88, 62.89) rectangle (223.38, 67.39);

\path[draw=drawColor,line width= 0.4pt,line join=round,line cap=round] (218.88, 62.89) -- (223.38, 67.39);

\path[draw=drawColor,line width= 0.4pt,line join=round,line cap=round] (218.88, 67.39) -- (223.38, 62.89);

\path[draw=drawColor,line width= 0.4pt,line join=round,line cap=round] ( 55.81,114.40) --
	( 74.18, 73.02) --
	( 92.55, 66.13) --
	(129.29, 65.14) --
	(166.03, 65.14) --
	(221.13, 65.14);
\definecolor{drawColor}{RGB}{0,0,0}

\path[draw=drawColor,line width= 0.4pt,line join=round,line cap=round] (156.94,167.61) rectangle (227.75,107.61);
\definecolor{drawColor}{RGB}{0,0,255}

\path[draw=drawColor,line width= 0.4pt,line join=round,line cap=round] (159.64,155.61) -- (177.64,155.61);
\definecolor{drawColor}{RGB}{0,0,0}

\path[draw=drawColor,line width= 0.4pt,dash pattern=on 1pt off 3pt on 4pt off 3pt ,line join=round,line cap=round] (159.64,143.61) -- (177.64,143.61);
\definecolor{drawColor}{RGB}{255,165,0}

\path[draw=drawColor,line width= 0.4pt,dash pattern=on 7pt off 3pt ,line join=round,line cap=round] (159.64,131.61) -- (177.64,131.61);
\definecolor{drawColor}{RGB}{255,0,0}

\path[draw=drawColor,line width= 0.4pt,line join=round,line cap=round] (159.64,119.61) -- (177.64,119.61);
\definecolor{drawColor}{RGB}{0,0,255}

\path[draw=drawColor,line width= 0.4pt,line join=round,line cap=round] (168.64,155.61) circle (  2.25);
\definecolor{drawColor}{RGB}{0,0,0}

\path[draw=drawColor,line width= 0.4pt,line join=round,line cap=round] (165.46,143.61) -- (171.82,143.61);

\path[draw=drawColor,line width= 0.4pt,line join=round,line cap=round] (168.64,140.43) -- (168.64,146.79);
\definecolor{drawColor}{RGB}{255,165,0}

\path[draw=drawColor,line width= 0.4pt,line join=round,line cap=round] (165.46,131.61) --
	(168.64,134.79) --
	(171.82,131.61) --
	(168.64,128.43) --
	cycle;
\definecolor{drawColor}{RGB}{255,0,0}

\path[draw=drawColor,line width= 0.4pt,line join=round,line cap=round] (166.39,117.36) rectangle (170.89,121.86);

\path[draw=drawColor,line width= 0.4pt,line join=round,line cap=round] (166.39,117.36) -- (170.89,121.86);

\path[draw=drawColor,line width= 0.4pt,line join=round,line cap=round] (166.39,121.86) -- (170.89,117.36);
\definecolor{drawColor}{RGB}{0,0,0}

\node[text=drawColor,anchor=base west,inner sep=0pt, outer sep=0pt, scale=  1.00] at (186.64,152.17) {Pairwise};

\node[text=drawColor,anchor=base west,inner sep=0pt, outer sep=0pt, scale=  1.00] at (186.64,140.17) {HSIC};

\node[text=drawColor,anchor=base west,inner sep=0pt, outer sep=0pt, scale=  1.00] at (186.64,128.17) {ICA};

\node[text=drawColor,anchor=base west,inner sep=0pt, outer sep=0pt, scale=  1.00] at (186.64,116.17) {MMI};
\end{scope}
\end{tikzpicture}

%% file: result_81.tex
\begin{tikzpicture}[x=1pt,y=1pt]
\definecolor{fillColor}{RGB}{255,255,255}
\path[use as bounding box,fill=fillColor,fill opacity=0.00] (0,0) rectangle (252.94,216.81);
\begin{scope}
\path[clip] (  0.00,  0.00) rectangle (252.94,216.81);
\definecolor{drawColor}{RGB}{0,0,0}

\path[draw=drawColor,line width= 0.4pt,line join=round,line cap=round] ( 74.18, 61.20) -- (221.13, 61.20);

\path[draw=drawColor,line width= 0.4pt,line join=round,line cap=round] ( 74.18, 61.20) -- ( 74.18, 55.20);

\path[draw=drawColor,line width= 0.4pt,line join=round,line cap=round] (110.92, 61.20) -- (110.92, 55.20);

\path[draw=drawColor,line width= 0.4pt,line join=round,line cap=round] (147.66, 61.20) -- (147.66, 55.20);

\path[draw=drawColor,line width= 0.4pt,line join=round,line cap=round] (184.39, 61.20) -- (184.39, 55.20);

\path[draw=drawColor,line width= 0.4pt,line join=round,line cap=round] (221.13, 61.20) -- (221.13, 55.20);

\node[text=drawColor,anchor=base,inner sep=0pt, outer sep=0pt, scale=  1.00] at ( 74.18, 39.60) {200};

\node[text=drawColor,anchor=base,inner sep=0pt, outer sep=0pt, scale=  1.00] at (110.92, 39.60) {400};

\node[text=drawColor,anchor=base,inner sep=0pt, outer sep=0pt, scale=  1.00] at (147.66, 39.60) {600};

\node[text=drawColor,anchor=base,inner sep=0pt, outer sep=0pt, scale=  1.00] at (184.39, 39.60) {800};

\node[text=drawColor,anchor=base,inner sep=0pt, outer sep=0pt, scale=  1.00] at (221.13, 39.60) {1000};

\path[draw=drawColor,line width= 0.4pt,line join=round,line cap=round] ( 49.20, 65.14) -- ( 49.20,163.67);

\path[draw=drawColor,line width= 0.4pt,line join=round,line cap=round] ( 49.20, 65.14) -- ( 43.20, 65.14);

\path[draw=drawColor,line width= 0.4pt,line join=round,line cap=round] ( 49.20, 84.85) -- ( 43.20, 84.85);

\path[draw=drawColor,line width= 0.4pt,line join=round,line cap=round] ( 49.20,104.55) -- ( 43.20,104.55);

\path[draw=drawColor,line width= 0.4pt,line join=round,line cap=round] ( 49.20,124.26) -- ( 43.20,124.26);

\path[draw=drawColor,line width= 0.4pt,line join=round,line cap=round] ( 49.20,143.96) -- ( 43.20,143.96);

\path[draw=drawColor,line width= 0.4pt,line join=round,line cap=round] ( 49.20,163.67) -- ( 43.20,163.67);

\node[text=drawColor,rotate= 90.00,anchor=base,inner sep=0pt, outer sep=0pt, scale=  1.00] at ( 34.80, 65.14) {0.0};

\node[text=drawColor,rotate= 90.00,anchor=base,inner sep=0pt, outer sep=0pt, scale=  1.00] at ( 34.80,104.55) {0.4};

\node[text=drawColor,rotate= 90.00,anchor=base,inner sep=0pt, outer sep=0pt, scale=  1.00] at ( 34.80,143.96) {0.8};

\path[draw=drawColor,line width= 0.4pt,line join=round,line cap=round] ( 49.20, 61.20) --
	(227.75, 61.20) --
	(227.75,167.61) --
	( 49.20,167.61) --
	cycle;
\end{scope}
\begin{scope}
\path[clip] (  0.00,  0.00) rectangle (252.94,216.81);
\definecolor{drawColor}{RGB}{0,0,0}

\node[text=drawColor,anchor=base,inner sep=0pt, outer sep=0pt, scale=  1.20] at (138.47,188.07) {\bfseries p=30: Criteria B};

\node[text=drawColor,anchor=base,inner sep=0pt, outer sep=0pt, scale=  1.00] at (138.47, 15.60) {Sample Size};

\node[text=drawColor,rotate= 90.00,anchor=base,inner sep=0pt, outer sep=0pt, scale=  1.00] at ( 10.80,114.41) {Error Rate};
\end{scope}
\begin{scope}
\path[clip] ( 49.20, 61.20) rectangle (227.75,167.61);
\definecolor{drawColor}{RGB}{0,0,255}

\path[draw=drawColor,line width= 0.4pt,line join=round,line cap=round] ( 55.81,109.48) circle (  2.25);

\path[draw=drawColor,line width= 0.4pt,line join=round,line cap=round] ( 74.18,103.57) circle (  2.25);

\path[draw=drawColor,line width= 0.4pt,line join=round,line cap=round] ( 92.55,100.61) circle (  2.25);

\path[draw=drawColor,line width= 0.4pt,line join=round,line cap=round] (129.29, 99.63) circle (  2.25);

\path[draw=drawColor,line width= 0.4pt,line join=round,line cap=round] (166.03, 96.67) circle (  2.25);

\path[draw=drawColor,line width= 0.4pt,line join=round,line cap=round] (221.13, 94.70) circle (  2.25);

\path[draw=drawColor,line width= 0.4pt,line join=round,line cap=round] ( 55.81,109.48) --
	( 74.18,103.57) --
	( 92.55,100.61) --
	(129.29, 99.63) --
	(166.03, 96.67) --
	(221.13, 94.70);
\definecolor{drawColor}{RGB}{0,0,0}

\path[draw=drawColor,line width= 0.4pt,line join=round,line cap=round] ( 52.63, 67.11) -- ( 58.99, 67.11);

\path[draw=drawColor,line width= 0.4pt,line join=round,line cap=round] ( 55.81, 63.93) -- ( 55.81, 70.29);

\path[draw=drawColor,line width= 0.4pt,line join=round,line cap=round] ( 71.00, 67.11) -- ( 77.36, 67.11);

\path[draw=drawColor,line width= 0.4pt,line join=round,line cap=round] ( 74.18, 63.93) -- ( 74.18, 70.29);

\path[draw=drawColor,line width= 0.4pt,line join=round,line cap=round] ( 89.37, 65.14) -- ( 95.73, 65.14);

\path[draw=drawColor,line width= 0.4pt,line join=round,line cap=round] ( 92.55, 61.96) -- ( 92.55, 68.32);

\path[draw=drawColor,line width= 0.4pt,line join=round,line cap=round] (126.11, 65.14) -- (132.47, 65.14);

\path[draw=drawColor,line width= 0.4pt,line join=round,line cap=round] (129.29, 61.96) -- (129.29, 68.32);

\path[draw=drawColor,line width= 0.4pt,line join=round,line cap=round] (162.84, 65.14) -- (169.21, 65.14);

\path[draw=drawColor,line width= 0.4pt,line join=round,line cap=round] (166.03, 61.96) -- (166.03, 68.32);

\path[draw=drawColor,line width= 0.4pt,line join=round,line cap=round] (217.95, 65.14) -- (224.31, 65.14);

\path[draw=drawColor,line width= 0.4pt,line join=round,line cap=round] (221.13, 61.96) -- (221.13, 68.32);

\path[draw=drawColor,line width= 0.4pt,dash pattern=on 1pt off 3pt on 4pt off 3pt ,line join=round,line cap=round] ( 55.81, 67.11) --
	( 74.18, 67.11) --
	( 92.55, 65.14) --
	(129.29, 65.14) --
	(166.03, 65.14) --
	(221.13, 65.14);
\definecolor{drawColor}{RGB}{255,165,0}

\path[draw=drawColor,line width= 0.4pt,line join=round,line cap=round] ( 52.63, 88.79) --
	( 55.81, 91.97) --
	( 58.99, 88.79) --
	( 55.81, 85.61) --
	cycle;

\path[draw=drawColor,line width= 0.4pt,line join=round,line cap=round] ( 71.00, 91.74) --
	( 74.18, 94.93) --
	( 77.36, 91.74) --
	( 74.18, 88.56) --
	cycle;

\path[draw=drawColor,line width= 0.4pt,line join=round,line cap=round] ( 89.37, 84.85) --
	( 92.55, 88.03) --
	( 95.73, 84.85) --
	( 92.55, 81.66) --
	cycle;

\path[draw=drawColor,line width= 0.4pt,line join=round,line cap=round] (126.11, 80.91) --
	(129.29, 84.09) --
	(132.47, 80.91) --
	(129.29, 77.72) --
	cycle;

\path[draw=drawColor,line width= 0.4pt,line join=round,line cap=round] (162.84, 73.02) --
	(166.03, 76.21) --
	(169.21, 73.02) --
	(166.03, 69.84) --
	cycle;

\path[draw=drawColor,line width= 0.4pt,line join=round,line cap=round] (217.95, 65.14) --
	(221.13, 68.32) --
	(224.31, 65.14) --
	(221.13, 61.96) --
	cycle;

\path[draw=drawColor,line width= 0.4pt,dash pattern=on 7pt off 3pt ,line join=round,line cap=round] ( 55.81, 88.79) --
	( 74.18, 91.74) --
	( 92.55, 84.85) --
	(129.29, 80.91) --
	(166.03, 73.02) --
	(221.13, 65.14);
\definecolor{drawColor}{RGB}{255,0,0}

\path[draw=drawColor,line width= 0.4pt,line join=round,line cap=round] ( 53.56, 63.88) rectangle ( 58.06, 68.38);

\path[draw=drawColor,line width= 0.4pt,line join=round,line cap=round] ( 53.56, 63.88) -- ( 58.06, 68.38);

\path[draw=drawColor,line width= 0.4pt,line join=round,line cap=round] ( 53.56, 68.38) -- ( 58.06, 63.88);

\path[draw=drawColor,line width= 0.4pt,line join=round,line cap=round] ( 71.93, 63.88) rectangle ( 76.43, 68.38);

\path[draw=drawColor,line width= 0.4pt,line join=round,line cap=round] ( 71.93, 63.88) -- ( 76.43, 68.38);

\path[draw=drawColor,line width= 0.4pt,line join=round,line cap=round] ( 71.93, 68.38) -- ( 76.43, 63.88);

\path[draw=drawColor,line width= 0.4pt,line join=round,line cap=round] ( 90.30, 63.88) rectangle ( 94.80, 68.38);

\path[draw=drawColor,line width= 0.4pt,line join=round,line cap=round] ( 90.30, 63.88) -- ( 94.80, 68.38);

\path[draw=drawColor,line width= 0.4pt,line join=round,line cap=round] ( 90.30, 68.38) -- ( 94.80, 63.88);

\path[draw=drawColor,line width= 0.4pt,line join=round,line cap=round] (127.04, 63.88) rectangle (131.54, 68.38);

\path[draw=drawColor,line width= 0.4pt,line join=round,line cap=round] (127.04, 63.88) -- (131.54, 68.38);

\path[draw=drawColor,line width= 0.4pt,line join=round,line cap=round] (127.04, 68.38) -- (131.54, 63.88);

\path[draw=drawColor,line width= 0.4pt,line join=round,line cap=round] (163.78, 63.88) rectangle (168.28, 68.38);

\path[draw=drawColor,line width= 0.4pt,line join=round,line cap=round] (163.78, 63.88) -- (168.28, 68.38);

\path[draw=drawColor,line width= 0.4pt,line join=round,line cap=round] (163.78, 68.38) -- (168.28, 63.88);

\path[draw=drawColor,line width= 0.4pt,line join=round,line cap=round] (218.88, 63.88) rectangle (223.38, 68.38);

\path[draw=drawColor,line width= 0.4pt,line join=round,line cap=round] (218.88, 63.88) -- (223.38, 68.38);

\path[draw=drawColor,line width= 0.4pt,line join=round,line cap=round] (218.88, 68.38) -- (223.38, 63.88);

\path[draw=drawColor,line width= 0.4pt,line join=round,line cap=round] ( 55.81, 66.13) --
	( 74.18, 66.13) --
	( 92.55, 66.13) --
	(129.29, 66.13) --
	(166.03, 66.13) --
	(221.13, 66.13);
\definecolor{drawColor}{RGB}{0,0,0}

\path[draw=drawColor,line width= 0.4pt,line join=round,line cap=round] (156.94,167.61) rectangle (227.75,107.61);
\definecolor{drawColor}{RGB}{0,0,255}

\path[draw=drawColor,line width= 0.4pt,line join=round,line cap=round] (159.64,155.61) -- (177.64,155.61);
\definecolor{drawColor}{RGB}{0,0,0}

\path[draw=drawColor,line width= 0.4pt,dash pattern=on 1pt off 3pt on 4pt off 3pt ,line join=round,line cap=round] (159.64,143.61) -- (177.64,143.61);
\definecolor{drawColor}{RGB}{255,165,0}

\path[draw=drawColor,line width= 0.4pt,dash pattern=on 7pt off 3pt ,line join=round,line cap=round] (159.64,131.61) -- (177.64,131.61);
\definecolor{drawColor}{RGB}{255,0,0}

\path[draw=drawColor,line width= 0.4pt,line join=round,line cap=round] (159.64,119.61) -- (177.64,119.61);
\definecolor{drawColor}{RGB}{0,0,255}

\path[draw=drawColor,line width= 0.4pt,line join=round,line cap=round] (168.64,155.61) circle (  2.25);
\definecolor{drawColor}{RGB}{0,0,0}

\path[draw=drawColor,line width= 0.4pt,line join=round,line cap=round] (165.46,143.61) -- (171.82,143.61);

\path[draw=drawColor,line width= 0.4pt,line join=round,line cap=round] (168.64,140.43) -- (168.64,146.79);
\definecolor{drawColor}{RGB}{255,165,0}

\path[draw=drawColor,line width= 0.4pt,line join=round,line cap=round] (165.46,131.61) --
	(168.64,134.79) --
	(171.82,131.61) --
	(168.64,128.43) --
	cycle;
\definecolor{drawColor}{RGB}{255,0,0}

\path[draw=drawColor,line width= 0.4pt,line join=round,line cap=round] (166.39,117.36) rectangle (170.89,121.86);

\path[draw=drawColor,line width= 0.4pt,line join=round,line cap=round] (166.39,117.36) -- (170.89,121.86);

\path[draw=drawColor,line width= 0.4pt,line join=round,line cap=round] (166.39,121.86) -- (170.89,117.36);
\definecolor{drawColor}{RGB}{0,0,0}

\node[text=drawColor,anchor=base west,inner sep=0pt, outer sep=0pt, scale=  1.00] at (186.64,152.17) {Pairwise};

\node[text=drawColor,anchor=base west,inner sep=0pt, outer sep=0pt, scale=  1.00] at (186.64,140.17) {HSIC};

\node[text=drawColor,anchor=base west,inner sep=0pt, outer sep=0pt, scale=  1.00] at (186.64,128.17) {ICA};

\node[text=drawColor,anchor=base west,inner sep=0pt, outer sep=0pt, scale=  1.00] at (186.64,116.17) {MMI};
\end{scope}
\end{tikzpicture}

%% file: result_91.tex
\begin{tikzpicture}[x=1pt,y=1pt]
\definecolor{fillColor}{RGB}{255,255,255}
\path[use as bounding box,fill=fillColor,fill opacity=0.00] (0,0) rectangle (252.94,216.81);
\begin{scope}
\path[clip] (  0.00,  0.00) rectangle (252.94,216.81);
\definecolor{drawColor}{RGB}{0,0,0}

\path[draw=drawColor,line width= 0.4pt,line join=round,line cap=round] ( 74.18, 61.20) -- (221.13, 61.20);

\path[draw=drawColor,line width= 0.4pt,line join=round,line cap=round] ( 74.18, 61.20) -- ( 74.18, 55.20);

\path[draw=drawColor,line width= 0.4pt,line join=round,line cap=round] (110.92, 61.20) -- (110.92, 55.20);

\path[draw=drawColor,line width= 0.4pt,line join=round,line cap=round] (147.66, 61.20) -- (147.66, 55.20);

\path[draw=drawColor,line width= 0.4pt,line join=round,line cap=round] (184.39, 61.20) -- (184.39, 55.20);

\path[draw=drawColor,line width= 0.4pt,line join=round,line cap=round] (221.13, 61.20) -- (221.13, 55.20);

\node[text=drawColor,anchor=base,inner sep=0pt, outer sep=0pt, scale=  1.00] at ( 74.18, 39.60) {200};

\node[text=drawColor,anchor=base,inner sep=0pt, outer sep=0pt, scale=  1.00] at (110.92, 39.60) {400};

\node[text=drawColor,anchor=base,inner sep=0pt, outer sep=0pt, scale=  1.00] at (147.66, 39.60) {600};

\node[text=drawColor,anchor=base,inner sep=0pt, outer sep=0pt, scale=  1.00] at (184.39, 39.60) {800};

\node[text=drawColor,anchor=base,inner sep=0pt, outer sep=0pt, scale=  1.00] at (221.13, 39.60) {1000};

\path[draw=drawColor,line width= 0.4pt,line join=round,line cap=round] ( 49.20, 65.14) -- ( 49.20,163.67);

\path[draw=drawColor,line width= 0.4pt,line join=round,line cap=round] ( 49.20, 65.14) -- ( 43.20, 65.14);

\path[draw=drawColor,line width= 0.4pt,line join=round,line cap=round] ( 49.20, 84.85) -- ( 43.20, 84.85);

\path[draw=drawColor,line width= 0.4pt,line join=round,line cap=round] ( 49.20,104.55) -- ( 43.20,104.55);

\path[draw=drawColor,line width= 0.4pt,line join=round,line cap=round] ( 49.20,124.26) -- ( 43.20,124.26);

\path[draw=drawColor,line width= 0.4pt,line join=round,line cap=round] ( 49.20,143.96) -- ( 43.20,143.96);

\path[draw=drawColor,line width= 0.4pt,line join=round,line cap=round] ( 49.20,163.67) -- ( 43.20,163.67);

\node[text=drawColor,rotate= 90.00,anchor=base,inner sep=0pt, outer sep=0pt, scale=  1.00] at ( 34.80, 65.14) {0.0};

\node[text=drawColor,rotate= 90.00,anchor=base,inner sep=0pt, outer sep=0pt, scale=  1.00] at ( 34.80,104.55) {0.4};

\node[text=drawColor,rotate= 90.00,anchor=base,inner sep=0pt, outer sep=0pt, scale=  1.00] at ( 34.80,143.96) {0.8};

\path[draw=drawColor,line width= 0.4pt,line join=round,line cap=round] ( 49.20, 61.20) --
	(227.75, 61.20) --
	(227.75,167.61) --
	( 49.20,167.61) --
	cycle;
\end{scope}
\begin{scope}
\path[clip] (  0.00,  0.00) rectangle (252.94,216.81);
\definecolor{drawColor}{RGB}{0,0,0}

\node[text=drawColor,anchor=base,inner sep=0pt, outer sep=0pt, scale=  1.20] at (138.47,188.07) {\bfseries p=15: Criteria A};

\node[text=drawColor,anchor=base,inner sep=0pt, outer sep=0pt, scale=  1.00] at (138.47, 15.60) {Sample Size};

\node[text=drawColor,rotate= 90.00,anchor=base,inner sep=0pt, outer sep=0pt, scale=  1.00] at ( 10.80,114.41) {Error Rate};
\end{scope}
\begin{scope}
\path[clip] ( 49.20, 61.20) rectangle (227.75,167.61);
\definecolor{drawColor}{RGB}{160,32,240}

\path[draw=drawColor,line width= 0.4pt,line join=round,line cap=round] ( 55.81,163.67) circle (  2.25);

\path[draw=drawColor,line width= 0.4pt,line join=round,line cap=round] ( 53.56,163.67) -- ( 58.06,163.67);

\path[draw=drawColor,line width= 0.4pt,line join=round,line cap=round] ( 55.81,161.42) -- ( 55.81,165.92);

\path[draw=drawColor,line width= 0.4pt,line join=round,line cap=round] (129.29,158.74) circle (  2.25);

\path[draw=drawColor,line width= 0.4pt,line join=round,line cap=round] (127.04,158.74) -- (131.54,158.74);

\path[draw=drawColor,line width= 0.4pt,line join=round,line cap=round] (129.29,156.49) -- (129.29,160.99);

\path[draw=drawColor,line width= 0.4pt,line join=round,line cap=round] (221.13,111.45) circle (  2.25);

\path[draw=drawColor,line width= 0.4pt,line join=round,line cap=round] (218.88,111.45) -- (223.38,111.45);

\path[draw=drawColor,line width= 0.4pt,line join=round,line cap=round] (221.13,109.20) -- (221.13,113.70);

\path[draw=drawColor,line width= 0.4pt,dash pattern=on 1pt off 3pt on 4pt off 3pt ,line join=round,line cap=round] ( 55.81,163.67) --
	(129.29,158.74) --
	(221.13,111.45);
\definecolor{drawColor}{RGB}{255,0,0}

\path[draw=drawColor,line width= 0.4pt,line join=round,line cap=round] ( 53.56, 77.67) rectangle ( 58.06, 82.17);

\path[draw=drawColor,line width= 0.4pt,line join=round,line cap=round] ( 53.56, 77.67) -- ( 58.06, 82.17);

\path[draw=drawColor,line width= 0.4pt,line join=round,line cap=round] ( 53.56, 82.17) -- ( 58.06, 77.67);

\path[draw=drawColor,line width= 0.4pt,line join=round,line cap=round] (127.04, 62.89) rectangle (131.54, 67.39);

\path[draw=drawColor,line width= 0.4pt,line join=round,line cap=round] (127.04, 62.89) -- (131.54, 67.39);

\path[draw=drawColor,line width= 0.4pt,line join=round,line cap=round] (127.04, 67.39) -- (131.54, 62.89);

\path[draw=drawColor,line width= 0.4pt,line join=round,line cap=round] (218.88, 62.89) rectangle (223.38, 67.39);

\path[draw=drawColor,line width= 0.4pt,line join=round,line cap=round] (218.88, 62.89) -- (223.38, 67.39);

\path[draw=drawColor,line width= 0.4pt,line join=round,line cap=round] (218.88, 67.39) -- (223.38, 62.89);

\path[draw=drawColor,line width= 0.4pt,line join=round,line cap=round] ( 55.81, 79.92) --
	(129.29, 65.14) --
	(221.13, 65.14);
\definecolor{drawColor}{RGB}{0,0,0}

\path[draw=drawColor,line width= 0.4pt,line join=round,line cap=round] (162.44,167.61) rectangle (227.75,131.61);
\definecolor{drawColor}{RGB}{160,32,240}

\path[draw=drawColor,line width= 0.4pt,dash pattern=on 1pt off 3pt on 4pt off 3pt ,line join=round,line cap=round] (165.14,155.61) -- (183.14,155.61);
\definecolor{drawColor}{RGB}{255,0,0}

\path[draw=drawColor,line width= 0.4pt,line join=round,line cap=round] (165.14,143.61) -- (183.14,143.61);
\definecolor{drawColor}{RGB}{160,32,240}

\path[draw=drawColor,line width= 0.4pt,line join=round,line cap=round] (174.14,155.61) circle (  2.25);

\path[draw=drawColor,line width= 0.4pt,line join=round,line cap=round] (171.89,155.61) -- (176.39,155.61);

\path[draw=drawColor,line width= 0.4pt,line join=round,line cap=round] (174.14,153.36) -- (174.14,157.86);
\definecolor{drawColor}{RGB}{255,0,0}

\path[draw=drawColor,line width= 0.4pt,line join=round,line cap=round] (171.89,141.36) rectangle (176.39,145.86);

\path[draw=drawColor,line width= 0.4pt,line join=round,line cap=round] (171.89,141.36) -- (176.39,145.86);

\path[draw=drawColor,line width= 0.4pt,line join=round,line cap=round] (171.89,145.86) -- (176.39,141.36);
\definecolor{drawColor}{RGB}{0,0,0}

\node[text=drawColor,anchor=base west,inner sep=0pt, outer sep=0pt, scale=  1.00] at (192.14,152.17) {Copula};

\node[text=drawColor,anchor=base west,inner sep=0pt, outer sep=0pt, scale=  1.00] at (192.14,140.17) {MMI};
\end{scope}
\end{tikzpicture}

%% file: result_101.tex
\begin{tikzpicture}[x=1pt,y=1pt]
\definecolor{fillColor}{RGB}{255,255,255}
\path[use as bounding box,fill=fillColor,fill opacity=0.00] (0,0) rectangle (252.94,216.81);
\begin{scope}
\path[clip] (  0.00,  0.00) rectangle (252.94,216.81);
\definecolor{drawColor}{RGB}{0,0,0}

\path[draw=drawColor,line width= 0.4pt,line join=round,line cap=round] ( 74.18, 61.20) -- (221.13, 61.20);

\path[draw=drawColor,line width= 0.4pt,line join=round,line cap=round] ( 74.18, 61.20) -- ( 74.18, 55.20);

\path[draw=drawColor,line width= 0.4pt,line join=round,line cap=round] (110.92, 61.20) -- (110.92, 55.20);

\path[draw=drawColor,line width= 0.4pt,line join=round,line cap=round] (147.66, 61.20) -- (147.66, 55.20);

\path[draw=drawColor,line width= 0.4pt,line join=round,line cap=round] (184.39, 61.20) -- (184.39, 55.20);

\path[draw=drawColor,line width= 0.4pt,line join=round,line cap=round] (221.13, 61.20) -- (221.13, 55.20);

\node[text=drawColor,anchor=base,inner sep=0pt, outer sep=0pt, scale=  1.00] at ( 74.18, 39.60) {200};

\node[text=drawColor,anchor=base,inner sep=0pt, outer sep=0pt, scale=  1.00] at (110.92, 39.60) {400};

\node[text=drawColor,anchor=base,inner sep=0pt, outer sep=0pt, scale=  1.00] at (147.66, 39.60) {600};

\node[text=drawColor,anchor=base,inner sep=0pt, outer sep=0pt, scale=  1.00] at (184.39, 39.60) {800};

\node[text=drawColor,anchor=base,inner sep=0pt, outer sep=0pt, scale=  1.00] at (221.13, 39.60) {1000};

\path[draw=drawColor,line width= 0.4pt,line join=round,line cap=round] ( 49.20, 65.14) -- ( 49.20,163.67);

\path[draw=drawColor,line width= 0.4pt,line join=round,line cap=round] ( 49.20, 65.14) -- ( 43.20, 65.14);

\path[draw=drawColor,line width= 0.4pt,line join=round,line cap=round] ( 49.20, 84.85) -- ( 43.20, 84.85);

\path[draw=drawColor,line width= 0.4pt,line join=round,line cap=round] ( 49.20,104.55) -- ( 43.20,104.55);

\path[draw=drawColor,line width= 0.4pt,line join=round,line cap=round] ( 49.20,124.26) -- ( 43.20,124.26);

\path[draw=drawColor,line width= 0.4pt,line join=round,line cap=round] ( 49.20,143.96) -- ( 43.20,143.96);

\path[draw=drawColor,line width= 0.4pt,line join=round,line cap=round] ( 49.20,163.67) -- ( 43.20,163.67);

\node[text=drawColor,rotate= 90.00,anchor=base,inner sep=0pt, outer sep=0pt, scale=  1.00] at ( 34.80, 65.14) {0.0};

\node[text=drawColor,rotate= 90.00,anchor=base,inner sep=0pt, outer sep=0pt, scale=  1.00] at ( 34.80,104.55) {0.4};

\node[text=drawColor,rotate= 90.00,anchor=base,inner sep=0pt, outer sep=0pt, scale=  1.00] at ( 34.80,143.96) {0.8};

\path[draw=drawColor,line width= 0.4pt,line join=round,line cap=round] ( 49.20, 61.20) --
	(227.75, 61.20) --
	(227.75,167.61) --
	( 49.20,167.61) --
	cycle;
\end{scope}
\begin{scope}
\path[clip] (  0.00,  0.00) rectangle (252.94,216.81);
\definecolor{drawColor}{RGB}{0,0,0}

\node[text=drawColor,anchor=base,inner sep=0pt, outer sep=0pt, scale=  1.20] at (138.47,188.07) {\bfseries p=15: Criteria B};

\node[text=drawColor,anchor=base,inner sep=0pt, outer sep=0pt, scale=  1.00] at (138.47, 15.60) {Sample Size};

\node[text=drawColor,rotate= 90.00,anchor=base,inner sep=0pt, outer sep=0pt, scale=  1.00] at ( 10.80,114.41) {Error Rate};
\end{scope}
\begin{scope}
\path[clip] ( 49.20, 61.20) rectangle (227.75,167.61);
\definecolor{drawColor}{RGB}{160,32,240}

\path[draw=drawColor,line width= 0.4pt,line join=round,line cap=round] ( 55.81,105.54) circle (  2.25);

\path[draw=drawColor,line width= 0.4pt,line join=round,line cap=round] ( 53.56,105.54) -- ( 58.06,105.54);

\path[draw=drawColor,line width= 0.4pt,line join=round,line cap=round] ( 55.81,103.29) -- ( 55.81,107.79);

\path[draw=drawColor,line width= 0.4pt,line join=round,line cap=round] (129.29,107.51) circle (  2.25);

\path[draw=drawColor,line width= 0.4pt,line join=round,line cap=round] (127.04,107.51) -- (131.54,107.51);

\path[draw=drawColor,line width= 0.4pt,line join=round,line cap=round] (129.29,105.26) -- (129.29,109.76);

\path[draw=drawColor,line width= 0.4pt,line join=round,line cap=round] (221.13, 88.79) circle (  2.25);

\path[draw=drawColor,line width= 0.4pt,line join=round,line cap=round] (218.88, 88.79) -- (223.38, 88.79);

\path[draw=drawColor,line width= 0.4pt,line join=round,line cap=round] (221.13, 86.54) -- (221.13, 91.04);

\path[draw=drawColor,line width= 0.4pt,dash pattern=on 1pt off 3pt on 4pt off 3pt ,line join=round,line cap=round] ( 55.81,105.54) --
	(129.29,107.51) --
	(221.13, 88.79);
\definecolor{drawColor}{RGB}{255,0,0}

\path[draw=drawColor,line width= 0.4pt,line join=round,line cap=round] ( 53.56, 69.79) rectangle ( 58.06, 74.29);

\path[draw=drawColor,line width= 0.4pt,line join=round,line cap=round] ( 53.56, 69.79) -- ( 58.06, 74.29);

\path[draw=drawColor,line width= 0.4pt,line join=round,line cap=round] ( 53.56, 74.29) -- ( 58.06, 69.79);

\path[draw=drawColor,line width= 0.4pt,line join=round,line cap=round] (127.04, 62.89) rectangle (131.54, 67.39);

\path[draw=drawColor,line width= 0.4pt,line join=round,line cap=round] (127.04, 62.89) -- (131.54, 67.39);

\path[draw=drawColor,line width= 0.4pt,line join=round,line cap=round] (127.04, 67.39) -- (131.54, 62.89);

\path[draw=drawColor,line width= 0.4pt,line join=round,line cap=round] (218.88, 62.89) rectangle (223.38, 67.39);

\path[draw=drawColor,line width= 0.4pt,line join=round,line cap=round] (218.88, 62.89) -- (223.38, 67.39);

\path[draw=drawColor,line width= 0.4pt,line join=round,line cap=round] (218.88, 67.39) -- (223.38, 62.89);

\path[draw=drawColor,line width= 0.4pt,line join=round,line cap=round] ( 55.81, 72.04) --
	(129.29, 65.14) --
	(221.13, 65.14);
\definecolor{drawColor}{RGB}{0,0,0}

\path[draw=drawColor,line width= 0.4pt,line join=round,line cap=round] (162.44,167.61) rectangle (227.75,131.61);
\definecolor{drawColor}{RGB}{160,32,240}

\path[draw=drawColor,line width= 0.4pt,dash pattern=on 1pt off 3pt on 4pt off 3pt ,line join=round,line cap=round] (165.14,155.61) -- (183.14,155.61);
\definecolor{drawColor}{RGB}{255,0,0}

\path[draw=drawColor,line width= 0.4pt,line join=round,line cap=round] (165.14,143.61) -- (183.14,143.61);
\definecolor{drawColor}{RGB}{160,32,240}

\path[draw=drawColor,line width= 0.4pt,line join=round,line cap=round] (174.14,155.61) circle (  2.25);

\path[draw=drawColor,line width= 0.4pt,line join=round,line cap=round] (171.89,155.61) -- (176.39,155.61);

\path[draw=drawColor,line width= 0.4pt,line join=round,line cap=round] (174.14,153.36) -- (174.14,157.86);
\definecolor{drawColor}{RGB}{255,0,0}

\path[draw=drawColor,line width= 0.4pt,line join=round,line cap=round] (171.89,141.36) rectangle (176.39,145.86);

\path[draw=drawColor,line width= 0.4pt,line join=round,line cap=round] (171.89,141.36) -- (176.39,145.86);

\path[draw=drawColor,line width= 0.4pt,line join=round,line cap=round] (171.89,145.86) -- (176.39,141.36);
\definecolor{drawColor}{RGB}{0,0,0}

\node[text=drawColor,anchor=base west,inner sep=0pt, outer sep=0pt, scale=  1.00] at (192.14,152.17) {Copula};

\node[text=drawColor,anchor=base west,inner sep=0pt, outer sep=0pt, scale=  1.00] at (192.14,140.17) {MMI};
\end{scope}
\end{tikzpicture}

%% file: result_1111.tex
\begin{tikzpicture}[x=1pt,y=1pt]
\definecolor{fillColor}{RGB}{255,255,255}
\path[use as bounding box,fill=fillColor,fill opacity=0.00] (0,0) rectangle (252.94,216.81);
\begin{scope}
\path[clip] (  0.00,  0.00) rectangle (252.94,216.81);
\definecolor{drawColor}{RGB}{0,0,0}

\path[draw=drawColor,line width= 0.4pt,line join=round,line cap=round] ( 74.18, 61.20) -- (221.13, 61.20);

\path[draw=drawColor,line width= 0.4pt,line join=round,line cap=round] ( 74.18, 61.20) -- ( 74.18, 55.20);

\path[draw=drawColor,line width= 0.4pt,line join=round,line cap=round] (110.92, 61.20) -- (110.92, 55.20);

\path[draw=drawColor,line width= 0.4pt,line join=round,line cap=round] (147.66, 61.20) -- (147.66, 55.20);

\path[draw=drawColor,line width= 0.4pt,line join=round,line cap=round] (184.39, 61.20) -- (184.39, 55.20);

\path[draw=drawColor,line width= 0.4pt,line join=round,line cap=round] (221.13, 61.20) -- (221.13, 55.20);

\node[text=drawColor,anchor=base,inner sep=0pt, outer sep=0pt, scale=  1.00] at ( 74.18, 39.60) {200};

\node[text=drawColor,anchor=base,inner sep=0pt, outer sep=0pt, scale=  1.00] at (110.92, 39.60) {400};

\node[text=drawColor,anchor=base,inner sep=0pt, outer sep=0pt, scale=  1.00] at (147.66, 39.60) {600};

\node[text=drawColor,anchor=base,inner sep=0pt, outer sep=0pt, scale=  1.00] at (184.39, 39.60) {800};

\node[text=drawColor,anchor=base,inner sep=0pt, outer sep=0pt, scale=  1.00] at (221.13, 39.60) {1000};

\path[draw=drawColor,line width= 0.4pt,line join=round,line cap=round] ( 49.20, 65.14) -- ( 49.20,163.67);

\path[draw=drawColor,line width= 0.4pt,line join=round,line cap=round] ( 49.20, 65.14) -- ( 43.20, 65.14);

\path[draw=drawColor,line width= 0.4pt,line join=round,line cap=round] ( 49.20, 84.85) -- ( 43.20, 84.85);

\path[draw=drawColor,line width= 0.4pt,line join=round,line cap=round] ( 49.20,104.55) -- ( 43.20,104.55);

\path[draw=drawColor,line width= 0.4pt,line join=round,line cap=round] ( 49.20,124.26) -- ( 43.20,124.26);

\path[draw=drawColor,line width= 0.4pt,line join=round,line cap=round] ( 49.20,143.96) -- ( 43.20,143.96);

\path[draw=drawColor,line width= 0.4pt,line join=round,line cap=round] ( 49.20,163.67) -- ( 43.20,163.67);

\node[text=drawColor,rotate= 90.00,anchor=base,inner sep=0pt, outer sep=0pt, scale=  1.00] at ( 34.80, 65.14) {0.0};

\node[text=drawColor,rotate= 90.00,anchor=base,inner sep=0pt, outer sep=0pt, scale=  1.00] at ( 34.80,104.55) {0.4};

\node[text=drawColor,rotate= 90.00,anchor=base,inner sep=0pt, outer sep=0pt, scale=  1.00] at ( 34.80,143.96) {0.8};

\path[draw=drawColor,line width= 0.4pt,line join=round,line cap=round] ( 49.20, 61.20) --
	(227.75, 61.20) --
	(227.75,167.61) --
	( 49.20,167.61) --
	cycle;
\end{scope}
\begin{scope}
\path[clip] (  0.00,  0.00) rectangle (252.94,216.81);
\definecolor{drawColor}{RGB}{0,0,0}

\node[text=drawColor,anchor=base,inner sep=0pt, outer sep=0pt, scale=  1.20] at (138.47,188.07) {\bfseries p=15: Criteria A};

\node[text=drawColor,anchor=base,inner sep=0pt, outer sep=0pt, scale=  1.00] at (138.47, 15.60) {Sample Size};

\node[text=drawColor,rotate= 90.00,anchor=base,inner sep=0pt, outer sep=0pt, scale=  1.00] at ( 10.80,114.41) {Error Rate};
\end{scope}
\begin{scope}
\path[clip] ( 49.20, 61.20) rectangle (227.75,167.61);
\definecolor{drawColor}{RGB}{160,32,240}

\path[draw=drawColor,line width= 0.4pt,line join=round,line cap=round] ( 55.81,163.67) circle (  2.25);

\path[draw=drawColor,line width= 0.4pt,line join=round,line cap=round] ( 53.56,163.67) -- ( 58.06,163.67);

\path[draw=drawColor,line width= 0.4pt,line join=round,line cap=round] ( 55.81,161.42) -- ( 55.81,165.92);

\path[draw=drawColor,line width= 0.4pt,line join=round,line cap=round] (129.29,163.67) circle (  2.25);

\path[draw=drawColor,line width= 0.4pt,line join=round,line cap=round] (127.04,163.67) -- (131.54,163.67);

\path[draw=drawColor,line width= 0.4pt,line join=round,line cap=round] (129.29,161.42) -- (129.29,165.92);

\path[draw=drawColor,line width= 0.4pt,line join=round,line cap=round] (221.13,163.67) circle (  2.25);

\path[draw=drawColor,line width= 0.4pt,line join=round,line cap=round] (218.88,163.67) -- (223.38,163.67);

\path[draw=drawColor,line width= 0.4pt,line join=round,line cap=round] (221.13,161.42) -- (221.13,165.92);

\path[draw=drawColor,line width= 0.4pt,dash pattern=on 1pt off 3pt on 4pt off 3pt ,line join=round,line cap=round] ( 55.81,163.67) --
	(129.29,163.67) --
	(221.13,163.67);
\definecolor{drawColor}{RGB}{255,0,0}

\path[draw=drawColor,line width= 0.4pt,line join=round,line cap=round] ( 53.56, 80.63) rectangle ( 58.06, 85.13);

\path[draw=drawColor,line width= 0.4pt,line join=round,line cap=round] ( 53.56, 80.63) -- ( 58.06, 85.13);

\path[draw=drawColor,line width= 0.4pt,line join=round,line cap=round] ( 53.56, 85.13) -- ( 58.06, 80.63);

\path[draw=drawColor,line width= 0.4pt,line join=round,line cap=round] (127.04, 62.89) rectangle (131.54, 67.39);

\path[draw=drawColor,line width= 0.4pt,line join=round,line cap=round] (127.04, 62.89) -- (131.54, 67.39);

\path[draw=drawColor,line width= 0.4pt,line join=round,line cap=round] (127.04, 67.39) -- (131.54, 62.89);

\path[draw=drawColor,line width= 0.4pt,line join=round,line cap=round] (218.88, 62.89) rectangle (223.38, 67.39);

\path[draw=drawColor,line width= 0.4pt,line join=round,line cap=round] (218.88, 62.89) -- (223.38, 67.39);

\path[draw=drawColor,line width= 0.4pt,line join=round,line cap=round] (218.88, 67.39) -- (223.38, 62.89);

\path[draw=drawColor,line width= 0.4pt,line join=round,line cap=round] ( 55.81, 82.88) --
	(129.29, 65.14) --
	(221.13, 65.14);
\definecolor{drawColor}{RGB}{0,0,0}

\path[draw=drawColor,line width= 0.4pt,line join=round,line cap=round] (162.44,167.61) rectangle (227.75,131.61);
\definecolor{drawColor}{RGB}{160,32,240}

\path[draw=drawColor,line width= 0.4pt,dash pattern=on 1pt off 3pt on 4pt off 3pt ,line join=round,line cap=round] (165.14,155.61) -- (183.14,155.61);
\definecolor{drawColor}{RGB}{255,0,0}

\path[draw=drawColor,line width= 0.4pt,line join=round,line cap=round] (165.14,143.61) -- (183.14,143.61);
\definecolor{drawColor}{RGB}{160,32,240}

\path[draw=drawColor,line width= 0.4pt,line join=round,line cap=round] (174.14,155.61) circle (  2.25);

\path[draw=drawColor,line width= 0.4pt,line join=round,line cap=round] (171.89,155.61) -- (176.39,155.61);

\path[draw=drawColor,line width= 0.4pt,line join=round,line cap=round] (174.14,153.36) -- (174.14,157.86);
\definecolor{drawColor}{RGB}{255,0,0}

\path[draw=drawColor,line width= 0.4pt,line join=round,line cap=round] (171.89,141.36) rectangle (176.39,145.86);

\path[draw=drawColor,line width= 0.4pt,line join=round,line cap=round] (171.89,141.36) -- (176.39,145.86);

\path[draw=drawColor,line width= 0.4pt,line join=round,line cap=round] (171.89,145.86) -- (176.39,141.36);
\definecolor{drawColor}{RGB}{0,0,0}

\node[text=drawColor,anchor=base west,inner sep=0pt, outer sep=0pt, scale=  1.00] at (192.14,152.17) {Copula};

\node[text=drawColor,anchor=base west,inner sep=0pt, outer sep=0pt, scale=  1.00] at (192.14,140.17) {MMI};
\end{scope}
\end{tikzpicture}

%% file: result_121.tex
\begin{tikzpicture}[x=1pt,y=1pt]
\definecolor{fillColor}{RGB}{255,255,255}
\path[use as bounding box,fill=fillColor,fill opacity=0.00] (0,0) rectangle (252.94,216.81);
\begin{scope}
\path[clip] (  0.00,  0.00) rectangle (252.94,216.81);
\definecolor{drawColor}{RGB}{0,0,0}

\path[draw=drawColor,line width= 0.4pt,line join=round,line cap=round] ( 74.18, 61.20) -- (221.13, 61.20);

\path[draw=drawColor,line width= 0.4pt,line join=round,line cap=round] ( 74.18, 61.20) -- ( 74.18, 55.20);

\path[draw=drawColor,line width= 0.4pt,line join=round,line cap=round] (110.92, 61.20) -- (110.92, 55.20);

\path[draw=drawColor,line width= 0.4pt,line join=round,line cap=round] (147.66, 61.20) -- (147.66, 55.20);

\path[draw=drawColor,line width= 0.4pt,line join=round,line cap=round] (184.39, 61.20) -- (184.39, 55.20);

\path[draw=drawColor,line width= 0.4pt,line join=round,line cap=round] (221.13, 61.20) -- (221.13, 55.20);

\node[text=drawColor,anchor=base,inner sep=0pt, outer sep=0pt, scale=  1.00] at ( 74.18, 39.60) {200};

\node[text=drawColor,anchor=base,inner sep=0pt, outer sep=0pt, scale=  1.00] at (110.92, 39.60) {400};

\node[text=drawColor,anchor=base,inner sep=0pt, outer sep=0pt, scale=  1.00] at (147.66, 39.60) {600};

\node[text=drawColor,anchor=base,inner sep=0pt, outer sep=0pt, scale=  1.00] at (184.39, 39.60) {800};

\node[text=drawColor,anchor=base,inner sep=0pt, outer sep=0pt, scale=  1.00] at (221.13, 39.60) {1000};

\path[draw=drawColor,line width= 0.4pt,line join=round,line cap=round] ( 49.20, 65.14) -- ( 49.20,163.67);

\path[draw=drawColor,line width= 0.4pt,line join=round,line cap=round] ( 49.20, 65.14) -- ( 43.20, 65.14);

\path[draw=drawColor,line width= 0.4pt,line join=round,line cap=round] ( 49.20, 84.85) -- ( 43.20, 84.85);

\path[draw=drawColor,line width= 0.4pt,line join=round,line cap=round] ( 49.20,104.55) -- ( 43.20,104.55);

\path[draw=drawColor,line width= 0.4pt,line join=round,line cap=round] ( 49.20,124.26) -- ( 43.20,124.26);

\path[draw=drawColor,line width= 0.4pt,line join=round,line cap=round] ( 49.20,143.96) -- ( 43.20,143.96);

\path[draw=drawColor,line width= 0.4pt,line join=round,line cap=round] ( 49.20,163.67) -- ( 43.20,163.67);

\node[text=drawColor,rotate= 90.00,anchor=base,inner sep=0pt, outer sep=0pt, scale=  1.00] at ( 34.80, 65.14) {0.0};

\node[text=drawColor,rotate= 90.00,anchor=base,inner sep=0pt, outer sep=0pt, scale=  1.00] at ( 34.80,104.55) {0.4};

\node[text=drawColor,rotate= 90.00,anchor=base,inner sep=0pt, outer sep=0pt, scale=  1.00] at ( 34.80,143.96) {0.8};

\path[draw=drawColor,line width= 0.4pt,line join=round,line cap=round] ( 49.20, 61.20) --
	(227.75, 61.20) --
	(227.75,167.61) --
	( 49.20,167.61) --
	cycle;
\end{scope}
\begin{scope}
\path[clip] (  0.00,  0.00) rectangle (252.94,216.81);
\definecolor{drawColor}{RGB}{0,0,0}

\node[text=drawColor,anchor=base,inner sep=0pt, outer sep=0pt, scale=  1.20] at (138.47,188.07) {\bfseries p=15: Criteria B};

\node[text=drawColor,anchor=base,inner sep=0pt, outer sep=0pt, scale=  1.00] at (138.47, 15.60) {Sample Size};

\node[text=drawColor,rotate= 90.00,anchor=base,inner sep=0pt, outer sep=0pt, scale=  1.00] at ( 10.80,114.41) {Error Rate};
\end{scope}
\begin{scope}
\path[clip] ( 49.20, 61.20) rectangle (227.75,167.61);
\definecolor{drawColor}{RGB}{160,32,240}

\path[draw=drawColor,line width= 0.4pt,line join=round,line cap=round] ( 55.81,105.54) circle (  2.25);

\path[draw=drawColor,line width= 0.4pt,line join=round,line cap=round] ( 53.56,105.54) -- ( 58.06,105.54);

\path[draw=drawColor,line width= 0.4pt,line join=round,line cap=round] ( 55.81,103.29) -- ( 55.81,107.79);

\path[draw=drawColor,line width= 0.4pt,line join=round,line cap=round] (129.29,108.49) circle (  2.25);

\path[draw=drawColor,line width= 0.4pt,line join=round,line cap=round] (127.04,108.49) -- (131.54,108.49);

\path[draw=drawColor,line width= 0.4pt,line join=round,line cap=round] (129.29,106.24) -- (129.29,110.74);

\path[draw=drawColor,line width= 0.4pt,line join=round,line cap=round] (221.13,106.52) circle (  2.25);

\path[draw=drawColor,line width= 0.4pt,line join=round,line cap=round] (218.88,106.52) -- (223.38,106.52);

\path[draw=drawColor,line width= 0.4pt,line join=round,line cap=round] (221.13,104.27) -- (221.13,108.77);

\path[draw=drawColor,line width= 0.4pt,dash pattern=on 1pt off 3pt on 4pt off 3pt ,line join=round,line cap=round] ( 55.81,105.54) --
	(129.29,108.49) --
	(221.13,106.52);
\definecolor{drawColor}{RGB}{255,0,0}

\path[draw=drawColor,line width= 0.4pt,line join=round,line cap=round] ( 53.56, 62.89) rectangle ( 58.06, 67.39);

\path[draw=drawColor,line width= 0.4pt,line join=round,line cap=round] ( 53.56, 62.89) -- ( 58.06, 67.39);

\path[draw=drawColor,line width= 0.4pt,line join=round,line cap=round] ( 53.56, 67.39) -- ( 58.06, 62.89);

\path[draw=drawColor,line width= 0.4pt,line join=round,line cap=round] (127.04, 62.89) rectangle (131.54, 67.39);

\path[draw=drawColor,line width= 0.4pt,line join=round,line cap=round] (127.04, 62.89) -- (131.54, 67.39);

\path[draw=drawColor,line width= 0.4pt,line join=round,line cap=round] (127.04, 67.39) -- (131.54, 62.89);

\path[draw=drawColor,line width= 0.4pt,line join=round,line cap=round] (218.88, 62.89) rectangle (223.38, 67.39);

\path[draw=drawColor,line width= 0.4pt,line join=round,line cap=round] (218.88, 62.89) -- (223.38, 67.39);

\path[draw=drawColor,line width= 0.4pt,line join=round,line cap=round] (218.88, 67.39) -- (223.38, 62.89);

\path[draw=drawColor,line width= 0.4pt,line join=round,line cap=round] ( 55.81, 65.14) --
	(129.29, 65.14) --
	(221.13, 65.14);
\definecolor{drawColor}{RGB}{0,0,0}

\path[draw=drawColor,line width= 0.4pt,line join=round,line cap=round] (162.44,167.61) rectangle (227.75,131.61);
\definecolor{drawColor}{RGB}{160,32,240}

\path[draw=drawColor,line width= 0.4pt,dash pattern=on 1pt off 3pt on 4pt off 3pt ,line join=round,line cap=round] (165.14,155.61) -- (183.14,155.61);
\definecolor{drawColor}{RGB}{255,0,0}

\path[draw=drawColor,line width= 0.4pt,line join=round,line cap=round] (165.14,143.61) -- (183.14,143.61);
\definecolor{drawColor}{RGB}{160,32,240}

\path[draw=drawColor,line width= 0.4pt,line join=round,line cap=round] (174.14,155.61) circle (  2.25);

\path[draw=drawColor,line width= 0.4pt,line join=round,line cap=round] (171.89,155.61) -- (176.39,155.61);

\path[draw=drawColor,line width= 0.4pt,line join=round,line cap=round] (174.14,153.36) -- (174.14,157.86);
\definecolor{drawColor}{RGB}{255,0,0}

\path[draw=drawColor,line width= 0.4pt,line join=round,line cap=round] (171.89,141.36) rectangle (176.39,145.86);

\path[draw=drawColor,line width= 0.4pt,line join=round,line cap=round] (171.89,141.36) -- (176.39,145.86);

\path[draw=drawColor,line width= 0.4pt,line join=round,line cap=round] (171.89,145.86) -- (176.39,141.36);
\definecolor{drawColor}{RGB}{0,0,0}

\node[text=drawColor,anchor=base west,inner sep=0pt, outer sep=0pt, scale=  1.00] at (192.14,152.17) {Copula};

\node[text=drawColor,anchor=base west,inner sep=0pt, outer sep=0pt, scale=  1.00] at (192.14,140.17) {MMI};
\end{scope}
\end{tikzpicture}

%% file: fig_new_new.tex
\begin{center}
\setlength\unitlength{0.8mm}
\begin{picture}(150,50)
\put(20,5){\circle{10}}
\put(20,25){\circle{10}}
\put(20,45){\circle{10}}
\put(50,15){\circle{10}}
\put(50,35){\circle{10}}
\put(75,25){\circle{10}}

\put(20,5){\makebox(0,0)[c]{$X_6$}}
\put(20,25){\makebox(0,0)[c]{$X_1$}}
\put(20,45){\makebox(0,0)[c]{$X_3$}}
\put(50,15){\makebox(0,0)[c]{$X_4$}}
\put(50,35){\makebox(0,0)[c]{$X_5$}}
\put(75,25){\makebox(0,0)[c]{$X_2$}}

\put(20,40){\vector(0,-1){10}}
\put(20,20){\vector(0,-1){10}}
\put(25,45){\vector(2,-1){20}}
\put(25,25){\vector(2,1){20}}
\put(25,25){\vector(2,-1){20}}
\put(25,5){\vector(2,1){20}}
\put(25,5){\vector(2,3){20}}
\put(55,35){\vector(3,-2){15}}
\put(55,15){\vector(3,2){15}}
\put(5,45){\line(0,-1){40}}
\put(15,45){\line(-1,0){10}}
\put(5,5){\vector(1,0){10}}
\put(50,30){\vector(0,-1){10}}
\put(90,25){
\begin{tabular}{ll}
$X_1$:& Father's Occupation \\
$X_2$:& Son's Income \\
$X_3$:& Father's Education \\
$X_4$:& Son's Occupation \\
$X_5$:& Son's Education \\
$X_6$:& Number of Siblings
\end{tabular}
}
\end{picture}
\end{center}